\documentclass{article}

\usepackage[main, final]{neurips_2026}

\usepackage[utf8]{inputenc}
\usepackage[T1]{fontenc}
\usepackage{hyperref}
\usepackage{url}
\usepackage{booktabs}
\usepackage{amsfonts}
\usepackage{nicefrac}
\usepackage{microtype}
\usepackage{xcolor}
\usepackage{graphicx}
\usepackage{amsmath}
\usepackage{multirow}
\usepackage{array}
\usepackage{enumitem}
\usepackage{float}
\usepackage{tikz}
\usepackage{fontawesome5}
\usepackage{amssymb}
\usetikzlibrary{positioning, arrows.meta, shapes, fit, backgrounds, calc, patterns}

\definecolor{neuripsblue}{RGB}{31, 119, 180}
\definecolor{neuripsorange}{RGB}{255, 127, 14}
\definecolor{neuripsgreen}{RGB}{44, 160, 44}
\definecolor{neuripsred}{RGB}{214, 39, 40}
\definecolor{neuripsgray}{RGB}{127, 127, 127}

\definecolor{innate1}{RGB}{31, 119, 180}
\definecolor{innate2}{RGB}{158, 202, 225}
\definecolor{cons1}{RGB}{230, 85, 13}
\definecolor{cons2}{RGB}{253, 174, 107}
\definecolor{cons3}{RGB}{254, 230, 206}

\definecolor{tokpatch}{RGB}{60, 90, 130}
\definecolor{tokvariate}{RGB}{120, 80, 130}
\definecolor{tokpoint}{RGB}{130, 80, 60}

\newcommand{\stdv}[1]{{\scriptsize\textcolor{neuripsgray}{$\pm$#1}}}

\providecommand{\answerYes}[1][]{[Yes]#1}
\providecommand{\answerNo}[1][]{[No]#1}
\providecommand{\answerNA}[1][]{[NA]#1}

\title{Tokenization and Architecture Jointly Allocate Component Roles in Time Series Transformers}

\author{%
  Mingyu Kim \\
  Department of Artificial Intelligence \\
  Inha University \\
  Incheon, Republic of Korea \\
  \texttt{mingyu@inha.edu} \\
  \And
  Doguk Kim\thanks{Corresponding author.} \\
  Department of Artificial Intelligence \\
  Inha University \\
  Incheon, Republic of Korea \\
  \texttt{dgkim@inha.ac.kr} \\
}

\begin{document}

\raggedbottom

\maketitle

\begin{abstract}
Time series transformers have achieved strong forecasting performance, yet it remains unclear which of their components, attention or the feed-forward network (FFN), constructs temporal features, and how tokenization such as patching changes this. We propose a temporal probing framework that uses 15 token-level diagnostic tasks to measure feature construction as the gain in linear decodability of temporal properties from one sub-layer to the next (a matched MLP probe already saturates at the embedding). It targets models whose probed representations are read by a linear forecasting head. Applied to 5 supervised models trained on 8 long-term forecasting benchmarks, it finds that the constructing component is not fixed but allocated by the interaction of tokenization and architecture. Under patch tokenization (PatchTST), the FFN's gain exceeds attention's in probing on all 8 datasets, and its 8-dataset mean is 3.2$\times$ attention's. Without attention, the FFN alone reaches nearly the same total gain, both when attention is removed from a trained model and when PatchTST is trained without it, and the retrained model forecasts within a few percent of the full model. In probing, attention dominates under variate tokenization (iTransformer); in CATS, cross-attention is the only input path. In a controlled manipulation of PatchTST, single-scalar tokens ($P{=}1$) nearly eliminate the FFN's construction, while one more point per token ($P{=}2$) restores it. Our forecasting-side interventions (activation patching, knockout, and training without attention) establish component-level reliance and sufficiency. These results offer a representational reading of prior findings on the gains from patching, the competitiveness of self-attention-free models (TSMixer, PatchMLP), and the distinct role of cross-attention in CATS.\\[0.4ex]
\null\hfill\faGithub~\url{https://github.com/mingyu-11/tokenization-component-roles}\hfill\null
\end{abstract}

\section{Introduction}
\label{sec:intro}

Time series transformers have evolved rapidly, and the results reported along the way appear contradictory. Early forecasters such as Informer~\citep{zhou2021informer}, Autoformer~\citep{wu2021autoformer}, and FEDformer~\citep{zhou2022fedformer} use time-step tokens, and a one-layer linear model, DLinear~\citep{zeng2023transformers}, surpassed all of them. PatchTST~\citep{nie2023time} restored transformer dominance through patch tokenization and channel independence (variates modeled separately), and subsequent work questioned whether attention is needed: TSMixer~\citep{ekambaram2023tsmixer} keeps patching but replaces self-attention with MLP-Mixer blocks~\citep{tolstikhin2021mlp}, and PatchMLP~\citep{tang2025patchmlp} attributes the gains mainly to patching rather than attention. CATS~\citep{kim2024cats} replaces self-attention with cross-attention from one learnable query per future patch, and iTransformer~\citep{liu2024itransformer} makes each variate's whole lookback window one token. These results may reflect an underlying pattern: transformers with time-step tokens have been reported to lose to linear models and those with patch or whole-variate tokens to win, and attention can be removed with little or no loss in some models while cross-attention alone suffices in others. The same name, ``attention,'' may thus cover different information flows.

Telling these flows apart requires observing model internals rather than comparing final outputs. We bring linear probing~\citep{alain2016understanding}, well established in NLP (\S\ref{sec:related}), to the sub-layers of supervised models and ask, along two axes, tokenization (point, patch, or variate) and attention type (self-attention, cross-attention, or none), how the relative roles of attention and the feed-forward network (FFN) shift and how the role of the same attention mechanism changes with tokenization. Because the models in our main analysis read their representations through a linear forecasting head, we treat a gain in linear decodability as feature construction and complement probing with forecasting-side interventions. Our contributions are:
\begin{itemize}[topsep=2pt,itemsep=2pt,leftmargin=*]
\item \textbf{Temporal probing framework.} We design 15 token-level diagnostic tasks and a controlled protocol that locates where temporal properties become linearly decodable, with sub-layer extraction, attention/FFN separation, and variants that remove attention or change the patch length or stride.
\item \textbf{Role allocation by tokenization and architecture.} In probing, the dominant constructing component is allocated by the interaction of tokenization and architecture: the FFN under patching with self-attention, attention under variate-level tokenization, the cross-attention path in CATS, and attention under single-scalar tokens (PatchTST with patch length $P{=}1$), where the total gain is smaller. Under patching, the allocation recurs across patch lengths, strides, and horizons (each on 4 datasets) and three added probing targets. On the forecasting side, activation patching, training without attention, and knockout establish component-level reliance and sufficiency in the standard PatchTST, and knockout ranks the architectures by attention's share as probing does.
\item \textbf{A representational reading of prior results.} We read reported results for DLinear, PatchTST, TSMixer, PatchMLP, and CATS qualitatively (\S\ref{sec:unified}) and suggest three design practices (\S\ref{sec:design}).
\end{itemize}

\section{Related Work}
\label{sec:related}

\textbf{Analyses of supervised time series transformers.} \citet{liang2025why} found that transformer blocks often degenerate into MLPs, with performance concentrated in the FFN, and traced this to embeddings that do not let attention operate in a well-structured latent space. They relied on forecasting error and attention weights, whereas we ask where temporal properties become linearly decodable. \citet{chen2025closer} compared point-wise, patch-wise, and variate-wise tokenization through mutual information, identifying intra-variate dependency as the primary contributor to benchmark performance, but did not separate the roles of attention and the FFN. Other analyses cover the geometry of layer representations~\citep{kaufman2024analyzing}, neuron-level attribution~\citep{wang2025pattern}, convergence failures~\citep{ilbert2024samformer}, and concept bottlenecks~\citep{vansprang2024concept}. Two concurrent works take other routes. \citet{yildirim2026superposition} found that the FFN of a single-layer PatchTST variant needs no superposition (packing more features than dimensions) and argued that benchmarks demand little nonlinear feature composition; this, however, does not rule out nonlinearly constructed features, and zeroing that FFN raised MSE in most of their settings. \citet{wang2026combinationts} reported that a parameter-free identity encoder often matches learned encoders once the embedding is well designed.

\textbf{Analyses of time series foundation models.} \citet{wilinski2025exploring} probed the residual stream after each block and found concepts such as trend and periodicity linearly separable at specific layers. \citet{bao2026universal} ablated MLPs and attention heads layer by layer, found that MLP ablation hurts decoder-only models much more than the encoder-decoder Chronos-Bolt, and asked how architecture determines what MLPs and attention heads each carry. \citet{yu2026understanding} showed that the numerical rank of patch embeddings is bounded by a quantity that grows with patch size rather than with the embedding dimension. Our work complements these studies by probing frozen supervised models, each trained on a single dataset, at sub-layer resolution, and \S\ref{sec:limitations} states when our framework covers foundation models.

\textbf{Probing and component roles in NLP.} Probing trains a predictor on frozen representations~\citep{conneau2018what,tenney2019bert,hewitt2019structural,hewitt2019designing,belinkov2022probing}, in its simplest form a linear one~\citep{alain2016understanding}. NLP work has separated component roles: \citet{geva2021kvmemories,geva2022transformer} showed that FFN layers function as key-value memories that build predictions across layers. \citet{elhage2021mathematical} and \citet{olsson2022induction} identified attention's induction heads and their role in in-context learning, building on earlier attention-head analyses~\citep{voita2019analyzing}. \citet{dong2025attention} used MixiT, whose attention scores are random, to show that attention largely drives in-context reasoning while MLPs, together with attention, store knowledge, and \citet{chen2025distributional} showed that feed-forward layers learn distributional associations. Probing shows what is linearly decodable rather than what a model uses, so we pair it with component-level interventions such as activation patching~\citep{meng2022locating} (\S\ref{sec:limitations}).

\section{Method: Temporal Probing Framework}

\subsection{Probing Methodology}

\begin{figure}[!htb]
\centering
\resizebox{0.72\linewidth}{!}{%
\begin{tikzpicture}[
  font=\small,
  >=Stealth,
  block/.style={
    draw, rounded corners=2pt,
    minimum width=1.6cm, minimum height=0.9cm,
    align=center, font=\small, fill=white
  },
  data/.style={
    draw=neuripsblue, line width=0.6pt,
    rounded corners=2pt,
    minimum width=1.6cm, minimum height=0.9cm,
    align=center, font=\small, fill=white
  },
  attn/.style={
    draw=neuripsorange, line width=0.8pt,
    rounded corners=2pt,
    minimum width=1.4cm, minimum height=0.9cm,
    align=center, font=\small,
    fill=neuripsorange!12,
    postaction={pattern=north east lines, pattern color=neuripsorange!30}
  },
  ffn/.style={
    draw=neuripsblue, line width=0.8pt,
    rounded corners=2pt,
    minimum width=1.4cm, minimum height=0.9cm,
    align=center, font=\small,
    fill=neuripsblue!18,
    postaction={pattern=dots, pattern color=neuripsblue!50}
  },
  probe/.style={
    draw=neuripsgray, dashed, line width=0.5pt,
    rounded corners=2pt,
    minimum width=1.5cm, minimum height=0.6cm,
    align=center, font=\footnotesize, fill=white
  },
  task/.style={
    draw=black, line width=0.4pt,
    rounded corners=2pt,
    minimum width=2.6cm, minimum height=0.5cm,
    align=center, font=\footnotesize, fill=gray!8
  },
  task-constructed/.style={
    draw=neuripsgreen, line width=0.7pt,
    rounded corners=2pt,
    minimum width=2.6cm, minimum height=0.5cm,
    align=center, font=\footnotesize, fill=neuripsgreen!18
  },
  task-innate/.style={
    draw=neuripsgreen!50, line width=0.5pt,
    rounded corners=2pt,
    minimum width=2.6cm, minimum height=0.5cm,
    align=center, font=\footnotesize, fill=neuripsgreen!5
  },
  task-more/.style={
    draw=neuripsgray, dashed, line width=0.5pt,
    rounded corners=2pt,
    minimum width=1.0cm, minimum height=0.5cm,
    align=center, font=\footnotesize\itshape, fill=gray!5,
    text=neuripsgray
  },
  arrow/.style={->, thick, black},
  probe-arrow/.style={->, dashed, thick, neuripsgray}
]
\node[data] (input) at (0,0) {Time\\Series};
\node[data] (embed) [right=0.6cm of input] {Token\\Embed};
\node[attn] (attn1) [right=0.7cm of embed] {Attention};
\node[ffn] (ffn1) [right=0.5cm of attn1] {FFN};
\node[block, draw=neuripsgray, fill=gray!3] (more) [right=0.5cm of ffn1, minimum width=0.9cm] {$\cdots$\\Layer $L$};
\draw[arrow] (input) -- (embed);
\draw[arrow] (embed) -- (attn1);
\draw[arrow] (attn1) -- (ffn1);
\draw[arrow] (ffn1) -- (more);
\begin{scope}[on background layer]
  \node[draw=neuripsgray, dashed, line width=0.4pt, rounded corners=4pt, fit=(attn1)(ffn1), inner sep=8pt, label={[neuripsgray, font=\footnotesize]above:Layer 1}] {};
\end{scope}
\node[probe] (p_embed) [below=0.85cm of embed] {embed};
\node[probe, draw=neuripsorange, fill=neuripsorange!8] at (attn1 |- p_embed) (p_attn) {1a\_attn};
\node[probe, draw=neuripsblue, fill=neuripsblue!8] at (ffn1 |- p_embed) (p_ffn) {1b\_ffn};
\node[font=\footnotesize\itshape, neuripsgray] (rep_label) at ($(p_embed.west) + (-0.6,0)$) {\rotatebox{90}{Representations}};
\draw[probe-arrow] (embed.south) -- (p_embed.north);
\draw[probe-arrow] (attn1.south) -- (p_attn.north);
\draw[probe-arrow] (ffn1.south) -- (p_ffn.north);
\node[task] (probe_box) [below=0.75cm of p_attn, minimum width=8.5cm] {%
  \textbf{Linear Probes}: 15 token-level diagnostic tasks};
\draw[probe-arrow] (p_embed.south) -- ([xshift=-2.8cm]probe_box.north);
\draw[probe-arrow] (p_attn.south) -- (probe_box.north);
\draw[probe-arrow] (p_ffn.south) -- ([xshift=2.8cm]probe_box.north);
\node[task-constructed] (t1) [below left=0.45cm and -1.5cm of probe_box] {Volatility (constructed)};
\node[task-constructed] (t2) [right=0.2cm of t1] {Spectral Entropy (constructed)};
\node[task-innate] (t3) [right=0.2cm of t2] {Trend Direction (innate)};
\node[task-more] (tdots) [right=0.2cm of t3] {$+\,12$ more};
\draw[probe-arrow] (probe_box.south) -- ([xshift=0]t1.north);
\draw[probe-arrow] (probe_box.south) -- ([xshift=0]t2.north);
\draw[probe-arrow] (probe_box.south) -- ([xshift=0]t3.north);
\end{tikzpicture}%
}
\caption{\textbf{Probing protocol with delta-based component attribution.} Linear probes on the frozen sub-layer outputs (\texttt{embed}, \texttt{1a\_attn}, \texttt{1b\_ffn}) predict 15 token-level temporal properties (\S\ref{sec:tasks}); per-component deltas ($\Delta$) attribute accuracy gains to sub-layers.}
\label{fig:probing_protocol}
\end{figure}
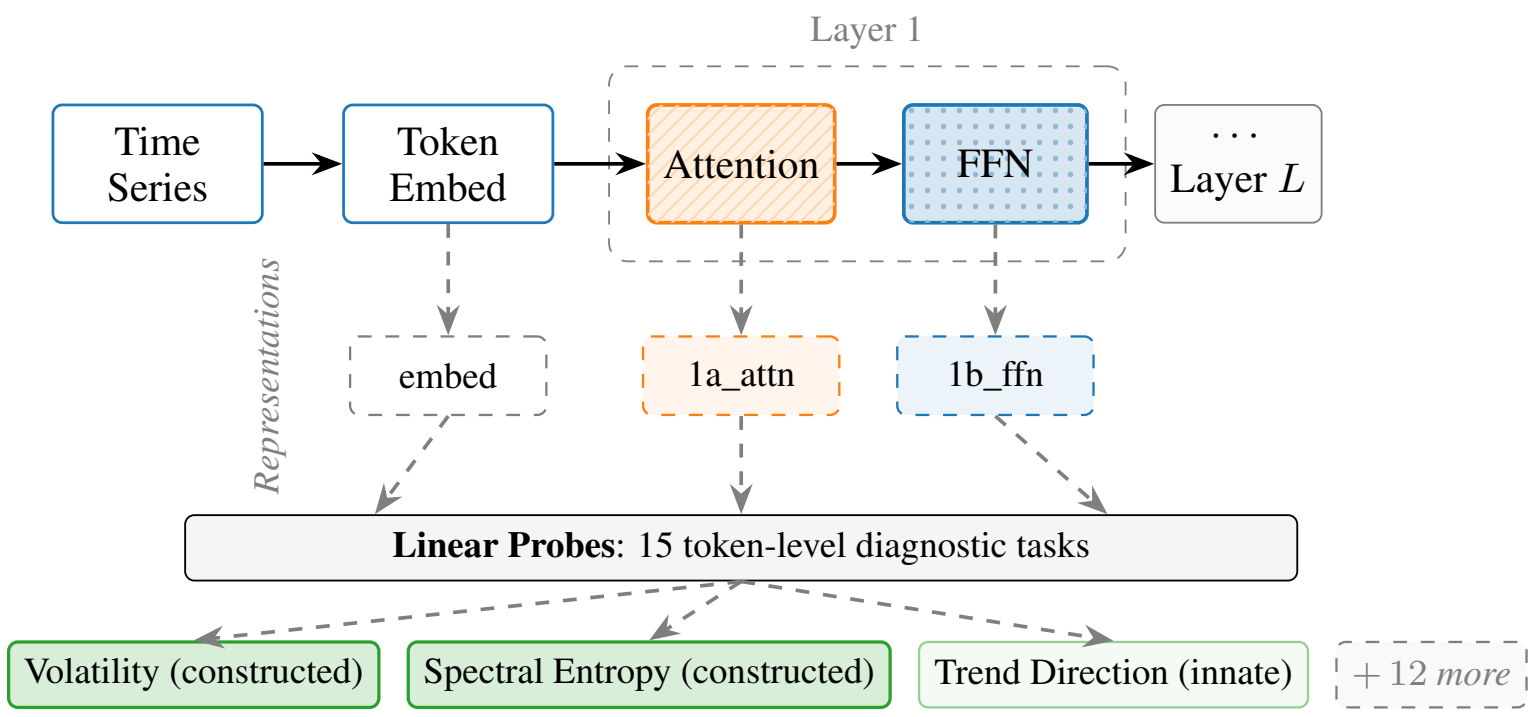

We freeze the weights of a trained transformer and extract intermediate representations at the embedding and after the two standard sub-layers~\citep{vaswani2017attention} of each layer $l$ (Figure~\ref{fig:probing_protocol}): \texttt{embed} (token embedding output), \texttt{$l$a\_attn} (after attention, residual connection, and normalization), and \texttt{$l$b\_ffn} (after FFN, residual connection, and normalization), e.g., \texttt{1a\_attn} and \texttt{1b\_ffn} for Layer 1. On each frozen representation, we train a linear probe~\citep{alain2016understanding} for each temporal property: its input is one token's representation $h \in \mathbb{R}^{d_{\text{model}}}$, and its output, $Wh + b$, has one logit per class of the property of the token's segment (one output for regression). Normalization is BatchNorm in PatchTST and its variants~\citep{nie2023time} and LayerNorm in iTransformer, CATS, and Informer. BatchNorm is affine at inference, so it does not change what a linear probe can decode; LayerNorm is a token-wise nonlinearity, so in these 3 models the attention delta defined below also reflects it.

\textbf{Construction as linear decodability.} A sub-layer \emph{constructs} a property when the property becomes linearly decodable from its output. This is a change in usable information (here, linearly extractable), not Shannon information~\citep{xu2020theory}: PatchTST embeds each token injectively, so no sub-layer adds Shannon information about a property computed from the token's own values~\citep{pimentel2020information}. As a proxy for this change, not an estimate of it, we use the gain in probing accuracy over the preceding probing point, in the spirit of baselined probing~\citep{hewitt2021conditional}:
\vspace{-0.2em}
\begin{equation*}
\Delta_{\text{attn}}(l) = \text{Acc}(\text{$l$a\_attn}) - \text{Acc}(\text{prev}), \qquad
\Delta_{\text{FFN}}(l) = \text{Acc}(\text{$l$b\_ffn}) - \text{Acc}(\text{$l$a\_attn})
\end{equation*}
\vspace{-1.6em}

\noindent where $\text{prev} = \text{embed}$ if $l=1$, else $(l{-}1)\text{b\_ffn}$. A delta near zero or below means no net gain in linear decodability, and the two deltas at a layer indicate which component constructs; a delta does not measure forecasting quality or whether the forecaster uses the property~\citep{elazar2021amnesic}. Like the heads through which these forecasters read their representations, our probes are linear rather than maximally expressive~\citep{pimentel2020information}: a matched MLP probe already saturates at the embedding (Appendix~\ref{app:mlp_probe}). On the forecasting side, three interventions complement probing: activation patching, which restores a component's clean output into a run on corrupted input, tests whether attention contributes at all (in PatchTST); training without attention tests whether the FFN suffices; and knockout, which zeroes a component's output in all layers of a frozen model, tests which component a trained model relies on.

\subsection{Task Design}
\label{sec:tasks}

\begin{table}[t]
\caption{Tasks (left: mainly intra-token; right: inter-token, regression, and control; Cl.: classes).}
\label{tab:tasks}
\centering
\footnotesize
\setlength{\tabcolsep}{2.5pt}
\begin{tabular}{@{}lcll@{\hspace{8pt}}lcll@{}}
\toprule
Task & Cl. & Dom. & Computation & Task & Cl. & Dom. & Computation \\
\midrule
\textbf{volatility}$^{\mathrm{C}}$ & 4 & stat & $\text{var}(\text{patch})$ quartile & adjacent\_similarity & 2 & stat & $\cos(p_i, p_{i+1}) > \text{median}$ \\
\textbf{diff\_energy}$^{\mathrm{C}}$ & 4 & diff & $\text{mean}((\Delta x)^2)$ quartile & relative\_magnitude & 2 & stat & $\text{mean}(p_i) > \text{mean}(\text{all})$ \\
\textbf{spectral\_entropy}$^{\mathrm{C}}$ & 4 & freq & $-\sum q_k \log q_k$ quartile & position\_quarter & 4 & time & token-index quarter \\
trend\_direction$^{\mathrm{I}}$ & 2 & time & $\text{sign}(x_P - x_1)$ & trend\_consistency & 2 & time & 3-token trend alignment \\
\cmidrule(l){5-8}
slope\_magnitude$^{\mathrm{B}}$ & 4 & time & $|\text{linear\_slope}|$ quartile & reg\_variance$^{\mathrm{C}}$ & $R^2$ & stat & $\text{var}(\text{patch})$ continuous \\
change\_point$^{\mathrm{B}}$ & 2 & time & $|\bar{x}_{\text{1st}} - \bar{x}_{\text{2nd}}|$ top 30\% & reg\_slope$^{\mathrm{I}}$ & $R^2$ & time & $x_P - x_1$ continuous \\
dominant\_period$^{\mathrm{D}}$ & 4 & freq & $\arg\max$ of 4 $|\text{rfft}|$ band sums & reg\_next\_mean & $R^2$ & pred & $\text{mean}(\text{next\_patch})$ \\
\cmidrule(l){5-8}
local\_global\_trend & 2 & time & $\text{sign(local)} = \text{sign(global)}$ & random\_label & 4 & --- & random label \\
\bottomrule
\end{tabular}
\end{table}

We design 15 token-level diagnostic tasks (Table~\ref{tab:tasks}), adapted to each model's tokenization (\S\ref{sec:adaptations}): eight \textbf{intra-token}, four \textbf{inter-token}, and three \textbf{regression} tasks, plus a random-label \textbf{control}. The suite is built to locate where temporal properties are constructed, not to inventory what forecasting needs; \S\ref{sec:noattn} tests what it may miss. Table~\ref{tab:tasks} marks innate (I), constructed (C), boundary (B), and degenerate (D) tasks, defined below; unmarked tasks fall outside this classification.

\textbf{Innate and constructed tasks.} PatchTST's embedding is an affine map $W_P\tilde{x} + W_{\text{pos}}$ of the normalized patch $\tilde{x}$ from which intra-token labels are computed, so a linear probe on it can decode only labels that a linear read-out of the patch (and its position) determines. \emph{Innate} tasks such as trend\_direction (whether the patch rises) have such labels. \emph{Constructed} tasks such as volatility, diff energy, and spectral entropy do not: the first two are quadratic forms and the third a scale-invariant function of the power spectrum, all functions of second-order statistics, so they can become linearly decodable only after a nonlinearity (the FFN's activation or attention's content-dependent weights).

\textbf{Main tasks.} The main analysis uses three constructed tasks: volatility, diff energy, and spectral entropy. Only constructed tasks can show which component constructs, since innate ones are linearly decodable at the embedding and leave nothing to attribute. Of the other constructed tasks, slope\_magnitude and change\_point only take the absolute value of a linear quantity the embedding already exposes, so we report them as boundary tasks, though on ETTm1 both also gain mainly at the Layer-1 FFN (Appendix~\ref{app:full_ettm1}). The main tasks lack such a linear part and view temporal complexity from statistical, differential, and spectral perspectives (results for all tasks in Appendix~\ref{app:full}).

\textbf{Supporting roles.} The other tasks either check the classification or ask a different question: trend\_direction shows what the embedding already exposes (\S\ref{sec:innate_constructed}), reg\_slope and reg\_variance repeat the innate-versus-constructed contrast with continuous labels (Appendix~\ref{app:reg_variance}), and the random-label control checks probe reliability. The inter-token tasks, reg\_next\_mean, and local\_global\_trend have labels that depend on other tokens, so they ask what reaches a token from elsewhere, not only what is constructed within it: routing between tokens (Appendix~\ref{app:intertoken}), what a token carries about the next segment, and whether its trend matches that of the whole window, respectively. Finally, dominant\_period, meant to capture the strongest frequency band, is degenerate: one band wins so often that its embedding accuracy matches the majority-class rate on every dataset (Appendix~\ref{app:embed_table}).

\textbf{Other temporal properties.} Forecasting work often stresses four further properties: trend shifts, seasonality, multi-scale structure, and long-range dependence. Trend shifts are probed by change\_point and local\_global\_trend. For the other three, at the granularity a 16-point token permits, we add three targets on PatchTST (results in \S\ref{sec:ffn_dominance}). Seasonal (spectral) concentration and a multi-scale ratio are both constructed ratios of second-order statistics; long-range consistency compares a token's slope with that of the preceding eight tokens and so also needs routing.

\subsection{Architecture-Specific Adaptations}
\label{sec:adaptations}

For patch-based models (PatchTST, CATS), task labels are computed on $P$ time points per token after the model's input normalization (for a CATS query, on the future patch it forecasts). For iTransformer, each token represents an entire variate's normalized lookback series, so task labels are computed per variate, before this normalization, which would make volatility nearly constant. As a token lacks its own series' mean and standard deviation, part of attention's gain can come from other tokens (Appendix~\ref{app:itrans_full}). For PatchTST ($P{=}1$) and Informer, we apply a mean adapter: 16 consecutive representations (stride 8) are mean-pooled; Informer's pooled vector mixes all variates yet is probed against per-variate labels (Appendix~\ref{app:informer}). Absolute accuracies are comparable only within the PatchTST family, which shares the same task granularity (16-point patches); for iTransformer, we compare the ratio of the two deltas.

\subsection{Probe Training Protocol}
\label{sec:protocol}

All probes are single linear layers trained with Adam ($\text{lr}=10^{-3}$), batch size 32{,}768, for 30k gradient steps on an 80/20 train/test split (Appendix~\ref{app:setup_details} describes how samples are drawn, labeled, and split). We evaluate every 500 steps and report the best test accuracy over the entire training trajectory. All experiments use 3 model seeds $\times$ 3 probe seeds = 9 runs. While probing accuracy rises slightly beyond 30k (e.g., $+1.6$ percentage points (pp) at \texttt{1b\_ffn} by 80k on volatility), the component that leads on each task is the same from 5k steps onward on all 4 datasets tested (Appendix~\ref{app:convergence}).

\section{Experimental Setup}
\label{sec:setup}

\label{sec:models_ablations}
We arrange 5 models (3 architectures and 2 PatchTST variants) by tokenization and attention type (Table~\ref{tab:model_matrix} in Appendix~\ref{app:setup_details}): at the patch level, PatchTST~\citep{nie2023time} (self-attention), CATS~\citep{kim2024cats} (cross-attention), and PatchTST-NoAttn (no attention); at the variate level, iTransformer~\citep{liu2024itransformer}; and at the point level, PatchTST ($P{=}1$), with single-scalar tokens. Informer~\citep{zhou2021informer}, whose time-step tokens hold all variates, is an auxiliary case.

\textbf{PatchTST-NoAttn} sets the self-attention output to zero, reducing the sub-layer to its residual identity with all else unchanged. We do so either in a trained checkpoint with the FFN weights fixed (\emph{ablated}; probing differences then reflect only the removal of attention) or from initialization on all 8 datasets (\emph{from scratch}, as in RAM~\citep{guo2024ram}, which found that this costs little in forecasting).

\textbf{PatchTST ($P{=}1$)} changes only the patch length $P$ and the stride to 1, so the model differs from PatchTST only in tokenization; its 16-point tasks are read from 16 consecutive tokens through an adapter (\S\ref{sec:adaptations}). PatchTST is also trained with $P \in \{2, 4, 8, 32\}$ (stride $P/2$) and with non-overlapping patches ($P = \text{stride} = 16$) on 4 datasets (ETTh1, ETTm1, Weather, Electricity). \textbf{Informer}, probed on the same 4 datasets, is not a point-level model: its tokens are time steps embedded by a three-step convolution over all variates (Appendix~\ref{app:informer}).

PatchTST and its variants, iTransformer, and CATS use per-dataset hyperparameters from the original papers and official scripts and, for datasets they do not cover, those of the most similar dataset in scale (Appendix~\ref{app:setup_details}). The input length $\text{seq\_len}$ is 336 for PatchTST and its variants and 96 for the other models. Inputs are instance-normalized with RevIN without affine parameters~\citep{kim2022reversible} in PatchTST and its variants and with built-in series normalization in iTransformer and CATS. All models use $\text{pred\_len}=96$ in the main analysis; PatchTST results for $\text{pred\_len} \in \{192, 336, 720\}$ on the same 4 datasets are in Appendix~\ref{app:predlen}. Appendices~\ref{app:patching} and~\ref{app:knockout} list the models, datasets, and seeds used for Layer-1 activation patching and component knockout.

\textbf{Datasets.}
\label{sec:datasets}
We use 8 long-term forecasting benchmarks with standard splits (Table~\ref{tab:datasets} in Appendix~\ref{app:setup_details}): ETTm1, ETTm2, ETTh1, and ETTh2~\citep{zhou2021informer}, Weather~\citep{wu2021autoformer}, and Electricity, Exchange, and Traffic~\citep{lai2018modeling}. Solar-Energy is excluded from the main analysis for data quality reasons (Appendix~\ref{app:dataquality}). Probe samples come from each dataset's test split.

\section{Results}
\label{sec:results_start}

Unless stated otherwise, we report Layer-1 deltas and their sum (the total) in pp of probing accuracy averaged over the three constructed tasks of \S\ref{sec:tasks}, as 8-dataset means of 9-run means. Each of these tasks labels a token by the quartile of one statistic of its time points (16 in PatchTST; chance $25\%$): \emph{volatility} is their variance, \emph{diff energy} the mean squared step between consecutive points (larger for rougher segments), and \emph{spectral entropy} how evenly their power spreads over frequencies (low for a clean oscillation, high for noise). The other tasks play the supporting roles of \S\ref{sec:tasks} and are analyzed in Appendices~\ref{app:full},~\ref{app:intertoken}, and~\ref{app:selectivity}. Deltas locate where a property becomes linearly decodable and do not measure forecasting quality; the forecasting-side interventions establish component-level reliance and sufficiency, not that a model builds and uses the individual probed features (\S\ref{sec:limitations}).

We separate innate from constructed properties at the embedding (\S\ref{sec:innate_constructed}), show in probing that the FFN constructs in PatchTST and test with activation patching whether attention contributes (\S\ref{sec:ffn_dominance}), remove attention from trained checkpoints and from training (\S\ref{sec:noattn}), reduce tokens to single scalars and vary patch length and stride (\S\ref{sec:pointwise}), and compare architectures in probing and with component knockout (\S\ref{sec:role_separation}). Datasets and exceptions are given with each claim.

\subsection{Innate vs. Constructed: Two Classes of Temporal Information}
\label{sec:innate_constructed}

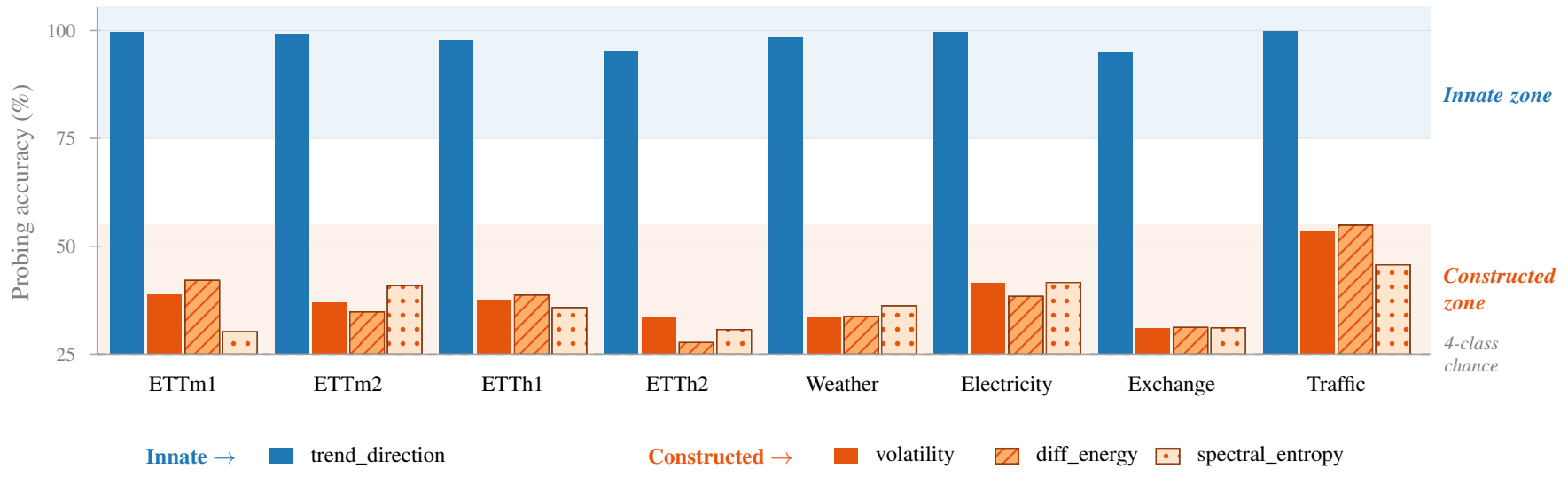
\begin{figure}[!h]
\centering
\resizebox{0.85\linewidth}{!}{%
\begin{tikzpicture}[
  font=\small,
  >=Stealth,
  axis-line/.style={line width=0.5pt, neuripsgray!70},
  grid-line/.style={line width=0.2pt, neuripsgray!25}
]

\def\ymin{25}
\def\ymax{100}
\def\yscale{0.055}
\def\barw{0.44}
\def\groupw{2.1}
\def\bargap{0.04}

\draw[axis-line] (0, 0) -- (0, {(\ymax-\ymin)*\yscale + 0.3});
\foreach \y in {25, 50, 75, 100} {
  \pgfmathsetmacro{\yp}{(\y-\ymin)*\yscale}
  \draw[axis-line] (-0.1, \yp) -- (0.05, \yp);
  \node[font=\scriptsize, neuripsgray, anchor=east] at (-0.15, \yp) {\y};
}
\node[font=\small, neuripsgray, rotate=90] at (-0.95, {(\ymax-\ymin)*\yscale/2})
  {Probing accuracy (\%)};

\foreach \y in {50, 75, 100} {
  \pgfmathsetmacro{\yp}{(\y-\ymin)*\yscale}
  \draw[grid-line] (0, \yp) -- (17.0, \yp);
}

\draw[dashed, neuripsgray, line width=0.5pt] (0, 0) -- (17.0, 0);
\node[font=\scriptsize\itshape, neuripsgray, anchor=west, align=left] at (17.05, 0) {4-class\\chance};

\def\datasets{%
  ETTm1/99.6/38.9/42.1/30.2,
  ETTm2/99.3/36.9/34.8/40.9,
  ETTh1/97.9/37.6/38.7/35.8,
  ETTh2/95.3/33.7/27.7/30.7,
  Weather/98.5/33.6/33.8/36.2,
  Electricity/99.6/41.6/38.4/41.6,
  Exchange/94.9/31.1/31.2/31.1,
  Traffic/99.9/53.7/54.9/45.7%
}

\foreach \name/\trend/\vol/\de/\se [count=\i from 0] in \datasets {
  \pgfmathsetmacro{\xc}{1.1 + \i * \groupw}
  \pgfmathsetmacro{\xstart}{\xc - 4*\barw/2 - 1.5*\bargap}

  \pgfmathsetmacro{\x}{\xstart}
  \pgfmathsetmacro{\h}{(\trend - \ymin)*\yscale}
  \fill[innate1] (\x, 0) rectangle ({\x+\barw}, \h);

  \pgfmathsetmacro{\x}{\xstart + (\barw + \bargap)}
  \pgfmathsetmacro{\h}{(\vol - \ymin)*\yscale}
  \fill[cons1] (\x, 0) rectangle ({\x+\barw}, \h);

  \pgfmathsetmacro{\x}{\xstart + 2*(\barw + \bargap)}
  \pgfmathsetmacro{\h}{(\de - \ymin)*\yscale}
  \fill[cons2] (\x, 0) rectangle ({\x+\barw}, \h);
  \draw[pattern=north east lines, pattern color=cons1] (\x, 0) rectangle ({\x+\barw}, \h);
  \draw[cons1!80!black, line width=0.3pt] (\x, 0) rectangle ({\x+\barw}, \h);

  \pgfmathsetmacro{\x}{\xstart + 3*(\barw + \bargap)}
  \pgfmathsetmacro{\h}{(\se - \ymin)*\yscale}
  \fill[cons3] (\x, 0) rectangle ({\x+\barw}, \h);
  \draw[pattern=dots, pattern color=cons1] (\x, 0) rectangle ({\x+\barw}, \h);
  \draw[cons1!80!black, line width=0.3pt] (\x, 0) rectangle ({\x+\barw}, \h);

  \node[font=\footnotesize, anchor=north] at (\xc, -0.15) {\name};
}

\draw[axis-line] (0, 0) -- (17.0, 0);

\begin{scope}[on background layer]
  \fill[innate1!8] (0, {(75-\ymin)*\yscale}) rectangle (17.0, {(\ymax-\ymin)*\yscale + 0.3});
  \fill[cons1!8] (0, 0) rectangle (17.0, {(55-\ymin)*\yscale});
\end{scope}

\node[font=\footnotesize\bfseries\itshape, text=innate1, anchor=west, align=left, text width=1.5cm]
  at (17.05, {(85-\ymin)*\yscale}) {Innate zone};
\node[font=\footnotesize\bfseries\itshape, text=cons1, anchor=west, align=left, text width=1.7cm]
  at (17.05, {(40-\ymin)*\yscale}) {Constructed zone};

\def\legy{-1.3}
\def\legysub{-1.3}

\node[font=\footnotesize\bfseries, text=innate1, anchor=west] at (0.5, \legy) {Innate $\rightarrow$};
\fill[innate1] (2.2, \legy-0.1) rectangle (2.5, \legy+0.1);
\node[font=\footnotesize, anchor=west] at (2.6, \legy) {trend\_direction};

\begin{scope}[xshift=6.4cm]
\node[font=\footnotesize\bfseries, text=cons1, anchor=west] at (0.5, \legysub) {Constructed $\rightarrow$};
\fill[cons1] (3.0, \legysub-0.1) rectangle (3.3, \legysub+0.1);
\node[font=\footnotesize, anchor=west] at (3.4, \legysub) {volatility};
\fill[cons2] (5.05, \legysub-0.1) rectangle (5.35, \legysub+0.1);
\draw[pattern=north east lines, pattern color=cons1] (5.05, \legysub-0.1) rectangle (5.35, \legysub+0.1);
\draw[cons1!80!black, line width=0.3pt] (5.05, \legysub-0.1) rectangle (5.35, \legysub+0.1);
\node[font=\footnotesize, anchor=west] at (5.45, \legysub) {diff\_energy};
\fill[cons3] (7.1, \legysub-0.1) rectangle (7.4, \legysub+0.1);
\draw[pattern=dots, pattern color=cons1] (7.1, \legysub-0.1) rectangle (7.4, \legysub+0.1);
\draw[cons1!80!black, line width=0.3pt] (7.1, \legysub-0.1) rectangle (7.4, \legysub+0.1);
\node[font=\footnotesize, anchor=west] at (7.5, \legysub) {spectral\_entropy};
\end{scope}

\end{tikzpicture}
}
\caption{\textbf{Embedding-layer probing reveals two classes of temporal information.} PatchTST's patch embedding preserves the \emph{innate} trend direction ($94.9$ to $99.9\%$), whereas \emph{constructed} properties (volatility, diff energy, spectral entropy) reach only $27.7$ to $54.9\%$ (Appendix~\ref{app:embed_table}).}
\label{fig:embed_dichotomy}
\end{figure}

PatchTST's embedding shows the dichotomy of \S\ref{sec:tasks} (Figure~\ref{fig:embed_dichotomy}): the linear patch projection keeps the \textbf{innate} trend direction at $94.9$ to $99.9\%$ accuracy, whereas the \textbf{constructed} properties stay at $27.7$ to $54.9\%$ (chance $25\%$) until a nonlinearity acts on them (\S\ref{sec:ffn_dominance}). Regression shows the same split: reg\_slope reaches $R^2 \geq 0.98$ on every dataset, while reg\_variance stays at most $0.15$ except on Traffic ($0.50$; Appendix~\ref{app:reg_variance}).

\textbf{Control validation.} As a pipeline sanity check, a random-label probe stays at chance ($\leq 26.4\%$) at every layer in all 4 models of Appendix~\ref{app:selectivity}, against PatchTST's Layer-1 constructed-task gain of $+26.7$\,pp (Cohen's $d_z = +1.67$, $p < 10^{-6}$, paired $t$-test pooled over the 24 dataset-task pairs).

\subsection{The FFN Dominates Feature Construction in PatchTST}
\label{sec:ffn_dominance}

Layer 1 raises the linear decodability of the constructed properties (volatility, diff energy, spectral entropy), and Table~\ref{tab:patchtst_delta} splits this gain between its two sub-layers. The FFN delta exceeds the attention delta on all 8 datasets: on 6 by at least 2$\times$ or with a negligible attention delta, and on ETTm1 and Weather by 1.9$\times$ and 1.3$\times$ (the 3.2$\times$ in Table~\ref{tab:patchtst_delta} is the ratio of the 8-dataset means). On ETTh1 the construction itself is weak (Attn $+0.0$, FFN $+3.8$). Per task, attention leads only on diff energy for ETTm1 and Weather (Appendix~\ref{app:full_ettm1}). Across horizons, $\Delta_{\text{FFN}} \geq \Delta_{\text{attn}}$ holds in all 16 pairs of dataset and horizon (Appendix~\ref{app:predlen}; Weather at $\text{pred\_len}=720$ is a near tie). The three targets added in \S\ref{sec:tasks} agree: their FFN deltas exceed their attention deltas on 6 or 7 of 8 datasets ($+12.0$, $+9.5$, and $+5.8$ against $+4.8$, $+6.4$, and $+2.1$ on average). Weather is the exception on the two ratio targets, and Traffic and ETTh2 (a tie) on long-range consistency (Appendix~\ref{app:new_targets}). A larger FFN delta on long-range consistency does not imply routing by the FFN.

\begin{table}[!htbp]
\caption{PatchTST Layer-1 constructed-task deltas (volatility, diff energy, spectral entropy; 9-run means). $\pm$ is the root-sum-square of the 9-run standard deviations at the two probing points of each delta, averaged over the three tasks; the mean averages the per-dataset values and their $\pm$ before rounding. ``alone'': negligible attention delta.}
\label{tab:patchtst_delta}
\centering
\footnotesize
\setlength{\tabcolsep}{1.5pt}
\begin{tabular}{lccccccccc}
\toprule
 & ETTm1 & ETTm2 & ETTh1 & ETTh2 & Weather & Electricity & Exchange & Traffic & \textbf{Mean} \\
\midrule
Attn $\Delta$ & $+13.5$\stdv{1.6} & $+0.2$\stdv{0.5} & $+0.0$\stdv{0.4} & $-0.6$\stdv{0.9} & $+16.7$\stdv{0.7} & $+11.3$\stdv{0.5} & $+0.1$\stdv{0.6} & $+9.5$\stdv{0.4} & $+6.4$\stdv{0.7} \\
FFN $\Delta$ & $\mathbf{+26.1}$\stdv{1.7} & $\mathbf{+43.0}$\stdv{0.7} & $\mathbf{+3.8}$\stdv{1.8} & $\mathbf{+11.1}$\stdv{2.2} & $\mathbf{+21.5}$\stdv{1.8} & $\mathbf{+23.7}$\stdv{0.8} & $\mathbf{+12.8}$\stdv{2.4} & $\mathbf{+20.8}$\stdv{0.6} & $\mathbf{+20.4}$\stdv{1.5} \\
Ratio & 1.9$\times$ & alone & alone & alone & 1.3$\times$ & 2.1$\times$ & alone & 2.2$\times$ & \textbf{3.2$\times$} \\
\bottomrule
\end{tabular}
\end{table}

\textbf{Why the FFN.} This allocation fits what each sub-layer computes on a patch token; the account is an argument, not a derivation (\S\ref{sec:limitations}). The FFN, a GELU MLP applied to each token's own representation, can compute nonlinear statistics of the 16-point patch such as its variance, whereas attention mixes linearly projected tokens with content-dependent weights and, for a fixed attention pattern, is linear in token content~\citep{elhage2021mathematical}.

\textbf{Forecasting side (activation patching).} We restore clean Layer-1 component outputs into a run whose input has its 16-step chunks shuffled in time and measure the forecasting MSE recovered~\citep{meng2022locating}, a denoising test of whether a component's clean output suffices to restore the forecast~\citep{heimersheim2024use} (Appendix~\ref{app:patching}). The test agrees with probing on whether attention contributes at all: restoring attention recovers at most $0.022$ MSE on the 4 datasets where probing puts $\Delta_{\text{attn}}$ near zero (ETTh1, ETTh2, ETTm2, Exchange) and $0.036$ to $0.229$ on the other~4. Restoring the FFN recovers more on 7 of 8 datasets (all but Weather, where it leads only from Layer~2 on; Appendix~\ref{app:patching_multilayer}), but the instrument favors it, since a restored FFN output is computed from the clean attention output as well; only attention's recovery is informative.

\subsection{The FFN Constructs Without Attention}
\label{sec:noattn}

We next ask whether the FFN needs attention for this construction, using the 2 NoAttn models of \S\ref{sec:models_ablations}: attention is zeroed either in a trained checkpoint whose FFN weights are kept (\emph{ablated}) or throughout training (\emph{from scratch}). Table~\ref{tab:noattn_r2} summarizes both; per-dataset values are in Appendices~\ref{app:noattn_full} and~\ref{app:noattn_mse}.

\begin{table}[!htbp]
\caption{Removing attention from PatchTST: Layer-1 constructed-task deltas (pp) and forecasting MSE, 8-dataset means. Total $= \Delta_{\text{attn}} + \Delta_{\text{FFN}}$ before rounding; MSE change is relative to PatchTST's mean MSE. $^\dagger$Zero by construction (\texttt{1a\_attn} is then an affine map of the embedding; per-dataset $|\Delta| \leq 0.7$): a sanity check of the probing pipeline, not a result.}
\label{tab:noattn_r2}
\centering
\footnotesize
\begin{tabular}{lccccc}
\toprule
Model & Attn $\Delta$ & FFN $\Delta$ & Total & MSE & MSE change \\
\midrule
PatchTST & $+6.4$ & $+20.4$ & $+26.7$ & $0.2305$ & \\
NoAttn, ablated & $0^\dagger$ & $+27.4$ & $+27.4$ & $0.2636$ & $+14.4\%$ \\
NoAttn, from scratch & $0^\dagger$ & $+28.8$ & $+28.8$ & $0.2352$ & $+2.0\%$ \\
\bottomrule
\end{tabular}
\end{table}

Both removals leave the probing total nearly unchanged ($+26.7$ with attention, $+27.4$ ablated, $+28.8$ from scratch): without attention, the FFN alone makes the constructed properties as linearly decodable as attention and the FFN do together. In the ablated model, the same FFN takes over attention's share: its delta barely changes where attention added almost nothing and rises by about attention's former delta elsewhere, so no dataset's total moves by more than $3$\,pp. Trained from scratch, the FFN alone reaches a similar total on every dataset ($-0.8$ to $+5.7$\,pp relative to PatchTST), so the FFN does not need attention during training either.

\textbf{Reliance is not construction.} In forecasting, the two removals differ: ablation (attention knockout) raises MSE on all 8 datasets ($+14.4\%$ on the mean MSE), whereas the from-scratch model comes within $2.0\%$ of PatchTST (at most $5.4\%$ on any dataset). Because ablation leaves the probed construction intact, its cost shows that the frozen model relies on attention for something the suite does not probe: on ETTm2, attention adds almost nothing in probing ($\Delta_{\text{attn}} = +0.2$), yet ablating it costs $+11.4\%$ MSE. The cost is consistent with a train/inference shift, since the FFN and downstream layers were trained on inputs that included attention outputs (cf.~\citealp{li2024optimal}).

\textbf{Sufficiency, not necessity.} Retraining removes most of this cost, whether by removing the shift or by recovering attention's contribution in other ways. At the component level, then, the FFN suffices: the retrained network reaches the PatchTST probing total and forecasts within a few percent of PatchTST, with the FFN as its only nonlinear transformation of token content (RevIN acts before patching, and BatchNorm is affine at inference) and with tokens interacting only through the linear head. Whether the FFN is necessary (training without it) is not tested here.

\subsection{Under Single-Scalar Tokens, the FFN Adds Little}
\label{sec:pointwise}

We now keep PatchTST's architecture and change its tokens. At $P{=}1$ (\S\ref{sec:models_ablations}), each token is a linear projection of one scalar, and the 16-point tasks are read through the mean adapter of \S\ref{sec:adaptations}; per-dataset values are in Appendix~\ref{app:p1_full}. The FFN's construction then nearly vanishes: the 8-dataset deltas shift from (Attn $+6.4$, FFN $+20.4$) at $P{=}16$ to (Attn $+10.9$, FFN $+2.3$) at $P{=}1$, a fall of 89\% in the FFN delta, and attention leads on 6 of 8 datasets, the exceptions being ETTh2 (FFN slightly ahead) and Exchange (flat). Deeper layers do not recover it: summed over the three layers, the FFN delta is $+1.2$ at $P{=}1$ against $+21.0$ at $P{=}16$. The total also halves on average ($+26.7$ to $+13.2$\,pp), though it grows on ETTh1 and ETTh2.

\textbf{Patch-length sweep.} One more point per token restores the FFN's construction. The sweep of \S\ref{sec:models_ablations} computes the tasks on each token's own $P$ points, so it asks whether the FFN constructs these statistics whenever a token holds the points they depend on, not how a fixed task changes with $P$ (at $P{=}2$, volatility and diff energy both become quartiles of the boundary-type $|x_2-x_1|$; Appendix~\ref{app:patch_sweep}). At $P{=}2$ the FFN delta is already $+13.0$ to $+31.4$, against $-0.9$ to $+3.4$ for the 16-point tasks at $P{=}1$, and for $P \geq 2$ it exceeds the attention delta in 18 of the 20 pairs of dataset and patch length (including $P{=}16$); the exceptions are Weather at $P{=}2$ (a near tie) and at $P{=}8$. The strength varies by dataset rather than growing with $P$ (on ETTh1 the FFN delta is only $+3.8$ at $P{=}16$ and $+0.5$ at $P{=}32$), and non-overlapping patches ($P = \text{stride} = 16$) reproduce the $P{=}16$ deltas within $3$\,pp.

\textbf{What the tokenization explains.} The FFN computes statistics of its own token (\S\ref{sec:ffn_dominance}), and a token needs at least two points to have any, so at $P{=}1$ only attention can bring several time points into a token. It can do so: a sufficiently wide multi-head attention layer can in principle stack lagged values into each token~\citep[Lemma~3.2]{wu2025transformers}, and the FFN could then compute statistics of what attention gathered. The argument thus does not predict which component leads at $P{=}1$. Empirically, in PatchTST, attention leads and the FFN adds little, and this is not an artifact of averaging: concatenating the 16 token representations instead of averaging them gives the same pattern on the 4 sweep datasets (Appendix~\ref{app:p1_full}).

\textbf{Forecasting side (knockout and activation patching).} The shift toward attention is representational: the frozen $P{=}1$ model still relies more on the FFN. Knocking out the FFN costs more than knocking out attention on all 4 datasets tested (ETTh1, ETTm1, Weather, Exchange; e.g., ETTh1 $+41.4\%$ vs.\ $+5.8\%$), as it does at $P{=}16$ (Appendix~\ref{app:knockout}). Layer-1 activation patching on ETTh1 and ETTm1 also restores more through the FFN, although attention's share rises relative to $P{=}16$ (Appendix~\ref{app:patching_p1}). Activation patching is biased toward the FFN (\S\ref{sec:ffn_dominance}), and knockout measures reliance, not construction (\S\ref{sec:noattn}), so neither instrument can confirm or refute the shift seen in probing.

\textbf{Informer} (auxiliary) is flat (Attn $+0.85$, FFN $+1.47$), but its tokens are not single scalars (\S\ref{sec:models_ablations}), so the result is confounded (Appendix~\ref{app:informer}).

\subsection{Component Roles Across Tokenizations and Architectures}
\label{sec:role_separation}

\begin{figure}[!htbp]
\centering
\resizebox{0.95\linewidth}{!}{%
\begin{tikzpicture}[
  font=\small,
  >=Stealth,
  attn-bar/.style={
    fill=neuripsorange!28,
    draw=neuripsorange,
    line width=0.6pt,
    postaction={pattern=north east lines, pattern color=neuripsorange!45}
  },
  ffn-bar/.style={
    fill=neuripsblue!28,
    draw=neuripsblue,
    line width=0.6pt,
    postaction={pattern=dots, pattern color=neuripsblue!55}
  },
  arch-name/.style={
    font=\small\bfseries, anchor=west
  },
  arch-token/.style={
    font=\scriptsize\itshape, text=neuripsgray, anchor=west
  },
  value-label/.style={
    font=\footnotesize, align=center
  },
  dom-label/.style={
    font=\footnotesize\bfseries, anchor=west
  },
  tok-tag/.style={
    rounded corners=2pt,
    font=\scriptsize\bfseries,
    inner sep=2.5pt,
    anchor=east,
    minimum width=1.0cm
  }
]

\def\barscale{0.07}
\def\centerx{0}
\def\barheight{0.42cm}
\def\labelright{2.6}
\def\namex{-6.6}

\node[font=\small\bfseries\itshape, text=neuripsorange, anchor=east] at (-0.3, 1.95)
  {$\leftarrow$ Attention / Cross-attention};
\node[font=\small\bfseries\itshape, text=neuripsblue, anchor=west] at (0.3, 1.95)
  {FFN $\rightarrow$};

\draw[neuripsgray, line width=0.8pt] (\centerx, 1.65) -- (\centerx, -1.30);

\def\rowA{1.4}
\def\rowB{0.60}
\def\rowC{-0.20}
\def\rowD{-1.00}

\node[arch-name] (n1) at (\namex, \rowA + 0.12) {PatchTST};
\node[tok-tag, fill=tokpatch!15, draw=tokpatch!50, text=tokpatch!90!black]
  at ([xshift=-0.15cm]n1.west) {PATCH};
\node[arch-token] at (\namex, \rowA - 0.22) {Self-Attn};

\node[attn-bar, minimum width={6.4*\barscale*1cm}, minimum height=\barheight, anchor=east]
  (a1) at (\centerx - 0.05, \rowA - 0.05) {};
\node[value-label, anchor=east] at ([xshift=-2pt]a1.west) {+6.4};

\node[ffn-bar, minimum width={20.4*\barscale*1cm}, minimum height=\barheight, anchor=west]
  (b1) at (\centerx + 0.05, \rowA - 0.05) {};
\node[value-label, anchor=west] at ([xshift=2pt]b1.east) {+20.4};

\node[dom-label, text=neuripsblue] at (\labelright, \rowA - 0.05)
  {$\bigstar$ \textbf{FFN} dominates ($3.2\times$)};

\node[arch-name] (n2) at (\namex, \rowB + 0.12) {iTransformer};
\node[tok-tag, fill=tokvariate!15, draw=tokvariate!50, text=tokvariate!90!black]
  at ([xshift=-0.15cm]n2.west) {VARIATE};
\node[arch-token] at (\namex, \rowB - 0.22) {Self-Attn};

\node[attn-bar, minimum width={13.8*\barscale*1cm}, minimum height=\barheight, anchor=east]
  (a2) at (\centerx - 0.05, \rowB - 0.05) {};
\node[value-label, anchor=east] at ([xshift=-2pt]a2.west) {+13.8};

\node[ffn-bar, minimum width={5.5*\barscale*1cm}, minimum height=\barheight, anchor=west]
  (b2) at (\centerx + 0.05, \rowB - 0.05) {};
\node[value-label, anchor=west] at ([xshift=2pt]b2.east) {+5.5};

\node[dom-label, text=neuripsorange] at (\labelright, \rowB - 0.05)
  {$\bigstar$ \textbf{Attention} dominates ($2.5\times$)};

\node[arch-name] (n3) at (\namex, \rowC + 0.12) {CATS};
\node[tok-tag, fill=tokpatch!15, draw=tokpatch!50, text=tokpatch!90!black]
  at ([xshift=-0.15cm]n3.west) {PATCH};
\node[arch-token] at (\namex, \rowC - 0.22) {Cross-Attn};

\node[attn-bar, minimum width={44.2*\barscale*1cm}, minimum height=\barheight, anchor=east]
  (a3) at (\centerx - 0.05, \rowC - 0.05) {};
\node[value-label, anchor=east] at ([xshift=-2pt]a3.west) {44.2$^*$};

\node[ffn-bar, minimum width={2.5*\barscale*1cm}, minimum height=\barheight, anchor=west]
  (b3) at (\centerx + 0.05, \rowC - 0.05) {};
\node[value-label, anchor=west] at ([xshift=2pt]b3.east) {+2.5};

\node[dom-label, text=neuripsorange] at (\labelright, \rowC - 0.05)
  {$\bigstar$ \textbf{Cross-attn}: only input path};

\node[arch-name] (n4) at (\namex, \rowD + 0.12) {PatchTST ($P{=}1$)};
\node[tok-tag, fill=tokpoint!15, draw=tokpoint!50, text=tokpoint!90!black]
  at ([xshift=-0.15cm]n4.west) {POINT};
\node[arch-token] at (\namex, \rowD - 0.22) {Self-Attn};

\node[attn-bar, minimum width={10.9*\barscale*1cm}, minimum height=\barheight, anchor=east]
  (a4) at (\centerx - 0.05, \rowD - 0.05) {};
\node[value-label, anchor=east] at ([xshift=-2pt]a4.west) {+10.9};

\node[ffn-bar, minimum width={2.3*\barscale*1cm}, minimum height=\barheight, anchor=west]
  (b4) at (\centerx + 0.05, \rowD - 0.05) {};
\node[value-label, anchor=west] at ([xshift=2pt]b4.east) {+2.3};

\node[dom-label, text=neuripsgray] at (\labelright, \rowD - 0.05)
  {\textbf{FFN adds little}; attention leads};

\draw[neuripsgray!60, line width=0.4pt] (-3.6, -1.45) -- (2.0, -1.45);
\draw[neuripsgray, line width=0.5pt] (0, -1.40) -- (0, -1.50);
\node[font=\scriptsize, neuripsgray, anchor=north] at (0, -1.55) {0};

\foreach \x in {10, 20, 30, 40, 50} {
  \draw[neuripsgray!60, line width=0.3pt] ({-\x*\barscale}, -1.40) -- ({-\x*\barscale}, -1.50);
  \node[font=\scriptsize, neuripsgray, anchor=north] at ({-\x*\barscale}, -1.55) {\x};
}

\foreach \x in {10, 20} {
  \draw[neuripsgray!60, line width=0.3pt] ({\x*\barscale}, -1.40) -- ({\x*\barscale}, -1.50);
  \node[font=\scriptsize, neuripsgray, anchor=north] at ({\x*\barscale}, -1.55) {\x};
}

\node[font=\scriptsize, neuripsgray, anchor=north] at (-1.0, -1.90)
  {$\Delta$ (probing accuracy gain, pp)};

\end{tikzpicture}
}
\caption{\textbf{Component dominance varies with tokenization and architecture.} Layer-1 constructed-task deltas (volatility, diff energy, spectral entropy; 8-dataset means, 9 runs) for 4 settings, tagged by tokenization. $^*$CATS cross-attention is reported as absolute accuracy, not as a delta, because the learnable queries start without per-instance input information; its bar is thus not on the delta scale. Its labels are those of the future patches the queries forecast.}
\label{fig:component_dominance}
\end{figure}
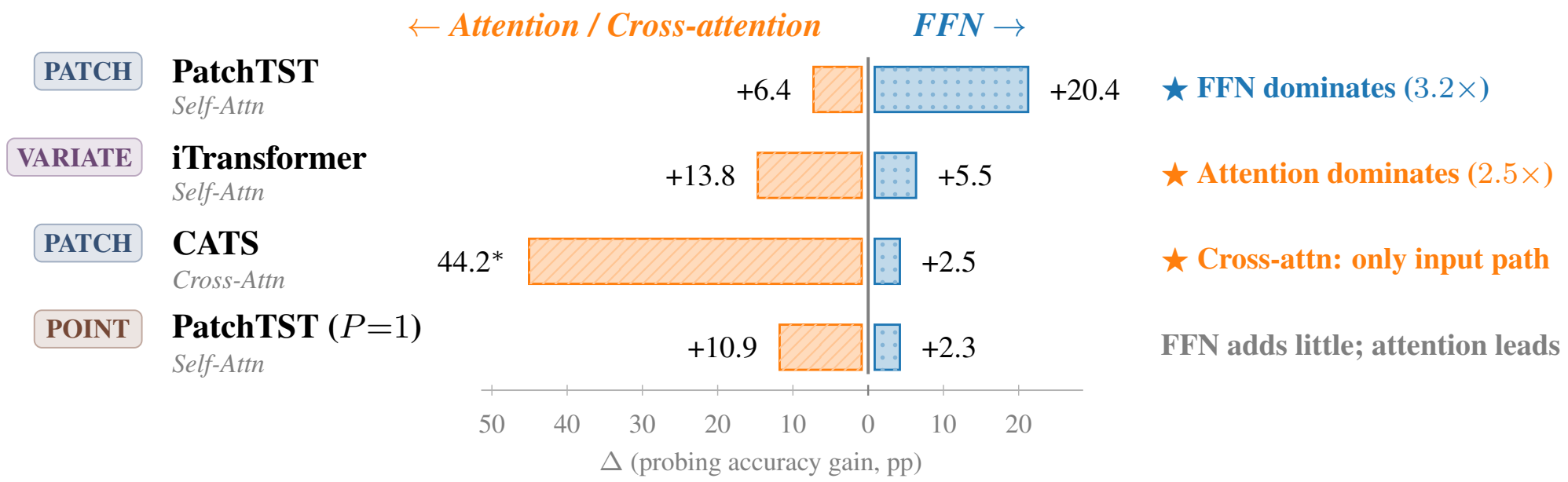

Figure~\ref{fig:component_dominance} compares PatchTST, iTransformer, CATS, and PatchTST ($P{=}1$). In iTransformer, whose tokens are whole variates, $\Delta_{\text{attn}} \geq \Delta_{\text{FFN}}$ on all 8 datasets (a $2.5\times$ ratio of the 8-dataset means), with near ties on Exchange ($+3.6$ vs.\ $+3.3$) and Weather ($+13.9$ vs.\ $+12.2$; Appendix~\ref{app:itrans_full}). In CATS, cross-attention is the only path from the input to the learnable queries, so its dominance is expected by construction (Appendix~\ref{app:cats_full}). Both are representational results: in iTransformer, attention's gain in linear decodability does not show that attention is necessary for forecasting.

\textbf{Forecasting side (component knockout).} Activation patching restores mostly through the FFN in iTransformer as well, which we attribute to the instrument (Appendix~\ref{app:patching_itrans}), so across architectures we compare the MSE cost of component knockout on all 8 datasets (Appendix~\ref{app:knockout}). The ordering across architectures matches probing: knocking out the FFN raises MSE by $35.1\%$ in PatchTST, $14.0\%$ in iTransformer, and $6.1\%$ in CATS, and the ratio of the mean attention cost to the mean FFN cost rises from $0.45$ to $1.32$ to $21.9$; both orderings hold when any one dataset is left out. Since knocking out cross-attention removes CATS's only input path, the informative comparison is PatchTST against iTransformer ($0.45$ against $1.32$). Per dataset the agreement is looser: PatchTST relies more on attention on 2 datasets and iTransformer more on the FFN on 4. Knockout measures what a frozen model relies on, not which component constructs (ETTm2 in \S\ref{sec:noattn}), and within PatchTST it does not track the shift at $P{=}1$ (\S\ref{sec:pointwise}).

\textbf{A common pattern.} Patch tokens with self-attention place construction in the FFN, variate tokens place it in attention, the cross-attention path carries it in CATS, and under single-scalar tokens, attention (the first operator to combine time points) leads while the FFN adds little. We read this as a regularity across the evaluated models rather than a mechanism. Within PatchTST, tokenization can be varied separately from the presence of a token mixer (the sub-layer that mixes tokens, here self-attention), and their effects on the total differ. Changing only the tokenization halves the total ($P{=}16 \to 1$, which also changes the read-out), whereas removing only the mixer leaves it nearly unchanged (\S\ref{sec:noattn}, \S\ref{sec:pointwise}); we did not vary the form of the mixer. Tokenization and what attention mixes cannot be separated, because switching from patch to variate tokens also switches attention from mixing time to mixing variates; nor is PatchTST versus iTransformer a controlled comparison (input length 336 vs.\ 96; channel independence vs.\ dependence,~\citealp{han2024capacity}; scope in \S\ref{sec:limitations}).

\section{A Representational Reading of Prior Results}
\label{sec:unified}

Our reading of the four findings of \S\ref{sec:intro} is qualitative, not a causal account of their MSE orderings. \textbf{(1) DLinear $>$ early transformers}~\citep{zeng2023transformers}. Like our $P{=}1$ tokens, their tokens cover one time step each, and at $P{=}1$ the FFN adds little (\S\ref{sec:pointwise}). Yet their tokens also mix all variates and neighboring steps, as Informer's do (\S\ref{sec:models_ablations}), so the $P{=}1$ result applies to them at most partially. \textbf{(2) PatchTST $>$ DLinear}~\citep{nie2023time}. A linear map of the raw window, such as DLinear, reads only linear functionals of it, whereas patching gives multi-point tokens to PatchTST's FFN, which makes within-patch statistics linearly decodable (\S\ref{sec:ffn_dominance}). \textbf{(3) Self-attention-free patch models}~\citep[TSMixer, PatchMLP;][]{ekambaram2023tsmixer,tang2025patchmlp}. At $P{=}16$, attention is not needed for the probed construction, and retraining without it costs little (\S\ref{sec:noattn}), a component-level result consistent with the forecasting accuracy of these models. Probing one such model, ModernTCN~\citep{luo2024moderntcn}, gives the same allocation: summed over its three blocks on ETTm1, the ConvFFNs, not the convolution mixer, carry the construction (Appendix~\ref{app:moderntcn_breakdown}). \textbf{(4) Cross-attention alone}~\citep[CATS;][]{kim2024cats}. CATS's learnable queries start without per-instance input information, so cross-attention, their only path from the input, supplies it by construction, and the FFN adds little (\S\ref{sec:role_separation}). Unlike the self-attention dropped in (3), which mixes tokens that already hold the input, this cross-attention is the input path itself. The same name, ``attention,'' thus covers different information flows: a removable mixer under patch tokens (3), the only input path in CATS (4), and, in probing, the main constructing component in iTransformer, whose tokens are whole variates (\S\ref{sec:role_separation}).

\section{Design Implications}
\label{sec:design}

Our results suggest three practices.

\textbf{(i) Diagnose before re-engineering.} Measure the Layer-1 attention and FFN deltas on the constructed tasks with a random-label control; Layer 1 suffices in PatchTST, whose deeper layers add little to the FFN delta (Table~\ref{tab:p1_depth}). A large FFN delta indicates that the tokens already hold the points the FFN needs, so attention is a candidate for cheaper mixing. A small one calls for trying another patch length on the same data: if construction appears there, the tokenization is the bottleneck.

\textbf{(ii) Choose the token before the mixer.} In PatchTST, shrinking tokens to single scalars halves the probing total, whereas removing the mixer leaves it nearly unchanged (\S\ref{sec:role_separation}). One extra point per token restores the FFN's construction, and longer patches need not add to it (\S\ref{sec:pointwise}). Under variate tokens, attention leads (iTransformer, \S\ref{sec:role_separation}), so the same mixer plays opposite roles.

\textbf{(iii) Under patch tokenization, self-attention is replaceable.} At $P{=}16$ even the lightest replacement, no mixer at all, suffices: PatchTST trained without attention keeps its probing total, with the FFN taking over attention's share, and stays within $2.0\%$ of the full model in mean MSE (Table~\ref{tab:noattn_r2}). This gives a component-level basis for the forecasting accuracy of TSMixer and PatchMLP (\S\ref{sec:unified}) and suggests spending the design budget on the tokenization and the FFN rather than on attention variants. For variate tokens we only predict the opposite: since attention leads in probing (\S\ref{sec:role_separation}), the same replacement would remove the constructing component rather than a redundant mixer.

\section{Limitations and Future Directions}
\label{sec:limitations}

\textbf{Reach of the evidence.} A delta locates where a property becomes linearly decodable, not whether the forecaster uses it~\citep{elazar2021amnesic}. In PatchTST at $P{=}16$, Layer-1 activation patching, training without attention, and knockout establish component-level reliance and sufficiency on the forecasting side (\S\ref{sec:ffn_dominance}, \S\ref{sec:noattn}, Appendix~\ref{app:knockout}). At $P{=}1$, activation patching and knockout point to the FFN but can neither confirm nor refute the shift seen in probing (\S\ref{sec:pointwise}). Across architectures, knockout supports only the ordering, so the construction claims for iTransformer and CATS stay representational (\S\ref{sec:role_separation}); Informer is only probed (Appendix~\ref{app:informer}). Below the component level, whether an individual probed property is used remains open: erasing one with iterative nullspace projection (INLP;~\citealp{ravfogel2020null}) removed 33 to 94\% of the 128 dimensions and was inconclusive against a random-label control (Appendix~\ref{app:erasure}).

\textbf{Scope.} Within PatchTST at $P{\geq}2$, the strength of construction varies with dataset and patch length (ETTh1 is weak at $P{=}16$ and $32$), but the allocation does not: it holds across patch lengths, strides, horizons, training without attention, both per-dataset widths ($d_{\text{model}}$ of 16 and 128), and three added targets, with Weather the recurring exception. Single-scalar and variate tokens lie outside this regime. These checks narrow but do not remove the generalization concern: we did not vary lookback, depth, or the form of the mixer, and a sweep with 16-point tasks fixed across $P$ is future work.

\textbf{Task suite.} The tasks are hand-picked to make construction visible, and the main conclusions rest on three second-order statistics of a token's segment (16 points in PatchTST), so the allocation may differ for other properties. The suite omits higher-order moments, multi-resolution structure beyond a two-scale split, dependence beyond eight tokens, and cross-token multi-scale structure.

\textbf{Where the framework applies.} The framework requires (i) an input embedding that is an affine map of each token's values, so that no construction happens inside it, and (ii) a linear head that reads the probed representations. Informer is a boundary case on both: its embedding mixes neighboring steps and variates, and its head reads decoder tokens. Any model meeting both is covered, including pretrained time series foundation models that do; we leave these to future work.

\textbf{No formal theory.} We establish the phenomenon and its boundary conditions experimentally but do not prove why they hold; a formal account is the next step. It should explain why the allocation shifts between single-scalar and multi-point tokens, and why, under patch tokens, the FFN alone reaches the full total when trained without attention. So far we have only an argument for patch tokens (\S\ref{sec:ffn_dominance}), a single-scalar analysis valid only before attention (\S\ref{sec:pointwise}), and nothing for variate tokens.

\section{Conclusion}

In time series transformers, the sub-layer that makes temporal properties linearly decodable is allocated by the interaction of tokenization and architecture: the FFN under patch tokens (PatchTST), attention under variate tokens (iTransformer, in probing), the cross-attention path in CATS, and attention under single-scalar tokens (in probing), where the FFN adds little. This reading accounts qualitatively for why patching helps and self-attention-free models compete, and suggests probing first, choosing the token before the mixer, and replacing self-attention only under patch tokens. Our forecasting-side interventions establish component-level reliance and sufficiency, not where individual features are causally constructed. Feature-level tests, pretrained foundation models, and a formal account are the next steps.

\clearpage
\begin{ack}
This work was supported by Inha University and by Institute of Information \& Communications Technology Planning \& Evaluation (IITP) grants funded by the Korea government (MSIT) under the Leading Generative AI Human Resources Development program (IITP-2026-RS-2024-00360227) and the Artificial Intelligence Innovation Human Resources Development program (IITP-2026-RS-2026-25548560).
\end{ack}

\bibliographystyle{plainnat}

\appendix
\clearpage

\section{Model Hyperparameters, Datasets, Probe Protocol, and Compute}
\label{app:setup_details}

\begin{table}[H]
\caption{Model coverage matrix: each cell indicates the model analyzed for the corresponding (tokenization, attention type) combination. PatchTST-NoAttn sets the self-attention output to zero, in a trained checkpoint (ablated) or from initialization (from scratch), so that the sub-layer is a residual identity and everything else, including BatchNorm, the FFN, and the per-dataset hyperparameters, is unchanged. Informer (4 datasets only) is a separate auxiliary case: its time-step tokens are embedded by a three-step convolution over all variates, so it is not a point-level model, and its probe is confounded (Appendix~\ref{app:informer}).}
\label{tab:model_matrix}
\centering
\small
\begin{tabular}{lccc}
\toprule
 & Self-Attention & Cross-Attention & No Attention \\
\midrule
Point ($P{=}1$) & PatchTST ($P{=}1$) & --- & --- \\
Time step (aux.) & Informer & --- & --- \\
Patch & PatchTST ($P{=}16$) & CATS & PatchTST-NoAttn ($P{=}16$) \\
Variate (whole series) & iTransformer & --- & --- \\
\bottomrule
\end{tabular}
\end{table}

\begin{table}[H]
\caption{Architecture-level configurations across the analyzed models (input length 336 for the PatchTST family and 96 for the other models).}
\label{tab:hparams_arch}
\centering
\small
\begin{tabular}{llccc}
\toprule
Model & Tokenization & $n_{\text{layers}}$ & patch\_len & stride \\
\midrule
PatchTST ($P{=}16$) & Patch & 3 & 16 & 8 \\
PatchTST ($P{=}1$) & Point & 3 & 1 & 1 \\
PatchTST-NoAttn & Patch (attn output $= 0$) & 3 & 16 & 8 \\
iTransformer & Variate (whole series) & varies (2--4) & --- & --- \\
CATS & Patch (cross-attn only) & 0 enc / 3 dec & 24 or 48 & 24 or 48 \\
Informer & Time step (conv $k{=}3$; ProbSparse) & 2 enc / 1 dec & --- & --- \\
\bottomrule
\end{tabular}
\end{table}

\paragraph{PatchTST (and variants).} PatchTST splits each univariate time series into 16-point patches with stride 8 and applies a vanilla 3-layer Transformer encoder per channel (channel-independent). Architectural width is determined by dataset scale (Table~\ref{tab:hparams_patchtst}); for datasets without original-paper specifications (Exchange, Solar), we use the configuration of the most similar dataset in scale: Exchange is grouped with ETTh1/ETTh2 (small), and Solar is grouped with Weather/Electricity/Traffic (large). PatchTST ($P{=}1$) sets \texttt{patch\_len} = \texttt{stride} = 1 with all other hyperparameters unchanged.

\begin{table}[H]
\caption{PatchTST per-dataset architectural configurations. Two width settings (small vs.\ large) are used.}
\label{tab:hparams_patchtst}
\centering
\small
\begin{tabular}{lccc}
\toprule
Datasets & $d_{\text{model}}$ & $d_{\text{ff}}$ & $n_{\text{heads}}$ \\
\midrule
ETTh1, ETTh2, Exchange & 16 & 128 & 4 \\
ETTm1, ETTm2, Weather, Electricity, Solar, Traffic & 128 & 256 & 16 \\
\bottomrule
\end{tabular}
\end{table}

\paragraph{iTransformer.} Each variate's whole lookback window becomes a single token; attention operates across variates. Width and depth follow the official per-dataset scripts (Table~\ref{tab:hparams_itrans}).

\begin{table}[H]
\caption{iTransformer per-dataset architectural configurations. $n_{\text{heads}}=8$ across all datasets; no patching.}
\label{tab:hparams_itrans}
\centering
\small
\begin{tabular}{lccc}
\toprule
Datasets & $d_{\text{model}}$ & $n_{\text{layers}}$ & $d_{\text{ff}}$ \\
\midrule
ETTh1 & 256 & 2 & 256 \\
ETTh2, ETTm1, ETTm2, Exchange & 128 & 2 & 128 \\
Solar & 512 & 2 & 512 \\
Weather, Electricity & 512 & 3 & 512 \\
Traffic & 512 & 4 & 512 \\
\bottomrule
\end{tabular}
\end{table}

\paragraph{CATS.} The encoder is removed ($e_{\text{layers}}=0$); only the cross-attention decoder is used. Width and depth are identical across datasets ($d_{\text{model}}=256$, $n_{\text{heads}}=32$, $d_{\text{layers}}=3$, $d_{\text{ff}}=512$), and \texttt{patch\_len}$=$\texttt{stride}$=48$ for ETTh1/ETTh2 and $24$ for the remaining datasets. Following the official scripts, the query-independence option (one set of queries shared by all variates; on for ETTm2, ETTh2, and Traffic), the batch size, the number of epochs, and the learning-rate schedule also vary by dataset; Exchange, which the scripts do not cover, uses the ETTm1 settings.

\paragraph{Informer.} A single architectural setting is used across all datasets: $d_{\text{model}}=512$, $n_{\text{heads}}=8$, $e_{\text{layers}}=2$, $d_{\text{layers}}=1$, $d_{\text{ff}}=2048$, ProbSparse factor $=3$; no patching. Training uses $\text{label\_len}=48$, batch size 128, learning rate $10^{-4}$, and 10 epochs (2 for Weather).

\begin{table}[H]
\caption{Dataset information. The splits follow the standard PatchTST protocol; Train/Val/Test sizes and Domain are as listed by \citet{liu2024itransformer}. Solar-Energy is excluded from the main analysis (Appendix~\ref{app:dataquality}).}\label{tab:datasets}
\centering
\small
\begin{tabular}{lcccl}
\toprule
Dataset & Variables & Train/Val/Test & Frequency & Domain \\
\midrule
ETTm1 & 7 & (34465, 11521, 11521) & 15 min & Electricity \\
ETTm2 & 7 & (34465, 11521, 11521) & 15 min & Electricity \\
ETTh1 & 7 & (8545, 2881, 2881) & 1 hour & Electricity \\
ETTh2 & 7 & (8545, 2881, 2881) & 1 hour & Electricity \\
Weather & 21 & (36792, 5271, 10540) & 10 min & Weather \\
Electricity & 321 & (18317, 2633, 5261) & 1 hour & Electricity \\
Exchange & 8 & (5120, 665, 1422) & Daily & Economy \\
Traffic & 862 & (12185, 1757, 3509) & 1 hour & Transportation \\
\midrule
Solar-Energy & 137 & (36601, 5161, 10417) & 10 min & Energy \\
\bottomrule
\end{tabular}
\end{table}

\paragraph{Forecaster training.} The models of Table~\ref{tab:model_matrix} are trained with each model's official code and scripts, which standardize each dataset with the statistics of its training split; datasets the scripts do not cover use the settings of the most similar dataset, as above, and Informer is trained with the implementation in the PatchTST code base. Averaged over the 3 seeds, the test MSE of PatchTST~\citep{nie2023time}, iTransformer~\citep{liu2024itransformer}, and CATS~\citep{kim2024cats} is within $3.7\%$ of the published result at the same input length on each of our 8 datasets that these papers report (PatchTST and CATS do not report Exchange).

\paragraph{Probe samples.} Probes are trained and evaluated on frozen representations of each dataset's test split, on which the forecasters are not trained. The split is read in order from its start, as the standard sliding windows that each start one time step after the previous one. Each window gives one token per variate and position (one per variate in iTransformer), and the tokens of all windows, variates, and positions form one pooled sample set per dataset, model seed, and probing point. Collection stops at a per-model budget. The PatchTST family uses 5{,}000 univariate input windows, that is, 210{,}000 tokens at $P{=}16$ and 205{,}000 mean-adapter samples at $P{=}1$. iTransformer uses whole batches of up to 32 windows until at least 50{,}000 variate tokens are collected (or the whole split, if it holds fewer), excluding the time-feature tokens that iTransformer appends. CATS uses full test batches, and Informer uses up to 250{,}000 pooled samples (Appendix~\ref{app:informer}).

\paragraph{Labels.} A token's labels are computed with the formulas of Table~\ref{tab:tasks} from its own time points, except that those of the inter-token tasks, reg\_next\_mean, and local\_global\_trend depend on its position or on other tokens of its window (\S\ref{sec:tasks}); \S\ref{sec:adaptations} gives, for each model, the time points from which a token's labels are computed. In spectral entropy, $q_k$ is the share of frequency bin $k$ in the power of the token's real FFT, zero frequency included. In PatchTST, dominant\_period's four bands are equal groups of consecutive bins starting at zero frequency (two bins each at $P{=}16$, with the Nyquist bin left out). The lowest band thus holds the zero-frequency bin, and it is the majority class on each of the 8 datasets ($84.93$ to $98.55\%$ of tokens, the majority column of Table~\ref{tab:embed}). Class boundaries (the quartiles, the 70th percentile for change\_point, and, for patch tokens, the median for adjacent\_similarity) are computed once over the pooled set, so all probing points and both parts of the probe's 80/20 split (below) share them.

\paragraph{Probe training.} One probe is trained for each dataset, model seed, probe seed, probing point, and task. It maps one token representation $h \in \mathbb{R}^{d_{\text{model}}}$ to $C$ class logits, $Wh + b$ with $W \in \mathbb{R}^{C \times d_{\text{model}}}$ (one output for regression; $16\,d_{\text{model}}$ inputs with the concatenation adapter of Appendix~\ref{app:p1_full}). The pooled set is split at random into 80\% for training and 20\% for evaluation. Each of the 30k Adam steps ($\text{lr}=10^{-3}$, no weight decay) uses 32{,}768 training samples (or as many as the training part holds, if fewer), drawn uniformly with replacement, and minimizes the cross-entropy, or the mean squared error for regression. Every 500 steps the probe is scored on the evaluation part (accuracy, or $R^2$ for regression), and the best score over training is reported. The probe seed sets the split, the initialization, and the batch draws.

\paragraph{Compute setup.} Experiments ran on NVIDIA A100 (40GB), A40, and RTX A6000 GPUs.

\section{Full Probing Results}
\label{app:full}

Appendix tables abbreviate trend\_direction as trend, dominant\_period as dom\_p, and volatility, diff\_energy, and spectral\_entropy as Vol, DE, and SE; the subscript $q4$ marks quartile labels. A probing accuracy's $\pm$ is the population standard deviation of its 9 runs, and the $\pm$ of a probing delta combines these standard deviations $s_a$ and $s_b$ of its two probing points as $\sqrt{s_a^2+s_b^2}$, as in Table~\ref{tab:patchtst_delta}; for a mean over tasks or datasets, these $\pm$ values are averaged.

\subsection{PatchTST Embed-Layer Probing (Numerical Values for Figure~\ref{fig:embed_dichotomy})}
\label{app:embed_table}

\begin{table}[H]
\caption{PatchTST embed-layer probing accuracy on intra-token tasks (9-run mean). trend is innate, and Vol, DE, and SE are constructed; dom\_p is degenerate (\S\ref{sec:tasks}): its probing accuracy is within $0.12$\,pp of the majority rate of its labels (shown to two decimals), the accuracy of always predicting the most common band. Figure~\ref{fig:embed_dichotomy} plots trend, Vol, DE, and SE.}
\label{tab:embed}
\centering
\footnotesize
\setlength{\tabcolsep}{5pt}
\begin{tabular}{lcccccc}
\toprule
 & & \multicolumn{2}{c}{dom\_p} & & & \\
\cmidrule(lr){3-4}
Dataset & trend & probe & majority & Vol & DE & SE \\
\midrule
ETTm1 & 99.6 & 96.2 & 96.25 & 38.9 & 42.1 & 30.2 \\
ETTm2 & 99.3 & 90.0 & 89.88 & 36.9 & 34.8 & 40.9 \\
ETTh1 & 97.9 & 92.3 & 92.36 & 37.6 & 38.7 & 35.8 \\
ETTh2 & 95.3 & 84.9 & 84.93 & 33.7 & 27.7 & 30.7 \\
Weather & 98.5 & 97.8 & 97.82 & 33.6 & 33.8 & 36.2 \\
Electricity & 99.6 & 94.2 & 94.26 & 41.6 & 38.4 & 41.6 \\
Exchange & 94.9 & 98.5 & 98.55 & 31.1 & 31.2 & 31.1 \\
Traffic & 99.9 & 95.0 & 94.91 & 53.7 & 54.9 & 45.7 \\
\bottomrule
\end{tabular}
\end{table}

\subsection{PatchTST Full Results}
\label{app:full_ettm1}

Table~\ref{tab:patchtst_pertask} gives the per-task Layer-1 deltas on all 8 datasets, and Table~\ref{tab:patchtst_full_ettm1} gives all tasks at every probing point on ETTm1 as an example; the released result files hold these values for every dataset.

\begin{table}[H]
\caption{PatchTST per-task Layer-1 delta (9-run mean $\pm$ std). Averaging the three tasks gives Table~\ref{tab:patchtst_delta}. The FFN delta exceeds the attention delta for every task on every dataset except diff\_energy on ETTm1 and Weather.}
\label{tab:patchtst_pertask}
\centering
\scriptsize
\setlength{\tabcolsep}{3pt}
\begin{tabular}{lcccccc}
\toprule
Dataset & Vol Attn $\Delta$ & Vol FFN $\Delta$ & DE Attn $\Delta$ & DE FFN $\Delta$ & SE Attn $\Delta$ & SE FFN $\Delta$ \\
\midrule
ETTm1 & $+10.3$\stdv{2.1} & $+34.5$\stdv{2.3} & $+13.5$\stdv{1.6} & $+10.6$\stdv{1.9} & $+16.8$\stdv{1.2} & $+33.2$\stdv{0.9} \\
ETTm2 & $-0.5$\stdv{0.4} & $+52.1$\stdv{0.6} & $+4.1$\stdv{0.4} & $+37.7$\stdv{0.6} & $-2.9$\stdv{0.8} & $+39.2$\stdv{0.8} \\
ETTh1 & $-0.1$\stdv{0.4} & $+2.9$\stdv{1.6} & $-0.2$\stdv{0.3} & $+3.1$\stdv{1.9} & $+0.3$\stdv{0.4} & $+5.4$\stdv{1.9} \\
ETTh2 & $-2.6$\stdv{1.1} & $+11.1$\stdv{3.3} & $+0.9$\stdv{1.1} & $+14.3$\stdv{1.3} & $-0.1$\stdv{0.6} & $+7.8$\stdv{1.9} \\
Weather & $+17.6$\stdv{0.4} & $+24.0$\stdv{1.8} & $+19.4$\stdv{0.6} & $+14.8$\stdv{1.5} & $+13.2$\stdv{1.0} & $+25.8$\stdv{2.1} \\
Electricity & $+12.5$\stdv{0.4} & $+30.5$\stdv{0.5} & $+11.7$\stdv{0.5} & $+26.4$\stdv{1.2} & $+9.8$\stdv{0.5} & $+14.3$\stdv{0.6} \\
Exchange & $+0.2$\stdv{0.4} & $+12.0$\stdv{2.3} & $+0.7$\stdv{0.5} & $+6.8$\stdv{1.5} & $-0.5$\stdv{0.8} & $+19.6$\stdv{3.5} \\
Traffic & $+11.5$\stdv{0.3} & $+26.5$\stdv{0.6} & $+10.9$\stdv{0.5} & $+20.0$\stdv{0.6} & $+6.2$\stdv{0.4} & $+15.8$\stdv{0.6} \\
\bottomrule
\end{tabular}
\end{table}

\begin{table}[H]
\caption{PatchTST full layer probing results on ETTm1 (9-run mean $\pm$ std). All 15 diagnostic tasks plus the random-label control. On the constructed tasks the Layer-1 FFN gives the largest rise for volatility and spectral\_entropy, and attention for diff\_energy; after Layer 1, no sub-layer changes their accuracy by more than $1.9$\,pp. Table~\ref{tab:p1_depth} gives, for all 8 datasets, the FFN deltas summed over the three layers.}
\label{tab:patchtst_full_ettm1}
\centering
\scriptsize
\setlength{\tabcolsep}{2pt}
\begin{tabular}{lccccccc}
\toprule
Task & embed & 1a\_attn & 1b\_ffn & 2a\_attn & 2b\_ffn & 3a\_attn & 3b\_ffn \\
\midrule
\multicolumn{8}{l}{\textit{Innate and degenerate (classification)}} \\
trend\_direction         & $99.6$\stdv{0.0} & $99.6$\stdv{0.0} & $99.4$\stdv{0.1} & $99.2$\stdv{0.2} & $98.9$\stdv{0.2} & $98.6$\stdv{0.1} & $98.5$\stdv{0.1} \\
dominant\_period         & $96.2$\stdv{0.1} & $96.2$\stdv{0.1} & $96.9$\stdv{0.2} & $96.8$\stdv{0.1} & $96.9$\stdv{0.1} & $96.8$\stdv{0.1} & $96.8$\stdv{0.1} \\
\midrule
\multicolumn{8}{l}{\textit{Constructed (classification)}} \\
volatility$_{q4}$        & $38.9$\stdv{0.4} & $49.2$\stdv{2.1} & $83.7$\stdv{1.0} & $83.4$\stdv{1.4} & $83.8$\stdv{0.8} & $83.2$\stdv{1.2} & $82.5$\stdv{0.9} \\
diff\_energy$_{q4}$      & $42.1$\stdv{0.4} & $55.6$\stdv{1.5} & $66.2$\stdv{1.1} & $68.1$\stdv{0.6} & $68.2$\stdv{0.8} & $68.3$\stdv{0.7} & $67.7$\stdv{1.0} \\
spectral\_entropy$_{q4}$ & $30.2$\stdv{1.0} & $47.0$\stdv{0.6} & $80.2$\stdv{0.7} & $79.5$\stdv{1.0} & $80.4$\stdv{0.7} & $79.7$\stdv{0.7} & $78.8$\stdv{0.9} \\
\midrule
\multicolumn{8}{l}{\textit{Boundary tasks and local\_global\_trend (classification)}} \\
slope\_magnitude$_{q4}$  & $33.4$\stdv{0.0} & $40.2$\stdv{0.8} & $80.5$\stdv{1.8} & $77.6$\stdv{1.9} & $79.1$\stdv{1.9} & $76.2$\stdv{2.3} & $76.5$\stdv{1.6} \\
change\_point            & $70.0$\stdv{0.2} & $74.5$\stdv{1.1} & $97.2$\stdv{0.5} & $96.8$\stdv{0.5} & $97.2$\stdv{0.5} & $96.7$\stdv{0.5} & $96.8$\stdv{0.3} \\
local\_global\_trend     & $60.1$\stdv{0.2} & $59.9$\stdv{0.3} & $60.6$\stdv{0.6} & $59.8$\stdv{0.6} & $60.2$\stdv{0.9} & $58.8$\stdv{0.7} & $59.1$\stdv{0.4} \\
\midrule
\multicolumn{8}{l}{\textit{Inter-token (classification)}} \\
adjacent\_similarity     & $54.5$\stdv{0.2} & $72.0$\stdv{0.6} & $87.1$\stdv{0.2} & $86.7$\stdv{0.2} & $86.9$\stdv{0.2} & $86.7$\stdv{0.5} & $86.7$\stdv{0.4} \\
relative\_magnitude      & $99.5$\stdv{0.1} & $99.8$\stdv{0.0} & $99.8$\stdv{0.0} & $99.8$\stdv{0.0} & $99.7$\stdv{0.0} & $99.7$\stdv{0.0} & $99.6$\stdv{0.0} \\
position\_quarter        & $100.0$\stdv{0.0} & $99.9$\stdv{0.0} & $95.5$\stdv{1.2} & $88.3$\stdv{1.9} & $82.4$\stdv{2.3} & $74.3$\stdv{1.7} & $69.5$\stdv{1.6} \\
trend\_consistency       & $61.5$\stdv{0.1} & $65.5$\stdv{0.5} & $72.5$\stdv{0.5} & $73.1$\stdv{0.3} & $73.2$\stdv{0.3} & $72.9$\stdv{0.5} & $73.0$\stdv{0.4} \\
\midrule
\multicolumn{8}{l}{\textit{Regression ($R^2$)}} \\
reg\_variance            & $0.009$\stdv{0.000} & $0.253$\stdv{0.041} & $0.961$\stdv{0.005} & $0.949$\stdv{0.009} & $0.951$\stdv{0.005} & $0.941$\stdv{0.009} & $0.944$\stdv{0.006} \\
reg\_slope               & $1.000$\stdv{0.000} & $1.000$\stdv{0.000} & $1.000$\stdv{0.000} & $0.999$\stdv{0.000} & $0.999$\stdv{0.000} & $0.998$\stdv{0.000} & $0.998$\stdv{0.000} \\
reg\_next\_mean         & $0.923$\stdv{0.001} & $0.933$\stdv{0.001} & $0.933$\stdv{0.001} & $0.934$\stdv{0.001} & $0.934$\stdv{0.001} & $0.935$\stdv{0.001} & $0.935$\stdv{0.001} \\
\midrule
\multicolumn{8}{l}{\textit{Control}} \\
random\_label            & $25.3$\stdv{0.2} & $25.3$\stdv{0.1} & $25.3$\stdv{0.1} & $25.3$\stdv{0.1} & $25.3$\stdv{0.1} & $25.3$\stdv{0.1} & $25.2$\stdv{0.0} \\
\bottomrule
\end{tabular}
\end{table}

\subsection{PatchTST Regression Results}
\label{app:reg_variance}

Table~\ref{tab:reg_variance} gives reg\_variance and reg\_slope on all 8 datasets.

\begin{table}[H]
\caption{PatchTST reg\_variance $R^2$ from embed to 1b\_ffn (9-run mean $\pm$ std); the last column is reg\_slope $R^2$ at the embedding (9-run mean). At 1b\_ffn, $R^2$ stays low on ETTh1, ETTh2, and Exchange ($0.175$, $0.310$, and $0.251$) and is at least $0.890$ on the other 5 datasets.}
\label{tab:reg_variance}
\centering
\small
\begin{tabular}{lcccc}
\toprule
Dataset & embed & 1b\_ffn & 1b\_ffn $-$ embed & reg\_slope (embed) \\
\midrule
ETTm1 & $0.009$\stdv{0.000} & $0.961$\stdv{0.005} & +0.952 & 1.000 \\
ETTm2 & $0.006$\stdv{0.003} & $0.890$\stdv{0.013} & +0.884 & 1.000 \\
ETTh1 & $0.043$\stdv{0.002} & $0.175$\stdv{0.072} & +0.132 & 0.996 \\
ETTh2 & $0.002$\stdv{0.001} & $0.310$\stdv{0.067} & +0.308 & 0.993 \\
Weather & $0.023$\stdv{0.002} & $0.923$\stdv{0.005} & +0.900 & 1.000 \\
Electricity & $0.152$\stdv{0.003} & $0.932$\stdv{0.003} & +0.780 & 1.000 \\
Exchange & $-0.015$\stdv{0.010} & $0.251$\stdv{0.069} & +0.265 & 0.983 \\
Traffic & $0.496$\stdv{0.003} & $0.943$\stdv{0.002} & +0.446 & 1.000 \\
\bottomrule
\end{tabular}
\end{table}

\subsection{iTransformer Full Results}
\label{app:itrans_full}

Labels are computed on each variate's series before the model's per-variate normalization, because after it every series has nearly the same variance and volatility would be nearly constant. A token, a linear map of the normalized series, therefore lacks the series' mean and standard deviation, on which the constructed-task labels depend. Volatility is the variance itself, diff\_energy scales with it, and spectral\_entropy, whose spectrum includes the zero-frequency bin, changes with the ratio of the mean to the standard deviation; trend\_direction is unaffected, since the normalization does not change its sign. Part of a gain on these tasks can therefore come from other tokens (other variates and the appended time-feature tokens), which only attention reads. Table~\ref{tab:itrans_full} gives the per-task deltas.

\begin{table}[H]
\caption{iTransformer per-task Layer-1 delta (9-run mean $\pm$ std). Avg: mean of the three tasks; Mean: mean of the 8 datasets.}
\label{tab:itrans_full}
\centering
\scriptsize
\setlength{\tabcolsep}{3pt}
\begin{tabular}{lcccccccc}
\toprule
Dataset & Vol Attn $\Delta$ & Vol FFN $\Delta$ & DE Attn $\Delta$ & DE FFN $\Delta$ & SE Attn $\Delta$ & SE FFN $\Delta$ & Avg Attn $\Delta$ & Avg FFN $\Delta$ \\
\midrule
ETTm1 & $+15.5$\stdv{1.5} & $+5.0$\stdv{2.1} & $+13.9$\stdv{0.8} & $+3.8$\stdv{0.9} & $+10.5$\stdv{1.0} & $+6.6$\stdv{1.9} & $+13.3$\stdv{1.1} & $+5.1$\stdv{1.6} \\
ETTm2 & $+5.7$\stdv{0.7} & $+1.9$\stdv{0.6} & $+8.8$\stdv{2.6} & $+4.2$\stdv{0.9} & $+14.1$\stdv{2.9} & $+1.5$\stdv{1.7} & $+9.5$\stdv{2.1} & $+2.5$\stdv{1.0} \\
ETTh1 & $+24.7$\stdv{1.5} & $+7.5$\stdv{1.5} & $+23.8$\stdv{0.9} & $+8.8$\stdv{1.4} & $+14.6$\stdv{1.0} & $+9.3$\stdv{0.9} & $+21.0$\stdv{1.1} & $+8.5$\stdv{1.3} \\
ETTh2 & $+10.4$\stdv{1.4} & $+7.1$\stdv{1.1} & $+6.2$\stdv{0.8} & $+3.4$\stdv{0.9} & $+11.9$\stdv{1.1} & $+2.4$\stdv{1.1} & $+9.5$\stdv{1.1} & $+4.3$\stdv{1.0} \\
Weather & $+12.0$\stdv{0.7} & $+13.3$\stdv{0.7} & $+15.4$\stdv{1.7} & $+12.7$\stdv{0.8} & $+14.4$\stdv{0.4} & $+10.6$\stdv{0.4} & $+13.9$\stdv{0.9} & $+12.2$\stdv{0.7} \\
Electricity & $+13.5$\stdv{0.4} & $+2.6$\stdv{0.4} & $+15.6$\stdv{0.6} & $+4.0$\stdv{0.5} & $+21.9$\stdv{0.5} & $+2.9$\stdv{0.4} & $+17.0$\stdv{0.5} & $+3.2$\stdv{0.4} \\
Exchange & $+2.6$\stdv{1.2} & $+4.8$\stdv{1.4} & $+3.5$\stdv{1.2} & $+2.4$\stdv{1.2} & $+4.8$\stdv{1.1} & $+2.8$\stdv{1.1} & $+3.6$\stdv{1.2} & $+3.3$\stdv{1.2} \\
Traffic & $+9.2$\stdv{0.4} & $+2.9$\stdv{0.5} & $+19.1$\stdv{0.6} & $+5.3$\stdv{0.5} & $+39.4$\stdv{0.6} & $+6.5$\stdv{0.9} & $+22.6$\stdv{0.5} & $+4.9$\stdv{0.6} \\
\midrule
\textbf{Mean} & $+11.7$\stdv{1.0} & $+5.6$\stdv{1.1} & $+13.3$\stdv{1.1} & $+5.6$\stdv{0.9} & $+16.5$\stdv{1.1} & $+5.3$\stdv{1.0} & $+13.8$\stdv{1.1} & $+5.5$\stdv{1.0} \\
\bottomrule\end{tabular}
\end{table}

\subsection{CATS Full Results}
\label{app:cats_full}

\begin{table}[H]
\caption{CATS per-task results (9-run mean $\pm$ std). Absolute accuracy of the cross-attention output (1a\_cross) and FFN delta from 1a\_cross to 1b\_ffn. Query-side labels are those of the future patch each query forecasts.}
\label{tab:cats_full}
\centering
\scriptsize
\setlength{\tabcolsep}{3pt}
\begin{tabular}{lcccccc}
\toprule
Dataset & Vol 1a\_cross & Vol FFN $\Delta$ & DE 1a\_cross & DE FFN $\Delta$ & SE 1a\_cross & SE FFN $\Delta$ \\
\midrule
ETTm1 & $41.8$\stdv{1.0} & $+2.2$\stdv{1.1} & $48.5$\stdv{0.5} & $+1.2$\stdv{0.7} & $41.3$\stdv{0.3} & $+1.4$\stdv{0.4} \\
ETTm2 & $39.3$\stdv{0.3} & $+1.2$\stdv{0.5} & $46.7$\stdv{0.3} & $+1.5$\stdv{0.7} & $38.7$\stdv{0.1} & $+1.5$\stdv{0.5} \\
ETTh1 & $36.5$\stdv{0.6} & $+2.5$\stdv{0.7} & $46.5$\stdv{0.4} & $+2.2$\stdv{0.5} & $43.7$\stdv{0.3} & $+0.2$\stdv{0.9} \\
ETTh2 & $37.0$\stdv{0.4} & $+1.9$\stdv{0.6} & $38.5$\stdv{0.7} & $+8.2$\stdv{1.1} & $37.1$\stdv{0.4} & $+3.5$\stdv{1.0} \\
Weather & $48.1$\stdv{0.3} & $+3.5$\stdv{0.4} & $56.9$\stdv{0.2} & $+3.8$\stdv{0.3} & $45.7$\stdv{0.1} & $+3.5$\stdv{0.2} \\
Electricity & $38.2$\stdv{0.2} & $+1.4$\stdv{0.2} & $51.6$\stdv{0.2} & $+1.0$\stdv{0.3} & $53.4$\stdv{0.4} & $+1.8$\stdv{0.4} \\
Exchange & $38.3$\stdv{0.6} & $+2.2$\stdv{0.7} & $41.5$\stdv{1.2} & $+3.6$\stdv{1.3} & $39.6$\stdv{0.8} & $+1.6$\stdv{0.9} \\
Traffic & $54.2$\stdv{0.1} & $+3.9$\stdv{0.2} & $52.1$\stdv{0.1} & $+3.7$\stdv{0.2} & $45.5$\stdv{0.1} & $+3.2$\stdv{0.1} \\
\bottomrule
\end{tabular}
\end{table}

CATS's learnable queries start without per-instance input information and hold no input points, so their labels are computed on the future patch each query forecasts, normalized with the input window's mean and standard deviation as in the model; the accuracies thus also reflect how predictable these properties are from the input. The absolute accuracy at 1a\_cross (cross-attention output) does not isolate cross-attention: the learned queries already encode the position and, without query independence (Appendix~\ref{app:setup_details}), the variate. The FFN adds $+0.2$ to $+8.2$\,pp afterward (Table~\ref{tab:cats_full}).

\subsection{PatchTST-NoAttn Full Results}
\label{app:noattn_full}

\begin{table}[H]
\caption{Per-dataset Layer-1 constructed-task deltas without attention (9-run mean $\pm$ std). FFN change and total change are differences from PatchTST, where the total is Attn $\Delta$ + FFN $\Delta$ (NoAttn's Attn $\Delta$ is in Table~\ref{tab:noattn_pertask}). The 8-dataset means are in Table~\ref{tab:noattn_r2}.}
\label{tab:noattn}
\centering
\scriptsize
\setlength{\tabcolsep}{3pt}
\begin{tabular}{lcccccccc}
\toprule
 & \multicolumn{2}{c}{PatchTST} & \multicolumn{3}{c}{NoAttn, ablated} & \multicolumn{3}{c}{NoAttn, from scratch} \\
\cmidrule(lr){2-3}\cmidrule(lr){4-6}\cmidrule(lr){7-9}
Dataset & Attn $\Delta$ & FFN $\Delta$ & FFN $\Delta$ & FFN change & Total change & FFN $\Delta$ & FFN change & Total change \\
\midrule
ETTm1 & $+13.5$\stdv{1.6} & $+26.1$\stdv{1.7} & $+38.8$\stdv{1.2} & $+12.7$ & $-0.8$ & $+44.9$\stdv{2.4} & $+18.8$ & $+5.2$ \\
ETTm2 & $+0.2$\stdv{0.5} & $+43.0$\stdv{0.7} & $+44.2$\stdv{0.5} & $+1.2$ & $+0.7$ & $+43.9$\stdv{0.6} & $+0.9$ & $+0.3$ \\
ETTh1 & $+0.0$\stdv{0.4} & $+3.8$\stdv{1.8} & $+4.1$\stdv{1.9} & $+0.3$ & $+0.3$ & $+4.1$\stdv{1.7} & $+0.3$ & $+0.3$ \\
ETTh2 & $-0.6$\stdv{0.9} & $+11.1$\stdv{2.2} & $+12.1$\stdv{2.8} & $+1.0$ & $+1.4$ & $+9.8$\stdv{3.9} & $-1.3$ & $-0.8$ \\
Weather & $+16.7$\stdv{0.7} & $+21.5$\stdv{1.8} & $+36.1$\stdv{2.5} & $+14.6$ & $-2.2$ & $+40.1$\stdv{0.9} & $+18.6$ & $+1.7$ \\
Electricity & $+11.3$\stdv{0.5} & $+23.7$\stdv{0.8} & $+38.1$\stdv{0.5} & $+14.3$ & $+3.0$ & $+40.6$\stdv{0.4} & $+16.9$ & $+5.7$ \\
Exchange & $+0.1$\stdv{0.6} & $+12.8$\stdv{2.4} & $+13.9$\stdv{3.0} & $+1.1$ & $+0.9$ & $+13.8$\stdv{4.3} & $+1.0$ & $+0.8$ \\
Traffic & $+9.5$\stdv{0.4} & $+20.8$\stdv{0.6} & $+32.3$\stdv{0.4} & $+11.5$ & $+2.1$ & $+33.0$\stdv{0.5} & $+12.2$ & $+3.3$ \\
\bottomrule
\end{tabular}
\end{table}

\begin{table}[H]
\caption{PatchTST-NoAttn per-task Layer-1 delta (9-run mean $\pm$ std), ablated (top) and trained from scratch (bottom). ``Attn $\Delta$'' is near zero because the attention output is zero.}
\label{tab:noattn_pertask}
\centering
\scriptsize
\setlength{\tabcolsep}{3pt}
\begin{tabular}{lcccccc}
\toprule
Dataset & Vol Attn $\Delta$ & Vol FFN $\Delta$ & DE Attn $\Delta$ & DE FFN $\Delta$ & SE Attn $\Delta$ & SE FFN $\Delta$ \\
\midrule
\multicolumn{7}{l}{\textit{Ablated}} \\
ETTm1 & $+0.1$\stdv{0.6} & $+46.2$\stdv{1.0} & $-0.6$\stdv{0.7} & $+21.0$\stdv{0.9} & $+0.6$\stdv{1.8} & $+49.1$\stdv{1.7} \\
ETTm2 & $+0.0$\stdv{0.4} & $+52.8$\stdv{0.6} & $-0.4$\stdv{0.2} & $+42.6$\stdv{0.4} & $-0.4$\stdv{0.4} & $+37.3$\stdv{0.5} \\
ETTh1 & $+0.0$\stdv{0.3} & $+2.9$\stdv{1.2} & $+0.0$\stdv{0.3} & $+3.5$\stdv{1.9} & $+0.0$\stdv{0.3} & $+5.8$\stdv{2.6} \\
ETTh2 & $-0.2$\stdv{1.2} & $+10.6$\stdv{4.2} & $-0.2$\stdv{1.0} & $+16.4$\stdv{1.7} & $-0.1$\stdv{0.6} & $+9.2$\stdv{2.5} \\
Weather & $+0.0$\stdv{0.3} & $+41.3$\stdv{2.7} & $+0.2$\stdv{0.6} & $+31.7$\stdv{1.9} & $-0.5$\stdv{0.7} & $+35.4$\stdv{2.9} \\
Electricity & $+0.0$\stdv{0.2} & $+49.6$\stdv{0.5} & $+0.1$\stdv{0.4} & $+42.0$\stdv{0.7} & $+0.0$\stdv{0.1} & $+22.6$\stdv{0.2} \\
Exchange & $+0.1$\stdv{0.4} & $+12.8$\stdv{2.6} & $+0.0$\stdv{0.4} & $+8.2$\stdv{2.0} & $-0.5$\stdv{0.6} & $+20.8$\stdv{4.2} \\
Traffic & $+0.3$\stdv{0.1} & $+40.2$\stdv{0.3} & $+0.0$\stdv{0.3} & $+33.6$\stdv{0.5} & $+0.0$\stdv{0.3} & $+23.0$\stdv{0.4} \\
\midrule
\multicolumn{7}{l}{\textit{From scratch}} \\
ETTm1 & $+0.2$\stdv{0.6} & $+50.8$\stdv{1.5} & $-1.0$\stdv{0.9} & $+31.5$\stdv{4.8} & $+0.6$\stdv{0.8} & $+52.4$\stdv{1.1} \\
ETTm2 & $-0.1$\stdv{0.2} & $+52.7$\stdv{0.7} & $-0.4$\stdv{0.3} & $+41.7$\stdv{0.4} & $-0.7$\stdv{0.7} & $+37.3$\stdv{0.8} \\
ETTh1 & $+0.0$\stdv{0.3} & $+3.3$\stdv{1.0} & $+0.1$\stdv{0.3} & $+3.8$\stdv{1.9} & $+0.0$\stdv{0.3} & $+5.1$\stdv{2.2} \\
ETTh2 & $-0.2$\stdv{1.8} & $+7.1$\stdv{5.1} & $+0.0$\stdv{1.3} & $+14.3$\stdv{3.4} & $-0.1$\stdv{1.2} & $+8.0$\stdv{3.4} \\
Weather & $+0.0$\stdv{0.3} & $+45.1$\stdv{0.9} & $+0.5$\stdv{0.6} & $+36.5$\stdv{0.7} & $-1.1$\stdv{0.7} & $+38.8$\stdv{1.1} \\
Electricity & $+0.2$\stdv{0.3} & $+50.2$\stdv{0.4} & $+0.1$\stdv{0.3} & $+47.5$\stdv{0.5} & $+0.3$\stdv{0.3} & $+24.1$\stdv{0.3} \\
Exchange & $+0.1$\stdv{0.4} & $+13.6$\stdv{3.9} & $+0.1$\stdv{0.3} & $+8.7$\stdv{2.8} & $-0.4$\stdv{0.6} & $+19.2$\stdv{6.2} \\
Traffic & $+0.9$\stdv{0.6} & $+40.9$\stdv{0.4} & $+0.3$\stdv{0.4} & $+34.0$\stdv{0.5} & $+0.6$\stdv{0.4} & $+24.0$\stdv{0.6} \\
\bottomrule
\end{tabular}
\end{table}

\subsection{PatchTST ($P{=}1$) Full Results}
\label{app:p1_full}

\begin{table}[H]
\caption{PatchTST ($P{=}1$) with the mean adapter, per-task Layer-1 delta (9-run mean $\pm$ std). Avg: mean of the three tasks; Mean: mean of the 8 datasets. The FFN delta is small on every dataset except ETTh2, while the attention delta (mean $+10.9$\,pp) is positive for every task on every dataset but Exchange.}
\label{tab:p1_full}
\centering
\scriptsize
\setlength{\tabcolsep}{3pt}
\begin{tabular}{lcccccccc}
\toprule
Dataset & Vol Attn $\Delta$ & Vol FFN $\Delta$ & DE Attn $\Delta$ & DE FFN $\Delta$ & SE Attn $\Delta$ & SE FFN $\Delta$ & Avg Attn $\Delta$ & Avg FFN $\Delta$ \\
\midrule
ETTm1 & $+15.0$\stdv{1.7} & $+1.4$\stdv{3.4} & $+16.5$\stdv{2.5} & $+2.5$\stdv{3.6} & $+4.1$\stdv{2.4} & $+2.6$\stdv{3.1} & $+11.9$\stdv{2.2} & $+2.2$\stdv{3.4} \\
ETTm2 & $+14.0$\stdv{0.7} & $+2.0$\stdv{0.9} & $+14.8$\stdv{1.1} & $+4.2$\stdv{1.7} & $+15.7$\stdv{1.0} & $+2.0$\stdv{1.3} & $+14.8$\stdv{1.0} & $+2.7$\stdv{1.3} \\
ETTh1 & $+18.2$\stdv{3.3} & $-3.4$\stdv{3.5} & $+14.3$\stdv{2.3} & $-3.4$\stdv{2.4} & $+3.5$\stdv{1.9} & $+4.1$\stdv{2.1} & $+12.0$\stdv{2.5} & $-0.9$\stdv{2.7} \\
ETTh2 & $+3.4$\stdv{1.0} & $+10.0$\stdv{1.8} & $+5.7$\stdv{1.3} & $+7.8$\stdv{1.9} & $+12.0$\stdv{4.7} & $+5.2$\stdv{5.7} & $+7.0$\stdv{2.3} & $+7.7$\stdv{3.1} \\
Weather & $+16.3$\stdv{0.8} & $+0.5$\stdv{1.0} & $+15.8$\stdv{1.3} & $+2.7$\stdv{1.3} & $+14.3$\stdv{0.6} & $-0.1$\stdv{0.5} & $+15.5$\stdv{0.9} & $+1.0$\stdv{0.9} \\
Electricity & $+19.8$\stdv{1.3} & $+5.7$\stdv{1.9} & $+10.7$\stdv{1.2} & $+3.7$\stdv{1.8} & $+8.0$\stdv{0.7} & $+0.9$\stdv{1.3} & $+12.8$\stdv{1.1} & $+3.4$\stdv{1.7} \\
Exchange & $-0.3$\stdv{0.5} & $+0.5$\stdv{0.6} & $+0.2$\stdv{0.4} & $+0.1$\stdv{0.5} & $+1.6$\stdv{1.3} & $-0.2$\stdv{1.6} & $+0.5$\stdv{0.8} & $+0.1$\stdv{0.9} \\
Traffic & $+19.1$\stdv{0.6} & $+1.9$\stdv{0.2} & $+14.6$\stdv{0.6} & $+1.1$\stdv{0.6} & $+4.4$\stdv{0.8} & $+2.2$\stdv{0.7} & $+12.7$\stdv{0.7} & $+1.7$\stdv{0.5} \\
\midrule
\textbf{Mean} & $+13.2$\stdv{1.2} & $+2.3$\stdv{1.7} & $+11.6$\stdv{1.3} & $+2.3$\stdv{1.7} & $+8.0$\stdv{1.7} & $+2.1$\stdv{2.0} & $+10.9$\stdv{1.4} & $+2.3$\stdv{1.8} \\
\bottomrule\end{tabular}
\end{table}

The constructed-task averages over the 8 datasets are $\Delta_{\text{attn}} = +10.9$ and $\Delta_{\text{FFN}} = +2.3$. ETTh2 is the exception: its FFN delta ($+7.7$) exceeds its attention delta ($+7.0$), although it stays below its $P{=}16$ value ($+11.1$, Table~\ref{tab:patchtst_delta}); on the other 7 datasets the FFN delta is at most $+3.4$. On Exchange both components stay small ($+0.5$ and $+0.1$).

\begin{table}[H]
\caption{Constructed-task deltas summed over the three encoder layers for PatchTST at $P{=}16$ and $P{=}1$ (deltas computed from the rounded 9-run means at each probing point). L1: Layer 1; $\Sigma$: sum over the three layers; ``Max deeper FFN'' is the larger of the Layer-2 and Layer-3 FFN deltas.}
\label{tab:p1_depth}
\centering
\footnotesize
\setlength{\tabcolsep}{4pt}
\begin{tabular}{lcccccc}
\toprule
 & \multicolumn{2}{c}{$P{=}16$} & \multicolumn{4}{c}{$P{=}1$} \\
\cmidrule(lr){2-3}\cmidrule(lr){4-7}
Dataset & L1 FFN $\Delta$ & $\Sigma$ FFN $\Delta$ & $\Sigma$ Attn $\Delta$ & L1 FFN $\Delta$ & $\Sigma$ FFN $\Delta$ & Max deeper FFN \\
\midrule
ETTm1 & $+26.1$ & $+25.8$ & $+17.0$ & $+2.2$ & $+1.4$ & $-0.2$ \\
ETTm2 & $+43.0$ & $+43.8$ & $+21.9$ & $+2.7$ & $+2.2$ & $-0.2$ \\
ETTh1 & $+3.8$ & $+5.2$ & $+17.5$ & $-0.9$ & $-3.0$ & $-0.4$ \\
ETTh2 & $+11.1$ & $+12.4$ & $+14.9$ & $+7.7$ & $+7.9$ & $+0.2$ \\
Weather & $+21.5$ & $+22.8$ & $+25.6$ & $+1.0$ & $-0.9$ & $-0.7$ \\
Electricity & $+23.7$ & $+23.6$ & $+14.7$ & $+3.4$ & $+2.9$ & $+1.3$ \\
Exchange & $+12.8$ & $+15.1$ & $+1.0$ & $+0.1$ & $-0.1$ & $0.0$ \\
Traffic & $+20.8$ & $+19.5$ & $+13.9$ & $+1.7$ & $-0.6$ & $+0.3$ \\
\midrule
\textbf{8-dataset mean} & $\mathbf{+20.4}$ & $\mathbf{+21.0}$ & $\mathbf{+15.8}$ & $\mathbf{+2.3}$ & $\mathbf{+1.2}$ & \\
\bottomrule
\end{tabular}
\end{table}

\textbf{Read-out.} The mean adapter averages the representations of 16 consecutive tokens, which could hide what the FFN writes into individual tokens. With a concatenation adapter that keeps all 16 vectors (a $16 d_{\text{model}}$-dimensional probe input), the FFN delta on the 4 sweep datasets is $+0.8$ (ETTh1), $+3.4$ (ETTm1), $+3.1$ (Weather), and $+3.4$ (Electricity), and attention still leads on all 4 ($+10.9$ to $+17.5$). With concatenation, the embedding already exposes every linear functional of the 16 points, as at $P{=}16$, so attention's lead is not an artifact of what mean pooling discards.

\subsection{Control Results}

\begin{table}[H]
\caption{Random-label control on PatchTST (top; 9-run mean $\pm$ std per cell) and on each model of Figure~\ref{fig:component_dominance} and Table~\ref{tab:noattn_r2} (bottom). Each row of the bottom block gives the mean $\pm$ std and the range of the model's 9-run cell means over the 8 datasets and its probing points (Layer 1 only for NoAttn from scratch). Labels have 4 classes drawn independently for each token (chance $25.0\%$); this control marks the chance floor used in Appendix~\ref{app:selectivity}.}
\label{tab:random-label-control}
\centering
\scriptsize
\setlength{\tabcolsep}{3pt}
\begin{tabular}{lccccccc}
\toprule
Dataset & embed & 1a\_attn & 1b\_ffn & 2a\_attn & 2b\_ffn & 3a\_attn & 3b\_ffn \\
\midrule
ETTm1       & $25.3$\stdv{0.2} & $25.3$\stdv{0.1} & $25.3$\stdv{0.1} & $25.3$\stdv{0.1} & $25.3$\stdv{0.1} & $25.3$\stdv{0.1} & $25.2$\stdv{0.0} \\
ETTm2       & $25.4$\stdv{0.1} & $25.3$\stdv{0.1} & $25.4$\stdv{0.2} & $25.6$\stdv{0.2} & $25.5$\stdv{0.2} & $25.4$\stdv{0.2} & $25.4$\stdv{0.2} \\
ETTh1       & $25.4$\stdv{0.2} & $25.4$\stdv{0.2} & $25.4$\stdv{0.1} & $25.4$\stdv{0.2} & $25.4$\stdv{0.1} & $25.4$\stdv{0.2} & $25.5$\stdv{0.1} \\
ETTh2       & $25.2$\stdv{0.2} & $25.1$\stdv{0.2} & $25.2$\stdv{0.2} & $25.2$\stdv{0.2} & $25.2$\stdv{0.2} & $25.2$\stdv{0.2} & $25.3$\stdv{0.3} \\
Weather     & $25.3$\stdv{0.1} & $25.3$\stdv{0.2} & $25.3$\stdv{0.1} & $25.3$\stdv{0.2} & $25.3$\stdv{0.1} & $25.4$\stdv{0.1} & $25.3$\stdv{0.1} \\
Electricity & $25.3$\stdv{0.1} & $25.3$\stdv{0.2} & $25.4$\stdv{0.1} & $25.3$\stdv{0.2} & $25.3$\stdv{0.2} & $25.3$\stdv{0.2} & $25.3$\stdv{0.2} \\
Exchange    & $25.2$\stdv{0.1} & $25.3$\stdv{0.1} & $25.2$\stdv{0.2} & $25.2$\stdv{0.2} & $25.2$\stdv{0.1} & $25.2$\stdv{0.2} & $25.3$\stdv{0.1} \\
Traffic     & $25.3$\stdv{0.2} & $25.4$\stdv{0.3} & $25.3$\stdv{0.2} & $25.3$\stdv{0.2} & $25.3$\stdv{0.3} & $25.2$\stdv{0.3} & $25.2$\stdv{0.3} \\
\midrule
Model & \multicolumn{3}{c}{Mean $\pm$ std} & \multicolumn{4}{c}{Range} \\
\midrule
PatchTST & \multicolumn{3}{c}{$25.3$\stdv{0.1}} & \multicolumn{4}{c}{25.1 to 25.6} \\
NoAttn, ablated & \multicolumn{3}{c}{$25.3$\stdv{0.1}} & \multicolumn{4}{c}{25.2 to 25.5} \\
NoAttn, from scratch & \multicolumn{3}{c}{$25.3$\stdv{0.1}} & \multicolumn{4}{c}{25.2 to 25.5} \\
PatchTST ($P{=}1$) & \multicolumn{3}{c}{$25.4$\stdv{0.1}} & \multicolumn{4}{c}{25.3 to 25.5} \\
iTransformer & \multicolumn{3}{c}{$25.6$\stdv{0.3}} & \multicolumn{4}{c}{25.2 to 26.4} \\
CATS & \multicolumn{3}{c}{$25.3$\stdv{0.2}} & \multicolumn{4}{c}{25.1 to 26.0} \\
\bottomrule
\end{tabular}
\end{table}

\subsection{Informer Multi-Dataset Probing}
\label{app:informer}

We probe Informer on 4 datasets (ETTm1, ETTh1, Weather, Electricity; Table~\ref{tab:informer_multi}). Informer's tokens are not single scalars: each encoder token is a three-step circular convolution over all $V$ variates plus time-feature and positional embeddings. The model also differs from PatchTST in its ProbSparse attention, the absence of RevIN, $\text{seq\_len}=96$, and a single architecture setting for all datasets; a nonlinear decoder, rather than a linear head, reads its encoder output. The single-scalar argument of \S\ref{sec:pointwise} therefore does not apply, and we report this as a confounded auxiliary result.

\begin{table}[H]
\caption{Informer Layer-1 deltas (constructed task average, 9-run mean, mean-adapter probe) across 4 datasets. Both components are flat, but Informer tokens are not single scalars, so this is a confounded auxiliary result (see text).}
\label{tab:informer_multi}
\centering
\small
\begin{tabular}{lcc}
\toprule
Dataset & Attn $\Delta$ & FFN $\Delta$ \\
\midrule
ETTm1 & $+1.62$ & $+2.66$ \\
ETTh1 & $+1.44$ & $+1.75$ \\
Weather & $+0.37$ & $+1.50$ \\
Electricity & $-0.04$ & $-0.02$ \\
\midrule
\textbf{Mean} & $\mathbf{+0.85}$ & $\mathbf{+1.47}$ \\
\bottomrule
\end{tabular}
\end{table}

\textbf{Probe-design confound.} Because an Informer token holds all variates, the mean adapter pools 16 consecutive token representations (stride 8) into one vector, replicates it $V$ times ($V=7/7/21/321$ for ETTm1/ETTh1/Weather/Electricity), and pairs each copy with a label from one variate's own 16-step patch. Each probe input thus appears $V$ times with labels that need not agree, which caps the attainable probing accuracy at every layer and can compress the deltas, more so as $V$ grows. An earlier run probed per-time-step representations against variate-averaged labels on the 16-step window ending at each step and gave FFN deltas of $+5.33$/$+5.09$/$+5.97$/$-0.94$ (ETTm1/ETTh1/Weather/Electricity). Because that run differed not only in the labels but also in the probe unit and (except on ETTh1) the number of test windows, the gap to the FFN deltas above cannot be attributed to label design alone. We therefore read the flat deltas as an open empirical observation, with the model differences above, Informer's training settings, and this probe design as candidate causes.

\subsection{Inter-Token Task Results}
\label{app:intertoken}

Table~\ref{tab:intertoken_results} reports Layer-1 probing deltas for the four inter-token tasks (adjacent\_similarity, relative\_magnitude, position\_quarter, trend\_consistency) across the 4 models analyzed here (PatchTST, iTransformer, CATS, and PatchTST-NoAttn). These tasks' labels depend on a token's position or on other tokens rather than on its own points alone.

\begin{table}[H]
\caption{Inter-token task Layer-1 deltas across architectures (9-run means averaged over the 8 datasets; PatchTST-NoAttn is the ablated model). For CATS, the Attn column gives the accuracy after cross-attention (abs), as in Table~\ref{tab:cats_full} (see note below).}
\label{tab:intertoken_results}
\centering
\small
\begin{tabular}{llcc}
\toprule
Model & Task & Attn $\Delta$ & FFN $\Delta$ \\
\midrule
\multirow{4}{*}{PatchTST}
& adjacent\_similarity & $+5.7$ & $+12.5$ \\
& relative\_magnitude & $+0.0$ & $-0.3$ \\
& position\_quarter & $-0.9$ & $-3.4$ \\
& trend\_consistency & $+0.7$ & $+2.9$ \\
\midrule
\multirow{4}{*}{iTransformer}
& adjacent\_similarity & $+2.2$ & $+1.2$ \\
& relative\_magnitude & $+5.0$ & $+3.5$ \\
& position\_quarter & $+11.3$ & $+8.1$ \\
& trend\_consistency & $+1.5$ & $+1.2$ \\
\midrule
\multirow{4}{*}{CATS$^{\dagger}$}
& adjacent\_similarity & $76.2$ (abs) & $+1.4$ \\
& relative\_magnitude & $63.4$ (abs) & $+2.2$ \\
& position\_quarter & $98.5$ (abs) & $+0.9$ \\
& trend\_consistency & $90.0$ (abs) & $+0.2$ \\
\midrule
\multirow{4}{*}{PatchTST-NoAttn}
& adjacent\_similarity & $+0.1$ & $+18.5$ \\
& relative\_magnitude & $+0.0$ & $-0.3$ \\
& position\_quarter & $-0.0$ & $-2.5$ \\
& trend\_consistency & $+0.0$ & $+3.5$ \\
\bottomrule
\multicolumn{4}{l}{\footnotesize $^{\dagger}$ Attn: absolute accuracy (abs) at \texttt{1a\_cross}, not a delta: the embedding lies on} \\
\multicolumn{4}{l}{\footnotesize \phantom{$^{\dagger}$} the key-value side and \texttt{1a\_cross} on the query side, whose labels are those of the} \\
\multicolumn{4}{l}{\footnotesize \phantom{$^{\dagger}$} forecast patches. FFN $\Delta$: from \texttt{1a\_cross} to \texttt{1b\_ffn}, within the query side.} \\
\end{tabular}
\end{table}

These results reveal architecture-dependent patterns on the inter-token tasks that are qualitatively distinct from the intra-token feature construction patterns analyzed in the main text. PatchTST's FFN strongly raises the linear decodability of inter-patch similarity ($+12.5$\,pp on adjacent\_similarity), and PatchTST-NoAttn's FFN raises it even more ($+18.5$\,pp). NoAttn has no token mixing, so this gain arises within each token: with stride 8, a 16-point patch already shares 8 time points with the next. iTransformer's attention across variates contributes more than its FFN to position and relative-magnitude information ($+11.3$\,pp on position\_quarter, $+5.0$\,pp on relative\_magnitude). For variate tokens, the tasks are defined across variates: adjacent\_similarity is whether a variate's series correlates positively with the next variate's (the previous variate's for the last variate), relative\_magnitude whether its mean exceeds the mean over variates, position\_quarter which quarter of the variate order it falls in, and trend\_consistency whether the sign of its end-minus-start change agrees with the majority of variates. Because relative\_magnitude is computed before the normalization, it depends on the variates' means, which no token holds. A mechanistic interpretation of these patterns, particularly of the role of attention heads in inter-token routing versus the FFN's encoding of token-level features, requires dedicated analytical machinery (e.g., per-head attention attribution) and is left for future work.

\subsection{Additional Probing Targets}
\label{app:new_targets}
Table~\ref{tab:new_targets} reports Layer-1 deltas for the three targets added in \S\ref{sec:tasks}. \emph{Seasonal concentration} is the share of nonzero-frequency power in the largest spectral bin, $\max_k p_k / \sum_k p_k$, where $p_k$ is bin $k$ of $|\text{rfft}(\tilde{x})|^2$ and $k$ runs over the nonzero frequencies. The \emph{multi-scale ratio} is $\text{var}(c)/\text{var}(\tilde{x} - c')$, where $c$ averages adjacent pairs of the patch and $c'$ repeats each average twice. Both are 4-class quartile tasks, with embedding probing accuracies of $29.5$ to $52.0\%$ and $29.2$ to $41.7\%$. \emph{Long-range consistency} is binary: whether the sign of the slope within token $i$ equals the sign of the slope through the means of tokens $i{-}8,\dots,i{-}1$ (tokens with $i<8$ are excluded). Unlike the class boundaries of Appendix~\ref{app:setup_details}, the quartile thresholds of the first two targets are computed on the probe's training part only. The FFN delta exceeds the attention delta on 7, 7, and 6 of the 8 datasets, and on 5 of the 8 for each target it also reaches at least $3$\,pp. The exceptions are Weather on the first two targets and, on long-range consistency, Traffic (attention ahead) and ETTh2 (a tie at zero); ETTh1 and Exchange are also weak on the first two targets, and ETTh1 is near flat on long-range consistency. Weather is attention-leaning on both intra-token targets, consistent with its having the weakest FFN/attention ratio in Table~\ref{tab:patchtst_delta} (1.3$\times$).
\begin{table}[H]
\caption{Layer-1 deltas for the additional probing targets on PatchTST (9-run mean $\pm$ std). The control is a binary random label on the long-range subset.}
\label{tab:new_targets}
\centering
\scriptsize
\setlength{\tabcolsep}{2.5pt}
\begin{tabular}{lcccccccc}
\toprule
& \multicolumn{2}{c}{Seasonal conc.} & \multicolumn{2}{c}{Multi-scale ratio} & \multicolumn{2}{c}{Long-range cons.} & \multicolumn{2}{c}{Control} \\
\cmidrule(lr){2-3} \cmidrule(lr){4-5} \cmidrule(lr){6-7} \cmidrule(lr){8-9}
Dataset & Attn $\Delta$ & FFN $\Delta$ & Attn $\Delta$ & FFN $\Delta$ & Attn $\Delta$ & FFN $\Delta$ & Attn $\Delta$ & FFN $\Delta$ \\
\midrule
ETTm1 & $+3.5$\stdv{0.3} & $+19.3$\stdv{0.5} & $+4.8$\stdv{0.9} & $+15.3$\stdv{0.9} & $+2.1$\stdv{0.2} & $+7.6$\stdv{0.4} & $+0.0$\stdv{0.3} & $+0.0$\stdv{0.3} \\
ETTm2 & $+10.2$\stdv{0.6} & $+13.3$\stdv{0.8} & $+14.5$\stdv{0.4} & $+15.7$\stdv{1.8} & $+3.0$\stdv{0.3} & $+8.7$\stdv{0.4} & $-0.2$\stdv{0.3} & $+0.1$\stdv{0.3} \\
ETTh1 & $-0.2$\stdv{0.4} & $+1.3$\stdv{1.2} & $+1.3$\stdv{0.5} & $+2.9$\stdv{1.4} & $+0.0$\stdv{0.4} & $+0.3$\stdv{0.6} & $-0.1$\stdv{0.1} & $+0.0$\stdv{0.1} \\
ETTh2 & $-0.6$\stdv{1.1} & $+5.9$\stdv{1.2} & $-0.4$\stdv{0.8} & $+4.3$\stdv{1.2} & $+0.0$\stdv{0.4} & $+0.0$\stdv{0.4} & $+0.0$\stdv{0.3} & $+0.0$\stdv{0.3} \\
Weather & $+10.0$\stdv{0.8} & $+7.4$\stdv{1.1} & $+12.5$\stdv{0.7} & $+8.1$\stdv{1.1} & $+5.8$\stdv{0.8} & $+16.4$\stdv{0.8} & $+0.0$\stdv{0.2} & $+0.0$\stdv{0.3} \\
Electricity & $+10.7$\stdv{0.4} & $+22.6$\stdv{0.7} & $+12.0$\stdv{0.4} & $+15.2$\stdv{1.3} & $+1.3$\stdv{0.6} & $+3.9$\stdv{0.7} & $+0.0$\stdv{0.2} & $+0.0$\stdv{0.2} \\
Exchange & $+0.4$\stdv{0.5} & $+1.0$\stdv{1.1} & $+0.0$\stdv{0.3} & $+2.7$\stdv{0.6} & $-0.5$\stdv{0.5} & $+8.6$\stdv{1.9} & $+0.1$\stdv{0.1} & $+0.0$\stdv{0.1} \\
Traffic & $+4.6$\stdv{0.4} & $+25.1$\stdv{0.8} & $+6.2$\stdv{0.3} & $+11.8$\stdv{0.4} & $+5.1$\stdv{0.4} & $+0.6$\stdv{0.4} & $+0.0$\stdv{0.2} & $+0.1$\stdv{0.1} \\
\midrule
\textbf{8-dataset mean} & $\mathbf{+4.8}$ & $\mathbf{+12.0}$ & $\mathbf{+6.4}$ & $\mathbf{+9.5}$ & $\mathbf{+2.1}$ & $\mathbf{+5.8}$ & $\mathbf{+0.0}$ & $\mathbf{+0.0}$ \\
\bottomrule
\end{tabular}
\end{table}

\subsection{Patch-Length Sweep}
\label{app:patch_sweep}

\begin{table}[H]
\caption{PatchTST Layer-1 constructed-task deltas by patch length (9-run mean $\pm$ std). $P{=}1$ and $P{=}16$ (stride 8) are the main checkpoints; $P \in \{2, 4, 8, 32\}$ (stride $P/2$) and the non-overlapping $P{=}16$ (stride 16) are separately trained PatchTST checkpoints (\S\ref{sec:models_ablations}). For $P \geq 2$ the tasks are computed on each token's own $P$ points, so rows probe statistics of different lengths (at $P{=}2$, volatility and diff\_energy are both quartiles of $|x_2 - x_1|$ and share their labels); at $P{=}1$ the tasks describe 16-point segments read through the mean adapter. We report deltas rather than FFN/attention ratios because Attn $\Delta$ crosses zero on ETTh1; on ETTm1 the ratio is $0.18\times$ at $P{=}1$ and $1.6$ to $3.0\times$ for $P \geq 2$, on Electricity $0.27\times$ and $2.0$ to $5.1\times$, and on Weather $0.07\times$ and $0.9$ to $1.9\times$.}
\label{tab:patch_sweep}
\centering
\scriptsize
\setlength{\tabcolsep}{3pt}
\begin{tabular}{lcccccccc}
\toprule
 & \multicolumn{2}{c}{ETTh1} & \multicolumn{2}{c}{ETTm1} & \multicolumn{2}{c}{Weather} & \multicolumn{2}{c}{Electricity} \\
\cmidrule(lr){2-3}\cmidrule(lr){4-5}\cmidrule(lr){6-7}\cmidrule(lr){8-9}
$P$ (stride) & Attn $\Delta$ & FFN $\Delta$ & Attn $\Delta$ & FFN $\Delta$ & Attn $\Delta$ & FFN $\Delta$ & Attn $\Delta$ & FFN $\Delta$ \\
\midrule
1 (1) & $+12.0$\stdv{2.5} & $-0.9$\stdv{2.7} & $+11.9$\stdv{2.2} & $+2.2$\stdv{3.4} & $+15.5$\stdv{0.9} & $+1.0$\stdv{0.9} & $+12.8$\stdv{1.1} & $+3.4$\stdv{1.7} \\
2 (1) & $-2.1$\stdv{0.8} & $+23.3$\stdv{3.3} & $+10.4$\stdv{0.7} & $+31.4$\stdv{4.2} & $+13.9$\stdv{2.4} & $+13.0$\stdv{3.2} & $+5.9$\stdv{4.7} & $+30.1$\stdv{4.8} \\
4 (2) & $+0.8$\stdv{0.8} & $+20.3$\stdv{2.9} & $+14.2$\stdv{1.6} & $+31.5$\stdv{4.3} & $+14.5$\stdv{1.2} & $+17.2$\stdv{1.6} & $+7.4$\stdv{0.8} & $+30.7$\stdv{2.9} \\
8 (4) & $+1.0$\stdv{0.7} & $+16.2$\stdv{2.6} & $+16.2$\stdv{1.5} & $+27.3$\stdv{3.1} & $+18.5$\stdv{1.0} & $+16.1$\stdv{1.5} & $+8.4$\stdv{0.6} & $+35.7$\stdv{1.1} \\
16 (8) & $+0.0$\stdv{0.4} & $+3.8$\stdv{1.8} & $+13.5$\stdv{1.6} & $+26.1$\stdv{1.7} & $+16.7$\stdv{0.7} & $+21.5$\stdv{1.8} & $+11.3$\stdv{0.5} & $+23.7$\stdv{0.8} \\
32 (16) & $-0.1$\stdv{0.6} & $+0.5$\stdv{1.2} & $+13.0$\stdv{0.9} & $+21.5$\stdv{1.1} & $+13.5$\stdv{1.0} & $+26.1$\stdv{1.3} & $+10.1$\stdv{0.6} & $+20.5$\stdv{1.0} \\
\midrule
16 (16) & $-0.2$\stdv{0.8} & $+4.0$\stdv{1.4} & $+14.3$\stdv{0.8} & $+26.8$\stdv{0.9} & $+16.1$\stdv{1.0} & $+18.6$\stdv{3.6} & $+10.8$\stdv{0.5} & $+25.5$\stdv{0.8} \\
\bottomrule
\end{tabular}
\end{table}

Whenever a token holds the points a task depends on ($P \geq 2$), the FFN delta is large on ETTm1, Weather, and Electricity, and on ETTh1 for $P \leq 8$. It exceeds the attention delta in 18 of the 20 pairs of dataset and patch length; on Weather, attention is slightly ahead at $P{=}2$ and ahead at $P{=}8$. The non-overlapping $P{=}16$ row differs from the overlapping one by less than $3$\,pp.

\subsection{Prediction Length Verification}
\label{app:predlen}

We check whether the FFN dominance pattern persists across forecasting horizons by retraining PatchTST with $\text{pred\_len} \in \{192, 336, 720\}$ on 4 datasets (ETTm1, ETTh1, Weather, Electricity) and applying our probing protocol.

\begin{table}[H]
\caption{Cross-horizon check of FFN dominance: constructed-task deltas for PatchTST at 4 forecasting horizons (9-run mean). FFN $\Delta \geq$ Attn $\Delta$ holds at every horizon. Values for pred\_len=96 are from Table~\ref{tab:patchtst_delta}.}
\label{tab:cross_horizon}
\centering
\small
\begin{tabular}{llccc}
\toprule
Dataset & pred\_len & Attn $\Delta$ & FFN $\Delta$ & Ratio \\
\midrule
ETTm1 & 96 & $+13.5$ & $+26.1$ & 1.9$\times$ \\
ETTm1 & 192 & $+15.7$ & $+24.5$ & 1.6$\times$ \\
ETTm1 & 336 & $+13.9$ & $+25.6$ & 1.8$\times$ \\
ETTm1 & 720 & $+13.3$ & $+26.9$ & 2.0$\times$ \\
\midrule
ETTh1 & 96 & $+0.0$ & $+3.8$ & --- \\
ETTh1 & 192 & $-0.1$ & $+4.1$ & --- \\
ETTh1 & 336 & $+0.0$ & $+4.0$ & --- \\
ETTh1 & 720 & $+0.0$ & $+4.3$ & --- \\
\midrule
Weather & 96 & $+16.7$ & $+21.5$ & 1.3$\times$ \\
Weather & 192 & $+16.6$ & $+20.2$ & 1.2$\times$ \\
Weather & 336 & $+16.6$ & $+18.7$ & 1.1$\times$ \\
Weather & 720 & $+15.4$ & $+15.8$ & 1.0$\times$ \\
\midrule
Electricity & 96 & $+11.3$ & $+23.7$ & 2.1$\times$ \\
Electricity & 192 & $+10.0$ & $+23.0$ & 2.3$\times$ \\
Electricity & 336 & $+11.3$ & $+25.1$ & 2.2$\times$ \\
Electricity & 720 & $+10.7$ & $+25.2$ & 2.4$\times$ \\
\bottomrule
\end{tabular}
\end{table}

FFN $\Delta \geq$ Attn $\Delta$ holds in all 16 pairs of dataset and horizon (Table~\ref{tab:cross_horizon}). On ETTm1 the FFN/Attn ratio ranges from 1.6$\times$ to 2.0$\times$ with no systematic trend; on Electricity it stays at 2.1 to 2.4$\times$; on Weather it falls from 1.3$\times$ to 1.0$\times$ as the horizon grows, and at pred\_len=720 the two deltas nearly tie ($+15.8$ against $+15.4$). ETTh1 shows minimal Attn $\Delta$ at all horizons. The FFN's leading role in construction under patch tokens thus extends to longer horizons on these 4 datasets.

\section{Data Quality Verification}
\label{app:dataquality}

\subsection{Quartile Boundary Verification}
\label{app:quartile}

\begin{table}[H]
\caption{Volatility quartile boundaries Q1 to Q3: quartiles of the variances of the 210{,}000 RevIN-normalized 16-point patches in each dataset's PatchTST probe sample (Appendix~\ref{app:setup_details}), that is, the class boundaries of the volatility labels. zero\%: share of these patches with variance below $10^{-6}$. Status: collapsed if two adjacent boundaries coincide, and normal otherwise.}
\label{tab:quartile}
\centering
\small
\begin{tabular}{lccccc}
\toprule
Dataset & Q1 & Q2 & Q3 & zero\% & Status \\
\midrule
ETTm1 & 0.064 & 0.198 & 0.399 & 0.3\% & \checkmark normal \\
ETTm2 & 0.092 & 0.348 & 0.668 & 12.9\% & \checkmark normal \\
ETTh1 & 0.327 & 0.646 & 1.041 & 1.3\% & \checkmark normal \\
ETTh2 & 0.244 & 0.582 & 0.951 & 11.4\% & \checkmark normal \\
Weather & 0.003 & 0.020 & 0.109 & 14.7\% & \checkmark normal \\
Electricity & 0.483 & 0.810 & 1.099 & 0.4\% & \checkmark normal \\
Exchange & 0.023 & 0.054 & 0.121 & 0.0\% & \checkmark normal \\
Traffic & 0.348 & 0.726 & 1.192 & 0.0\% & \checkmark normal \\
\midrule
Solar & 0.000 & 0.000 & 0.137 & 55.4\% & $\times$ collapsed \\
\bottomrule
\end{tabular}
\end{table}

In Solar (Table~\ref{tab:quartile}), Q1 and Q2 are both 0 because 55.4\% of patches (nighttime) have variance $\approx 0$. Quartile classification then effectively reduces to a 3-class problem, and high accuracy is achievable simply by detecting whether the variance is zero. The other 8 datasets in the table have well-separated quartile boundaries, making 4-class classification meaningful.

\subsection{Channel Correlation Verification}

\begin{table}[H]
\caption{Inter-channel Pearson correlation in the same probe samples, computed within each input window between the concatenated patches of two variates and averaged over the sample's first 20 windows (fewer for Electricity and Traffic) and over all pairs of variates. $>$0.5 ratio: share of pairs whose window-averaged correlation exceeds 0.5. Status describes the mean correlation: low below $0.3$, moderate from $0.3$ to $0.5$, high from $0.5$ to $0.8$, and redundant above $0.8$.}
\label{tab:channel_corr}
\centering
\small
\begin{tabular}{lcccc}
\toprule
Dataset & \#Variables & Mean correlation & $>$0.5 ratio & Status \\
\midrule
ETTm1 & 7 & 0.062 & 9.5\% & \checkmark low \\
ETTm2 & 7 & 0.241 & 19.0\% & \checkmark low \\
ETTh1 & 7 & 0.113 & 14.3\% & \checkmark low \\
ETTh2 & 7 & 0.276 & 28.6\% & \checkmark low \\
Weather & 21 & 0.033 & 22.4\% & \checkmark low \\
Electricity & 321 & 0.491 & 55.6\% & $\triangle$ moderate \\
Exchange & 8 & 0.335 & 35.7\% & $\triangle$ moderate \\
Traffic & 862 & 0.593 & 69.5\% & $\triangle$ high \\
\midrule
Solar & 137 & 0.855 & 100.0\% & $\times$ redundant \\
\bottomrule
\end{tabular}
\end{table}

Solar's 137 plants all lie in one state (Alabama;~\citealp{lai2018modeling}), yielding a mean correlation of 0.855 (Table~\ref{tab:channel_corr}). Under channel-independent treatment, Solar thus amounts to nearly the same time series repeated 137 times. Electricity, Exchange, and Traffic also show channel correlation but keep well-separated quartile boundaries, so their probing results remain interpretable.

\subsection{Solar-Energy Exclusion Rationale}

Solar-Energy exhibits both issues (quartile collapse and channel redundancy) and is therefore excluded from the main analysis. Table~\ref{tab:solar_compare} compares the main 8-dataset configuration with a 9-dataset configuration that adds Solar.

\begin{table}[H]
\caption{Effect of adding Solar to the main 8 datasets: Layer-1 deltas and, for CATS cross-attention, absolute accuracy (constructed task average, 9-run mean).}
\label{tab:solar_compare}
\centering
\small
\begin{tabular}{lccc}
\toprule
Metric & 8 datasets (main) & 8 datasets + Solar & Diff \\
\midrule
PatchTST Attn $\Delta$         & $+6.4$  & $+6.5$  & $+0.1$ \\
PatchTST FFN $\Delta$          & $+20.4$ & $+18.6$ & $-1.8$ \\
iTransformer Attn $\Delta$     & $+13.8$ & $+14.2$ & $+0.4$ \\
iTransformer FFN $\Delta$      & $+5.5$  & $+5.7$  & $+0.2$ \\
CATS cross-attn accuracy          & $44.2$  & $47.4$  & $+3.2$ \\
CATS FFN $\Delta$              & $+2.5$  & $+2.6$  & $+0.1$ \\
\bottomrule
\end{tabular}
\end{table}

Adding Solar introduces only minor numerical shifts on most metrics ($\leq 0.4$\,pp), with the largest changes on PatchTST FFN $\Delta$ ($-1.8$\,pp) and CATS cross-attention ($+3.2$\,pp). The architecture-level dominance pattern is preserved (PatchTST: FFN-dominant, iTransformer: attention-dominant, CATS: cross-attention-dominant), so excluding Solar does not change it.

\paragraph{Solar full probing results.} The quartile collapse is directly visible at the embedding layer (Table~\ref{tab:patchtst_full_solar}): volatility ($86.1\%$), diff\_energy ($83.7\%$), and spectral\_entropy ($80.4\%$) are already high on Solar, against $27.7$ to $54.9\%$ on the 8 main datasets (\S\ref{sec:innate_constructed}; e.g., ETTm1 volatility $38.9\%$, Table~\ref{tab:patchtst_full_ettm1}). With only $\sim 12$\,pp gained by \texttt{1b\_ffn} (vs.\ $\sim 40$\,pp on ETTm1), these tasks no longer test construction on Solar: their labels largely separate nighttime patches of near-zero variance from the rest (Appendix~\ref{app:quartile}), so they are decodable at the embedding for a reason specific to the data, not because a linear read-out of the patch determines them (\S\ref{sec:tasks}).

\begin{table}[H]
\caption{PatchTST full layer probing results on Solar (9-run mean $\pm$ std). The constructed-task group (volatility, diff\_energy, spectral\_entropy) starts above $80\%$ at the embedding layer due to Solar's quartile collapse, leaving little room for FFN-driven gains and undermining the constructed/innate dichotomy used in our main analysis. This is the data quality issue that motivates Solar's exclusion (Appendix~\ref{app:quartile}).}
\label{tab:patchtst_full_solar}
\centering
\scriptsize
\setlength{\tabcolsep}{2pt}
\begin{tabular}{lccccccc}
\toprule
Task & embed & 1a\_attn & 1b\_ffn & 2a\_attn & 2b\_ffn & 3a\_attn & 3b\_ffn \\
\midrule
\multicolumn{8}{l}{\textit{Innate and degenerate (classification)}} \\
trend\_direction         & $99.4$\stdv{0.0} & $99.4$\stdv{0.0} & $99.2$\stdv{0.0} & $98.9$\stdv{0.1} & $98.7$\stdv{0.0} & $98.3$\stdv{0.1} & $97.9$\stdv{0.2} \\
dominant\_period         & $99.4$\stdv{0.0} & $99.4$\stdv{0.0} & $99.6$\stdv{0.0} & $99.6$\stdv{0.0} & $99.6$\stdv{0.0} & $99.5$\stdv{0.0} & $99.5$\stdv{0.0} \\
\midrule
\multicolumn{8}{l}{\textit{Constructed (classification)}} \\
volatility$_{q4}$        & $86.1$\stdv{0.1} & $92.8$\stdv{0.5} & $97.0$\stdv{0.2} & $96.9$\stdv{0.2} & $96.6$\stdv{0.1} & $96.1$\stdv{0.2} & $95.6$\stdv{0.3} \\
diff\_energy$_{q4}$      & $83.7$\stdv{0.1} & $89.9$\stdv{0.4} & $95.5$\stdv{0.4} & $95.4$\stdv{0.2} & $95.2$\stdv{0.3} & $94.7$\stdv{0.1} & $94.2$\stdv{0.1} \\
spectral\_entropy$_{q4}$ & $80.4$\stdv{0.3} & $90.3$\stdv{0.7} & $93.1$\stdv{0.7} & $94.9$\stdv{0.5} & $94.8$\stdv{0.5} & $95.1$\stdv{0.4} & $94.2$\stdv{0.4} \\
\midrule
\multicolumn{8}{l}{\textit{Boundary tasks and local\_global\_trend (classification)}} \\
slope\_magnitude$_{q4}$  & $74.5$\stdv{0.3} & $82.4$\stdv{1.0} & $86.5$\stdv{0.3} & $86.9$\stdv{0.4} & $86.7$\stdv{0.4} & $87.0$\stdv{0.2} & $85.7$\stdv{0.3} \\
change\_point            & $88.4$\stdv{0.1} & $93.6$\stdv{0.6} & $97.1$\stdv{0.3} & $96.4$\stdv{0.3} & $96.3$\stdv{0.4} & $95.6$\stdv{0.2} & $95.3$\stdv{0.2} \\
local\_global\_trend     & $99.4$\stdv{0.0} & $99.4$\stdv{0.0} & $99.2$\stdv{0.0} & $98.9$\stdv{0.1} & $98.7$\stdv{0.0} & $98.3$\stdv{0.1} & $97.9$\stdv{0.2} \\
\midrule
\multicolumn{8}{l}{\textit{Inter-token (classification)}} \\
adjacent\_similarity     & $93.6$\stdv{0.1} & $95.2$\stdv{0.4} & $95.4$\stdv{0.2} & $95.5$\stdv{0.1} & $95.2$\stdv{0.2} & $94.8$\stdv{0.4} & $94.2$\stdv{0.2} \\
relative\_magnitude      & $100.0$\stdv{0.0} & $99.9$\stdv{0.0} & $99.9$\stdv{0.0} & $99.8$\stdv{0.0} & $99.7$\stdv{0.1} & $99.5$\stdv{0.1} & $99.4$\stdv{0.0} \\
position\_quarter        & $100.0$\stdv{0.0} & $100.0$\stdv{0.0} & $99.7$\stdv{0.0} & $98.8$\stdv{0.1} & $97.5$\stdv{0.3} & $93.9$\stdv{0.6} & $90.9$\stdv{0.9} \\
trend\_consistency       & $81.3$\stdv{0.1} & $87.5$\stdv{0.2} & $87.9$\stdv{0.3} & $87.4$\stdv{0.4} & $87.0$\stdv{0.3} & $85.3$\stdv{0.2} & $84.2$\stdv{0.6} \\
\midrule
\multicolumn{8}{l}{\textit{Regression ($R^2$)}} \\
reg\_variance            & $0.402$\stdv{0.002} & $0.823$\stdv{0.016} & $0.956$\stdv{0.003} & $0.950$\stdv{0.006} & $0.949$\stdv{0.004} & $0.934$\stdv{0.006} & $0.933$\stdv{0.008} \\
reg\_slope               & $1.000$\stdv{0.000} & $1.000$\stdv{0.000} & $0.999$\stdv{0.000} & $0.997$\stdv{0.000} & $0.995$\stdv{0.001} & $0.989$\stdv{0.001} & $0.984$\stdv{0.002} \\
reg\_next\_mean         & $0.976$\stdv{0.000} & $0.979$\stdv{0.000} & $0.979$\stdv{0.000} & $0.978$\stdv{0.000} & $0.978$\stdv{0.001} & $0.975$\stdv{0.001} & $0.975$\stdv{0.001} \\
\midrule
\multicolumn{8}{l}{\textit{Control}} \\
random\_label            & $25.3$\stdv{0.2} & $25.3$\stdv{0.2} & $25.3$\stdv{0.1} & $25.4$\stdv{0.2} & $25.5$\stdv{0.2} & $25.4$\stdv{0.3} & $25.4$\stdv{0.2} \\
\bottomrule
\end{tabular}
\end{table}

\section{ModernTCN Component Breakdown}
\label{app:moderntcn_breakdown}

We also probe ModernTCN~\citep{luo2024moderntcn}, a CNN-based time series model with structure Patch Embed $\to$ [DWConv $+$ ConvFFN1 $+$ ConvFFN2] $\times L$, where DWConv performs linear inter-patch mixing while the two ConvFFNs perform nonlinear processing. We evaluate on ETTm1 using ModernTCN's official 3-layer configuration ($L=3$); a 1-layer configuration cannot, by construction, show cancellation across layers.

ModernTCN's block decomposes into three components per layer: DWConv (linear inter-patch mixing), ConvFFN1 (cross-feature mixing, nonlinear), and ConvFFN2 (cross-variable mixing, nonlinear). Table~\ref{tab:moderntcn_breakdown} reports per-component cumulative deltas across all three layers on ETTm1.

\begin{table}[H]
\caption{ModernTCN per-component cumulative deltas across all three layers. ConvFFN1 contributes most of the nonlinear feature construction; ConvFFN2 contributes additionally; DWConv shows sign reversal across layers (positive at Layer 1, negative at Layers 2 and 3) with net cancellation when summed.}
\label{tab:moderntcn_breakdown}
\centering
\small
\begin{tabular}{lcccc}
\toprule
Task & Total DWConv $\Delta$ & Total ConvFFN1 $\Delta$ & Total ConvFFN2 $\Delta$ & Total ConvFFN $\Delta$ \\
\midrule
volatility & $-3.1$ & $+11.7$ & $+4.1$ & $+15.8$ \\
diff\_energy & $+2.2$ & $+23.9$ & $+2.3$ & $+26.2$ \\
spectral\_entropy & $-0.1$ & $+15.8$ & $+5.9$ & $+21.7$ \\
\midrule
\textbf{Constructed avg} & $\mathbf{-0.3}$ & $\mathbf{+17.1}$ & $\mathbf{+4.1}$ & $\mathbf{+21.2}$ \\
\bottomrule
\end{tabular}
\end{table}

Per-layer DWConv contributions are $+13.8$, $-7.9$, $-6.2$ (Layers 1, 2, and 3, respectively), illustrating the sign reversal and net cancellation: DWConv's gain at Layer 1 is offset by its negative contributions at Layers 2 and 3. ConvFFN1 contributions are stable across layers ($+6.5$, $+5.5$, $+5.1$), and ConvFFN2 adds a smaller but consistent increment ($+1.0$, $+1.9$, $+1.2$). The linear DWConv leads in the first block, which is where the main analysis compares components, but the ConvFFNs lead from the second block on and carry most of the construction summed over the three blocks, as the FFN does in PatchTST (\S\ref{sec:unified}).

\section{Probe Convergence and Expressiveness}
\label{app:convergence}

\subsection{Convergence}

\begin{table}[H]
\caption{Probing accuracy by gradient steps on ETTm1 (top) and the FFN delta minus the attention delta on three more datasets (bottom; positive: the FFN leads), for PatchTST with the three model seeds of the main analysis (9-run means). The probes were retrained for this table, and their random draws differ from those behind Table~\ref{tab:patchtst_full_ettm1}, so the ETTm1 30k column differs from it by up to $0.6$\,pp at embed and $0.1$\,pp after attention and the FFN. The component that leads on each task is the same at every step count on ETTm1, and from 5k on elsewhere.}
\label{tab:convergence}
\centering
\small
\begin{tabular}{llccccccc}
\toprule
Layer & Task & 1k & 5k & 10k & 20k & 30k & 50k & 80k \\
\midrule
embed & volatility & 38.6 & 38.7 & 38.7 & 38.7 & 38.7 & 38.7 & 38.7 \\
1a\_attn & volatility & 43.9 & 46.8 & 47.9 & 48.7 & 49.2 & 49.7 & 50.2 \\
1b\_ffn & volatility & 70.2 & 79.7 & 81.6 & 83.0 & 83.8 & 84.6 & 85.4 \\
\midrule
embed & diff\_energy & 41.4 & 41.5 & 41.5 & 41.5 & 41.5 & 41.5 & 41.5 \\
1a\_attn & diff\_energy & 49.6 & 52.8 & 54.1 & 55.1 & 55.7 & 56.5 & 57.1 \\
1b\_ffn & diff\_energy & 54.6 & 61.2 & 63.5 & 65.3 & 66.2 & 67.2 & 68.0 \\
\midrule
embed & spectral\_entropy & 28.4 & 29.8 & 30.0 & 30.2 & 30.2 & 30.3 & 30.4 \\
1a\_attn & spectral\_entropy & 41.5 & 44.4 & 45.5 & 46.5 & 47.1 & 47.8 & 48.4 \\
1b\_ffn & spectral\_entropy & 68.2 & 75.9 & 77.9 & 79.4 & 80.1 & 80.9 & 81.5 \\
\midrule
\multicolumn{9}{l}{\textit{FFN $\Delta$ minus Attn $\Delta$ (pp)}} \\
ETTh1 & volatility & $+0.8$ & $+2.3$ & $+2.8$ & $+3.0$ & $+3.0$ & $+3.0$ & $+3.0$ \\
ETTh1 & diff\_energy & $+2.0$ & $+3.0$ & $+3.1$ & $+3.1$ & $+3.1$ & $+3.1$ & $+3.1$ \\
ETTh1 & spectral\_entropy & $+1.2$ & $+4.3$ & $+5.2$ & $+5.3$ & $+5.2$ & $+5.1$ & $+5.0$ \\
Weather & volatility & $-1.4$ & $+2.8$ & $+4.3$ & $+5.7$ & $+6.3$ & $+6.9$ & $+7.4$ \\
Weather & diff\_energy & $-5.8$ & $-5.3$ & $-5.0$ & $-4.9$ & $-4.8$ & $-4.4$ & $-4.7$ \\
Weather & spectral\_entropy & $+2.7$ & $+6.8$ & $+9.4$ & $+11.2$ & $+11.9$ & $+12.7$ & $+13.6$ \\
Electricity & volatility & $+17.7$ & $+18.7$ & $+18.7$ & $+18.2$ & $+17.9$ & $+17.6$ & $+17.4$ \\
Electricity & diff\_energy & $+13.0$ & $+14.8$ & $+15.2$ & $+15.1$ & $+15.0$ & $+14.9$ & $+14.7$ \\
Electricity & spectral\_entropy & $+4.5$ & $+5.0$ & $+4.8$ & $+4.7$ & $+4.5$ & $+4.3$ & $+4.2$ \\
\bottomrule
\end{tabular}
\end{table}

On ETTm1, embed accuracy changes by at most $0.6$\,pp after 5k steps (Table~\ref{tab:convergence}, top). FFN-layer accuracy increases modestly even at 80k steps ($+1.6$, $+1.8$, and $+1.4$\,pp on volatility, diff\_energy, and spectral\_entropy from 30k to 80k), which we attribute to the standard logistic-regression dynamic of weight norms increasing under cross-entropy loss without weight decay. The component that leads on each task is the same at every step count (the FFN on volatility and spectral\_entropy, attention on diff\_energy). From 30k to 80k, the FFN/Attn $\Delta$ ratio falls by only 5 to 7\% (volatility: $3.3\times$ vs.\ $3.1\times$; diff\_energy: $0.74\times$ vs.\ $0.70\times$; spectral\_entropy: $2.0\times$ vs.\ $1.8\times$), so on ETTm1 the conclusion on component dominance does not depend on the number of training steps. On ETTh1, Weather, and Electricity, the leading component of each task is also the same at every step count from 5k on, and the FFN-layer accuracy, not shown in Table~\ref{tab:convergence}, rises by at most $3.0$\,pp from 30k to 80k. We adopt 30k steps as a balance between near-saturation and computational cost.

\subsection{Linear Probe vs. MLP Probe}
\label{app:mlp_probe}

Because a linear probe can miss nonlinearly encoded information, we compare it with a 2-layer MLP probe (hidden width 128, ReLU) on identical representations across 4 datasets (ETTm1, ETTh1, Weather, Electricity) and four tasks (three constructed: volatility, diff\_energy, spectral\_entropy; one innate: trend\_direction). The MLP probe is matched to the linear probes: it is trained and scored as in \S\ref{sec:protocol}, with the same split for each probe seed, Adam at the same learning rate, the same batch size and number of steps, and the best test accuracy over evaluations every 500 steps. Each cell of Table~\ref{tab:mlp_probe} is a 9-run mean.

\begin{table}[H]
\caption{Linear vs.\ MLP probe comparison on PatchTST across 4 datasets and four tasks (9-run mean). MLP probes saturate at the embed layer for constructed tasks; linear probes gain at the FFN instead. Linear probes measure what is linearly decodable, the format a linear head reads, while MLP probes measure information present in some form.}
\label{tab:mlp_probe}
\centering
\scriptsize
\setlength{\tabcolsep}{2.5pt}
\begin{tabular}{llcccccccc}
\toprule
& & \multicolumn{2}{c}{ETTm1} & \multicolumn{2}{c}{ETTh1} & \multicolumn{2}{c}{Weather} & \multicolumn{2}{c}{Electricity} \\
\cmidrule(lr){3-4} \cmidrule(lr){5-6} \cmidrule(lr){7-8} \cmidrule(lr){9-10}
Task & Probe & embed & FFN $\Delta$ & embed & FFN $\Delta$ & embed & FFN $\Delta$ & embed & FFN $\Delta$ \\
\midrule
\multirow{2}{*}{volatility} & Linear & 38.9 & $+34.5$ & 37.6 & $+2.9$ & 33.6 & $+24.0$ & 41.6 & $+30.5$ \\
& MLP & \textbf{99.0} & $-0.2$ & \textbf{95.5} & $-0.4$ & \textbf{99.1} & $-0.3$ & \textbf{97.5} & $+0.2$ \\
\midrule
\multirow{2}{*}{diff\_energy} & Linear & 42.1 & $+10.6$ & 38.7 & $+3.1$ & 33.8 & $+14.8$ & 38.4 & $+26.4$ \\
& MLP & \textbf{98.6} & $-0.2$ & \textbf{92.9} & $-1.4$ & \textbf{97.5} & $-0.8$ & \textbf{96.4} & $+0.5$ \\
\midrule
\multirow{2}{*}{spectral\_entropy} & Linear & 30.2 & $+33.2$ & 35.8 & $+5.4$ & 36.2 & $+25.8$ & 41.6 & $+14.3$ \\
& MLP & \textbf{98.5} & $-0.3$ & \textbf{91.0} & $-1.4$ & \textbf{98.7} & $-0.4$ & \textbf{96.0} & $-0.1$ \\
\midrule
\multirow{2}{*}{trend\_direction} & Linear & 99.6 & $-0.2$ & 97.9 & $-1.4$ & 98.5 & $-0.3$ & 99.6 & $-0.4$ \\
& MLP & 100.0 & $0.0$ & 99.6 & $-0.1$ & 99.9 & $-0.1$ & 99.9 & $-0.1$ \\
\bottomrule
\end{tabular}
\end{table}

\textbf{Observation.} On constructed tasks (volatility, diff\_energy, spectral\_entropy), MLP probes saturate at the embed layer across all 4 datasets, achieving 91 to 99\% with negligible change through attn/FFN ($|\Delta| \leq 1.4$\,pp). Linear probes, in contrast, start at about 30 to 42\% (chance 25\%) and gain $+2.9$ to $+34.5$\,pp at the FFN, least on ETTh1. On the innate task (trend\_direction), both probes saturate near ceiling at the embed layer.

\textbf{Interpretation.} PatchTST's patch embedding $W_P$ is a full-rank linear projection from the patch dimension (16) to $d_{\text{model}}$ (128, or 16 on ETTh1). MLP probes can effectively invert this projection to recover raw patches and reimplement task formulas (variance, diff energy, spectral entropy) directly. Hence, MLP probing accuracy reflects where information exists in some form, not where it is linearly decodable.

\textbf{Linear probes are appropriate.} PatchTST's forecasting head, like those of its variants and ModernTCN, is the linear projection $\text{Linear}(d_{\text{model}} \cdot N_{\text{patches}}, \text{pred\_len})$. iTransformer and CATS map each token linearly to its part of the forecast (a variate's horizon, a query's future patch); only Informer reads its encoder output with a nonlinear decoder (Appendix~\ref{app:informer}). Linear decodability is thus the format these heads read, and it is what our linear probes measure. The linear-probe deltas at the FFN show that the FFN re-encodes raw patch information into a linearly decodable form, the property relevant to our central question.

\section{Selectivity Analysis}
\label{app:selectivity}

We compute the selectivity of \citet{hewitt2019designing} on the same probing data used in \S\ref{sec:results_start}, with a random-label control in place of their word-type control tasks, which give each word type a fixed random label. For each (dataset, layer, task) triple we report
\[
  \mathrm{selectivity}(d, \ell, t)
  \;=\; \mathrm{acc}(d, \ell, t) \;-\; \mathrm{acc}_{\mathrm{ctrl}}(d, \ell),
\]
where $\mathrm{acc}_{\mathrm{ctrl}}$ is the accuracy of the same probe trained on a 4-class \emph{random-label} control task. All values are 9-run means and aggregated across the 8 datasets used in the main analysis. The control task stays near chance at every probing point of the 4 models analyzed here (PatchTST, PatchTST-NoAttn, iTransformer, and CATS; Table~\ref{tab:random-label-control}). Its labels are drawn independently for each token and carry nothing a probe could transfer to held-out tokens, so the control stays at chance by construction and marks the chance floor at each probing point. Unlike word-type control tasks, it does not measure probe capacity, which we fix instead by restricting the read-out to the forecasting heads' linear family (a matched MLP probe saturates at the embedding; Appendix~\ref{app:mlp_probe}). Throughout this section, we write $\Delta_a$ for the delta from the embedding to the output of the first sub-layer (attention or cross-attention) and $\Delta_b$ for the delta from there to the FFN output.

\paragraph{PatchTST Layer-1 selectivity.}
The three constructed tasks gain on average $+26.70$\,pp through Layer 1 (paired $t$-test pooled over the 24 dataset-task pairs, $p < 10^{-6}$, Cohen's $d_z = +1.67$, very large), with the FFN sub-layer contributing $+20.33$\,pp ($d_z = +1.59$) versus attention's $+6.37$\,pp ($d_z = +0.89$). The three tasks of a dataset are probed on the same forecasters, so these pairs are not independent; a test on the 8 dataset means gives $p < 10^{-3}$. Through Layer 1, the innate trend\_direction changes by $-1.2$\,pp, the degenerate dominant\_period by $+1.1$\,pp, and local\_global\_trend (outside the classification) by $+0.6$\,pp. In contrast, the boundary tasks slope\_magnitude$_{q4}$ and change\_point gain $+28.5$ and $+16.4$\,pp, mainly at the FFN, as \S\ref{sec:tasks} notes for ETTm1.

\begin{table}[H]
\caption{PatchTST Layer-1 selectivity (8-dataset mean; 9-run probing). Columns: selectivity at embedding, post-attention, and post-FFN; total Layer-1 delta; Cohen's $d_z$ and one-sided $p$ (for an increase) of a paired $t$-test on $\mathrm{embed}\to 1\mathrm{b\_ffn}$ over the 9-run means of the 8 datasets (task rows) or of the 24 dataset-task pairs (Constructed avg row). Constructed and boundary tasks show very large selectivity gains, mainly at the FFN, while the other tasks change by at most $1.2$\,pp.}
\label{tab:selectivity_patchtst_l1}
\centering
\small
\begin{tabular}{lrrrrrr}
\toprule
Task & embed & 1a\_attn & 1b\_ffn & Total $\Delta$ & Cohen's $d_z$ & $p$ \\
\midrule
\multicolumn{7}{l}{\textit{Constructed}} \\
volatility$_{q4}$        & $+13.1$ & $+19.2$ & $+43.4$ & $+30.3$ & $+1.58$ & $0.0014$ \\
diff\_energy$_{q4}$      & $+12.4$ & $+20.0$ & $+36.7$ & $+24.3$ & $+1.69$ & $0.0010$ \\
spectral\_entropy$_{q4}$ & $+11.2$ & $+16.6$ & $+36.7$ & $+25.5$ & $+1.65$ & $0.0011$ \\
\textbf{Constructed avg} & $\mathbf{+12.2}$ & $\mathbf{+18.6}$ & $\mathbf{+38.9}$ & $\mathbf{+26.7}$ & $\mathbf{+1.67}$ & $\mathbf{<10^{-6}}$ \\
\midrule
\multicolumn{7}{l}{\textit{Innate and degenerate}} \\
trend\_direction         & $+72.8$ & $+72.3$ & $+71.6$ & $-1.2$  & $-0.80$ & $0.97$ \\
dominant\_period         & $+68.3$ & $+68.4$ & $+69.4$ & $+1.1$  & $+0.91$ & $0.019$ \\
\midrule
\multicolumn{7}{l}{\textit{Boundary tasks and local\_global\_trend}} \\
slope\_magnitude$_{q4}$  & $+10.3$ & $+16.9$ & $+38.9$ & $+28.5$ & $+1.55$ & $0.0016$ \\
change\_point            & $+46.1$ & $+50.0$ & $+62.5$ & $+16.4$ & $+1.62$ & $0.0013$ \\
local\_global\_trend     & $+36.3$ & $+37.0$ & $+36.9$ & $+0.6$  & $+0.98$ & $0.014$ \\
\bottomrule
\end{tabular}
\end{table}

\paragraph{Selectivity tracks raw accuracy.}
For the 8-dataset means of all 4 models, the maximum gap between the selectivity-based delta and the raw-accuracy-based delta was $0.05$\,pp before rounding (up to $0.63$\,pp on a single dataset). For PatchTST constructed tasks, $\mathrm{embed}\to 1\mathrm{b\_ffn}$ selectivity increased by $+26.70$\,pp while raw accuracy increased by $+26.71$\,pp. This near-equality follows directly from the flatness of $\mathrm{acc}_{\mathrm{ctrl}}$ across layers ($25.30$ to $25.33$ for PatchTST), which the random-label design guarantees.

\paragraph{Cross-architecture validation.}
The architecture-specific dominance pattern reported in Figure~\ref{fig:component_dominance} (FFN-dominant in PatchTST, attention-dominant in iTransformer, cross-attention-dominant in CATS) is reproduced under the selectivity metric (Table~\ref{tab:selectivity_cross_arch}; for CATS, by the query-side trajectory of Table~\ref{tab:selectivity_cats_qside}).

\begin{table}[H]
\caption{Cross-architecture Layer-1 selectivity vs.\ raw accuracy (constructed task average; 8-dataset mean). The dominant sub-layer (\textbf{bold} in each row but CATS) and its sign are reproduced under selectivity. For PatchTST-NoAttn, $\Delta_a$ is near zero because the attention sub-layer is ablated.}
\label{tab:selectivity_cross_arch}
\centering
\small
\begin{tabular}{llrrrr}
\toprule
Architecture & First sub-layer & $\Delta_a$ (sel) & $\Delta_b$ (sel) & $\Delta_a$ (raw) & $\Delta_b$ (raw) \\
\midrule
PatchTST          & attn          & $+6.4$  & $\mathbf{+20.3}$ & $+6.4$  & $+20.4$ \\
PatchTST-NoAttn   & attn (ablated) & $-0.1$  & $\mathbf{+27.5}$ & $-0.1$  & $+27.4$ \\
iTransformer      & attn          & $\mathbf{+13.8}$ & $+5.5$ & $+13.8$ & $+5.5$ \\
CATS$^{\dagger}$  & cross         & $-9.9^{\dagger}$ & $+2.5$ & $-9.9$  & $+2.5$ \\
\bottomrule
\end{tabular}
\\[2pt]
\begin{minipage}{\linewidth}
\footnotesize
$^{\dagger}$ \textbf{CATS side-switch caveat.} CATS \texttt{embed} probes the KV-side (key and value) input embedding, while \texttt{1a\_cross} and subsequent layers probe the Q-side post-cross-attention representation (a learnable query attending to KV). Their direct subtraction therefore reflects a side switch rather than a layer transition, and the labels also change from the input patches to the future patches the queries forecast (the KV-side labels are computed on the input padded with zeros, whereas the model pads it by repeating its last value). The negative $\Delta_a$ is thus not information loss. Table~\ref{tab:selectivity_cats_qside} shows the within-Q trajectory: the first cross-attention output reaches $+18.85$\,pp of constructed-task selectivity, measured against chance rather than the query before cross-attention, while the subsequent five sub-layers add only $+4.03$\,pp cumulatively.
\end{minipage}
\end{table}

\begin{table}[H]
\caption{CATS within-Q selectivity trajectory (constructed task average, 8-dataset mean). Most of the Q-side selectivity is already present at the first cross-attention output.}
\label{tab:selectivity_cats_qside}
\centering
\small
\begin{tabular}{lrr}
\toprule
Q-side layer & Selectivity & Sub-layer $\Delta$ \\
\midrule
1a\_cross & $+18.85$ & --- (start; cross-attention output) \\
1b\_ffn   & $+21.39$ & $+2.54$ \\
2a\_cross & $+21.97$ & $+0.58$ \\
2b\_ffn   & $+22.65$ & $+0.68$ \\
3a\_cross & $+22.69$ & $+0.04$ \\
3b\_ffn   & $+22.88$ & $+0.19$ \\
\midrule
\textit{Total post-1a\_cross gain} & & \textit{$+4.03$ over 5 sub-layers} \\
\bottomrule
\end{tabular}
\end{table}

\paragraph{The boundary tasks.}
The FFN gives most of the Layer-1 gain of both boundary tasks in Table~\ref{tab:selectivity_patchtst_l1}: $+22.0$ of $+28.5$\,pp for slope\_magnitude$_{q4}$ and $+12.5$ of $+16.4$\,pp for change\_point. A linear read-out cannot realize an absolute value, but their labels only take the absolute value of a linear quantity that the embedding already exposes (the slope, or the difference between the two half means), so \S\ref{sec:tasks} reports them as boundary tasks rather than main tasks.

\section{Activation Patching}
\label{app:patching}

To complement the representational analysis (\S\ref{sec:ffn_dominance}) on the forecasting side, we apply activation patching~\citep{meng2022locating} to PatchTST on all 8 datasets, and additionally to PatchTST at $P{=}1$ and to iTransformer (Appendices~\ref{app:patching_p1} and~\ref{app:patching_itrans}). Restoring a clean component output into a corrupted run is a denoising test: it checks whether that output suffices to restore the forecast~\citep{heimersheim2024use}, not whether the component is necessary.

\subsection{Methodology}

\paragraph{Procedure.} Following~\citet{meng2022locating}, for each input we run two forward passes: a \emph{clean} run on the original input, and a \emph{corrupt} run where the input is replaced by a corrupted version. We then run a third \emph{patched} pass: the model receives the corrupt input, but we replace the output of a chosen component (e.g., the Layer-1 FFN) with its activation from the clean run. The patching effect is

\begin{equation*}
\text{effect}(c) = \text{MSE}_{\text{corrupt}} - \text{MSE}_{\text{patched}(c)},
\end{equation*}

where $c$ is the patched component. A positive effect means that injecting the clean component output into the corrupt run \emph{recovers} forecasting performance, that is, the clean output of $c$ suffices to restore part of the forecast. Because the FFN's input is the normalized sum of the residual stream and the attention output, a restored FFN output also carries the part of clean attention that passes through the FFN.

\paragraph{Corruption protocol.} Per-sample non-overlapping chunk shuffle: the input time axis is divided into 21 chunks of size 16, and each sample receives an independent permutation of its chunks. This preserves the per-sample marginal distribution exactly while disrupting temporal structure.

\paragraph{Hook position.} Hooks are placed at \texttt{dropout\_attn} and \texttt{dropout\_ffn} (pre-residual component outputs), not at post-residual norm layers. With post-residual hooks, $L_i\_\text{attn}$ and $L_i\_\text{ffn}$ collapse to identical effects in eval mode (once the post-attention stream is restored, the deterministic FFN and normalization reproduce the clean stream downstream); pre-residual hooks correctly separate the two components.

\paragraph{Why focus on Layer 1.} Two reasons: (i) probing places most construction at Layer 1 (Table~\ref{tab:p1_depth}), so Layer 1 is where a forecasting-side check of attention's contribution is most direct; (ii) deeper-layer patching is subject to \emph{cross-context incompatibility}: the clean output enters a corrupt residual stream that has had more layers to diverge from the clean one, so the patched activation can be off-distribution downstream. \citet{zhang2024best} raise this concern for any intervention that introduces out-of-distribution inputs.

\paragraph{Experimental scale.} In every patching experiment, each dataset contributes its first 100 test windows (consecutive, one step apart), and the same windows with the same corruption are used for each of the 3 model seeds, giving 300 paired comparisons per (dataset, component) cell. The main analysis thus has 8 datasets $\times$ 2 Layer-1 components = 16 cells, or 4{,}800 patched runs, and an additional 9{,}600 runs cover Layers 2 and 3 (Appendix~\ref{app:patching_multilayer}). Because each window enters a cell three times, the paired $t$-tests average each window's FFN-minus-attention difference over the 3 seeds and test the 100 seed-averaged windows; their $p$-values still treat these overlapping windows as independent. Cohen's $d_z$ is computed over the 300 pairs.

\subsection{Layer-1 Results}

\begin{table}[H]
\caption{Layer-1 activation patching effects on PatchTST (MSE recovered; mean over 300 paired samples per cell, 3 seeds $\times$ 100). Damage: MSE of the corrupt run minus that of the clean run, averaged over the same 300 samples. Bold marks the larger component within each dataset. Cohen's $d_z$ is the effect size of the FFN-minus-attention difference over the 300 pairs; the paired $t$-test on the 100 seed-averaged windows gives $p < 10^{-16}$ on every dataset.}
\label{tab:patching_l1}
\centering
\small
\begin{tabular}{lccccc}
\toprule
Dataset & Damage & L1\_attn & L1\_ffn & $\Delta$ (FFN $-$ attn) & Cohen's $d_z$ \\
\midrule
ETTm1 & $1.129$ & $+0.229$ & $\mathbf{+0.473}$ & $\mathbf{+0.243}$ & $\mathbf{+2.10}$ \\
ETTm2 & $0.100$ & $+0.022$ & $\mathbf{+0.042}$ & $\mathbf{+0.021}$ & $\mathbf{+1.64}$ \\
ETTh1 & $0.637$ & $+0.007$ & $\mathbf{+0.202}$ & $\mathbf{+0.194}$ & $\mathbf{+3.09}$ \\
ETTh2 & $0.089$ & $+0.000$ & $\mathbf{+0.038}$ & $\mathbf{+0.037}$ & $\mathbf{+1.07}$ \\
Weather & $0.112$ & $\mathbf{+0.036}$ & $+0.030$ & $-0.006$ & $-0.85$ \\
Electricity & $0.691$ & $+0.058$ & $\mathbf{+0.085}$ & $\mathbf{+0.027}$ & $\mathbf{+1.35}$ \\
Exchange & $0.183$ & $+0.005$ & $\mathbf{+0.075}$ & $\mathbf{+0.069}$ & $\mathbf{+1.08}$ \\
Traffic & $1.071$ & $+0.108$ & $\mathbf{+0.262}$ & $\mathbf{+0.153}$ & $\mathbf{+3.30}$ \\
\bottomrule
\end{tabular}
\end{table}

\paragraph{FFN restoration exceeds attention restoration on 7 of 8 datasets.} On every dataset except Weather, restoring the Layer-1 FFN recovers more forecasting performance than restoring Layer-1 attention ($d_z = 1.07$ to $3.30$). Because a restored FFN output is computed from the clean attention output as well (see Procedure), the instrument favors FFN restoration, which carries little signal by itself; only attention's recovery is informative. On ETTh1, for example, attention recovers $+0.007$ against the FFN's $+0.202$.

\paragraph{Agreement with probing on whether attention contributes.} Restoring attention recovers at most $0.022$ on the 4 datasets where probing puts Attn $\Delta$ near zero (ETTh1, ETTh2, ETTm2, Exchange; Table~\ref{tab:patchtst_delta}) and $0.036$ to $0.229$ on the other 4. The agreement is binary (near zero or not) and holds only in absolute MSE: relative to the damage (Table~\ref{tab:patching_l1}), attention recovers at most $3\%$ on ETTh1, ETTh2, and Exchange but $22\%$ on ETTm2, about as much as on ETTm1 ($20\%$) and more than on Electricity ($8\%$) or Traffic ($10\%$), so the recovered shares do not follow the ordering of the probing deltas.

\paragraph{Weather.} On Weather, attention restoration is slightly but significantly larger at Layer 1 ($0.036$ vs.\ $0.030$, $d_z = -0.85$), and the FFN leads at Layers 2 and 3 (Layer 3: $0.053$ vs.\ $0.044$). Weather also has the weakest FFN/attention probing ratio and favors attention on two additional probing targets (\S\ref{sec:ffn_dominance}, Appendix~\ref{app:new_targets}), so it recurs as an exception across instruments; we do not attribute this to a specific data property.

\subsection{Multi-Layer Results}
\label{app:patching_multilayer}

The Layer-1 conclusions largely extend to Layer 2 (Table~\ref{tab:patching_multilayer}); Layer 3 introduces cross-context incompatibility. On Traffic, attention restores more than the FFN at Layer 3 ($+0.167$ vs.\ $+0.035$).

\begin{table}[H]
\caption{Per-layer patching effects on PatchTST across all encoder layers (MSE recovered; mean across 300 paired comparisons). Bold marks the per-layer dominant component (largest positive effect).}
\label{tab:patching_multilayer}
\centering
\small
\setlength{\tabcolsep}{3pt}
\begin{tabular}{lcccccccc}
\toprule
Component & ETTm1 & ETTm2 & ETTh1 & ETTh2 & Weather & Electricity & Exchange & Traffic \\
\midrule
L1\_attn & $+0.229$ & $+0.022$ & $+0.007$ & $+0.000$ & $\mathbf{+0.036}$ & $+0.058$ & $+0.005$ & $+0.108$ \\
L1\_ffn & $\mathbf{+0.473}$ & $\mathbf{+0.042}$ & $\mathbf{+0.202}$ & $\mathbf{+0.038}$ & $+0.030$ & $\mathbf{+0.085}$ & $\mathbf{+0.075}$ & $\mathbf{+0.262}$ \\
\midrule
L2\_attn & $+0.177$ & $+0.028$ & $+0.019$ & $+0.002$ & $+0.039$ & $\mathbf{+0.040}$ & $+0.016$ & $+0.122$ \\
L2\_ffn & $\mathbf{+0.409}$ & $\mathbf{+0.037}$ & $\mathbf{+0.197}$ & $\mathbf{+0.041}$ & $\mathbf{+0.040}$ & $+0.039$ & $\mathbf{+0.079}$ & $\mathbf{+0.180}$ \\
\midrule
L3\_attn & $+0.187$ & $+0.029$ & $+0.035$ & $+0.006$ & $+0.044$ & $\mathbf{+0.049}$ & $+0.036$ & $\mathbf{+0.167}$ \\
L3\_ffn & $\mathbf{+0.356}$ & $\mathbf{+0.037}$ & $\mathbf{+0.147}$ & $\mathbf{+0.036}$ & $\mathbf{+0.053}$ & $-0.020$ & $\mathbf{+0.075}$ & $+0.035$ \\
\bottomrule
\end{tabular}
\end{table}

\paragraph{Cross-context incompatibility in deeper layers.} At Layer 3, a patched clean output enters a corrupt residual stream that has already passed through two layers, so it can lower forecasting performance instead of restoring it, as on Electricity's $L3\_\text{ffn}$ ($-0.020$). The main claims therefore rest on Layer 1, where the clean and corrupt streams have diverged least.

\subsection{PatchTST at $P{=}1$}
\label{app:patching_p1}

We repeat Layer-1 patching on PatchTST at $P{=}1$ for ETTh1 and ETTm1 with the same corruption. Two sanity checks (hooks off, and clean activations patched into the clean run) give zero difference, and the same code reproduces the $P{=}16$ ETTh1 values of Table~\ref{tab:patching_l1}. Restoration stays FFN-dominant at $P{=}1$ (Table~\ref{tab:patching_p1}), although probing puts construction in attention there (\S\ref{sec:pointwise}). Attention's restoration rises relative to the FFN's from $P{=}16$ to $P{=}1$. This instrument favors the FFN by construction and cannot confirm or refute the shift seen in probing.

\begin{table}[H]
\caption{Layer-1 activation patching at $P{=}16$ and $P{=}1$ (MSE recovered; mean over 300 paired samples, 3 seeds $\times$ 100). $d_z$ is the paired effect size of FFN minus attention at $P{=}1$ ($p < 10^{-45}$ on both datasets for the 100 seed-averaged windows).}
\label{tab:patching_p1}
\centering
\small
\begin{tabular}{lccccc}
\toprule
 & \multicolumn{2}{c}{$P{=}16$} & \multicolumn{3}{c}{$P{=}1$} \\
\cmidrule(lr){2-3}\cmidrule(lr){4-6}
Dataset & L1\_attn & L1\_ffn & L1\_attn & L1\_ffn & $d_z$ \\
\midrule
ETTh1 & $+0.007$ & $+0.202$ & $+0.065$ & $+0.269$ & $+2.68$ \\
ETTm1 & $+0.229$ & $+0.473$ & $+0.267$ & $+0.443$ & $+2.08$ \\
\bottomrule
\end{tabular}
\end{table}

\subsection{iTransformer}
\label{app:patching_itrans}

We also patch Layer 1 of iTransformer under two corruptions: the temporal chunk shuffle (6 chunks of 16 steps over the 96-step input) and a cross-variate corruption that circularly shifts each variate by an independent random offset of 8 to 88 steps, which destroys alignment across variates while keeping each variate's values. Both sanity checks give zero difference. Restoration is FFN-dominant for every dataset and corruption tested (Table~\ref{tab:patching_itrans}), although probing puts construction in attention for iTransformer (\S\ref{sec:role_separation}). In iTransformer's post-norm layer, too, the clean FFN output is computed from the clean attention output, so patching favors the FFN by construction. The circular shift also misaligns each variate with its own forecast target. We therefore attribute this result to the instrument and use component knockout for the forecasting-side comparison across architectures (Appendix~\ref{app:knockout}).

\begin{table}[H]
\caption{iTransformer Layer-1 activation patching (MSE recovered; mean over 300 paired samples, 3 seeds $\times$ 100). $d_z$ is the paired effect size of FFN minus attention ($p < 10^{-18}$ in every row for the 100 seed-averaged windows).}
\label{tab:patching_itrans}
\centering
\small
\begin{tabular}{llccc}
\toprule
Corruption & Dataset & L1\_attn & L1\_ffn & $d_z$ \\
\midrule
Temporal & ETTh1 & $+0.073$ & $+0.408$ & $+1.74$ \\
Temporal & Electricity & $+0.045$ & $+0.133$ & $+0.77$ \\
Cross-variate & ETTh1 & $+0.039$ & $+0.401$ & $+2.77$ \\
Cross-variate & Electricity & $+0.011$ & $+0.090$ & $+2.97$ \\
Cross-variate & Traffic & $+0.011$ & $+0.096$ & $+1.29$ \\
\bottomrule
\end{tabular}
\end{table}

\subsection{Limitations of the Patching Analysis}

In PatchTST at Layer 1, patching agrees with probing on whether attention contributes (8 datasets); the larger FFN restoration on 7 of 8 datasets reflects the instrument's bias toward the FFN and does not by itself locate construction. Activation patching gives component-level, forecasting-side evidence; finer mechanistic validation would require path patching~\citep{wang2023interpretability} across components, head-level decomposition, and interventions designed for each architecture.

\section{Forecasting Without a Component}
\label{app:noattn_mse}
\label{app:knockout}

We evaluate forecasting MSE on all 8 datasets, comparing the trained PatchTST checkpoint with two models without attention: the same checkpoint with its attention output set to zero (ablated) and PatchTST-NoAttn trained from scratch (Table~\ref{tab:noattn_mse}). Ablation raises MSE on all 8 datasets, with a mean increase of $+15.7\%$ over the 24 runs and $+14.4\%$ on the 8-dataset mean MSE; the increase is near zero on ETTh2 ($+0.3\%$, not positive for every seed) and small on ETTh1 ($+1.6\%$). The from-scratch model's MSE is $0.4$ to $5.4\%$ higher than PatchTST's on 7 datasets and $2.6\%$ lower on Exchange.

\begin{table}[H]
\caption{Forecasting MSE without attention across 8 datasets (3 seeds per dataset; mean $\pm$ sample std). Ablated: the trained checkpoint with the attention output set to zero; its change is computed per seed before averaging. From scratch: PatchTST-NoAttn trained with the attention output fixed to zero; its change is computed on the 3-seed means because its runs are not paired with PatchTST seeds.}
\label{tab:noattn_mse}
\centering
\small
\setlength{\tabcolsep}{4pt}
\begin{tabular}{lccccc}
\toprule
Dataset & PatchTST & Ablated & Change (\%) & From scratch & Change (\%) \\
\midrule
ETTm1 & $0.289$\stdv{0.002} & $0.340$\stdv{0.007} & $+17.6$\stdv{1.7} & $0.295$\stdv{0.001} & $+2.0$ \\
ETTm2 & $0.164$\stdv{0.001} & $0.183$\stdv{0.001} & $+11.4$\stdv{0.8} & $0.165$\stdv{0.001} & $+0.6$ \\
ETTh1 & $0.375$\stdv{0.001} & $0.381$\stdv{0.001} & $+1.6$\stdv{0.2} & $0.377$\stdv{0.001} & $+0.4$ \\
ETTh2 & $0.275$\stdv{0.000} & $0.276$\stdv{0.001} & $+0.3$\stdv{0.5} & $0.276$\stdv{0.000} & $+0.5$ \\
Weather & $0.151$\stdv{0.000} & $0.204$\stdv{0.002} & $+34.8$\stdv{1.4} & $0.157$\stdv{0.000} & $+4.3$ \\
Electricity & $0.130$\stdv{0.001} & $0.160$\stdv{0.004} & $+22.5$\stdv{4.0} & $0.135$\stdv{0.000} & $+3.1$ \\
Exchange & $0.092$\stdv{0.001} & $0.103$\stdv{0.005} & $+11.4$\stdv{4.6} & $0.090$\stdv{0.000} & $-2.6$ \\
Traffic & $0.367$\stdv{0.000} & $0.463$\stdv{0.002} & $+26.3$\stdv{0.5} & $0.386$\stdv{0.000} & $+5.4$ \\
\midrule
\textbf{Mean (24 runs)} & & & $\mathbf{+15.7}$ & & \\
\textbf{8-dataset mean MSE} & $0.2305$ & $0.2636$ & $+14.4$ & $0.2352$ & $+2.0$ \\
\bottomrule
\end{tabular}
\end{table}

\paragraph{Without attention.} The 4 datasets with the largest FFN delta change (ETTm1, Weather, Electricity, Traffic; Table~\ref{tab:noattn}) also show the largest ablation costs ($+17.6$ to $+34.8\%$), but ablation costs $+11.4\%$ on ETTm2 and Exchange with almost no change in probing, so the cost does not depend only on how much construction moves from attention to the FFN. Training from scratch removes most of it: the from-scratch model recovers 75 to 95\% of the ablation damage on 6 datasets and beats the ablated model on every dataset except ETTh2. On the 7 datasets shared with RAM~\citep{guo2024ram}, the from-scratch model's MSE at horizon 96 is within $0.003$ of RAM's attention-free PatchTST (mean of per-dataset changes $+2.3\%$, against RAM's $+2.2\%$). The setups are not identical. RAM deletes the attention layer but keeps the feed-forward, residual, and normalization layers, and it averages five runs; its text gives a lookback of 96 but labels the model PatchTST/42, the 336-step setting of \citet{nie2023time}. We fix the attention output to zero, use a lookback of 336, and average three seeds.

\paragraph{Knockout.} For each frozen checkpoint, we zero the output of every attention sub-layer or of every FFN sub-layer, keep the residual connections and normalization, and measure the relative increase in test MSE over the unmodified model at $\text{pred\_len}=96$. A no-op hook reproduces the baseline MSE exactly in every run; Appendix~\ref{app:setup_details} compares the baselines with the published MSEs. For PatchTST, the attention knockout is the ablated NoAttn model of \S\ref{sec:noattn}, and its mean costs match Table~\ref{tab:noattn_mse}. Table~\ref{tab:knockout} reports the costs.

\begin{table}[H]
\caption{Component knockout: MSE increase (\%) when the attention or FFN output is zeroed in all layers of a frozen model (mean $\pm$ sample std over 3 seeds). For CATS, ``Attn'' is cross-attention. Ratio: mean attention cost over the 8 datasets divided by the mean FFN cost.}
\label{tab:knockout}
\centering
\footnotesize
\setlength{\tabcolsep}{4pt}
\begin{tabular}{lcccccc}
\toprule
 & \multicolumn{2}{c}{PatchTST} & \multicolumn{2}{c}{iTransformer} & \multicolumn{2}{c}{CATS} \\
\cmidrule(lr){2-3}\cmidrule(lr){4-5}\cmidrule(lr){6-7}
Dataset & Attn & FFN & Attn & FFN & Attn & FFN \\
\midrule
ETTm1 & $+17.6$\stdv{1.7} & $+53.1$\stdv{8.5} & $+6.4$\stdv{1.5} & $+11.2$\stdv{2.2} & $+122.6$\stdv{4.2} & $+8.3$\stdv{0.5} \\
ETTm2 & $+11.4$\stdv{0.8} & $+22.8$\stdv{2.0} & $+7.5$\stdv{1.2} & $+1.8$\stdv{0.3} & $+29.1$\stdv{1.2} & $+1.0$\stdv{0.5} \\
ETTh1 & $+1.6$\stdv{0.2} & $+19.5$\stdv{2.0} & $+4.9$\stdv{1.1} & $+8.2$\stdv{0.7} & $+85.2$\stdv{0.9} & $+2.6$\stdv{0.3} \\
ETTh2 & $+0.3$\stdv{0.5} & $+12.4$\stdv{0.2} & $+1.6$\stdv{0.5} & $+3.2$\stdv{0.1} & $+19.5$\stdv{0.2} & $+5.2$\stdv{0.6} \\
Weather & $+34.8$\stdv{1.4} & $+36.9$\stdv{1.7} & $+28.8$\stdv{3.0} & $+13.6$\stdv{0.6} & $+31.4$\stdv{0.5} & $+7.9$\stdv{1.6} \\
Electricity & $+22.5$\stdv{4.0} & $+10.2$\stdv{2.2} & $+38.0$\stdv{0.6} & $+30.3$\stdv{1.0} & $+471.9$\stdv{0.4} & $+9.5$\stdv{0.5} \\
Exchange & $+11.4$\stdv{4.6} & $+106.1$\stdv{5.6} & $-1.6$\stdv{1.0} & $+3.6$\stdv{1.7} & $+68.5$\stdv{1.9} & $+1.4$\stdv{0.3} \\
Traffic & $+26.3$\stdv{0.5} & $+20.2$\stdv{1.0} & $+62.3$\stdv{3.0} & $+40.0$\stdv{0.4} & $+234.7$\stdv{0.1} & $+12.6$\stdv{0.4} \\
\midrule
\textbf{8-dataset mean} & $\mathbf{+15.7}$ & $\mathbf{+35.1}$ & $\mathbf{+18.5}$ & $\mathbf{+14.0}$ & $\mathbf{+132.8}$ & $\mathbf{+6.1}$ \\
Ratio (attn/FFN) & \multicolumn{2}{c}{$0.45$} & \multicolumn{2}{c}{$1.32$} & \multicolumn{2}{c}{$21.9$} \\
FFN cost $>$ attn cost & \multicolumn{2}{c}{6 of 8} & \multicolumn{2}{c}{4 of 8} & \multicolumn{2}{c}{0 of 8} \\
\bottomrule
\end{tabular}
\end{table}

\paragraph{Ordering.} The FFN cost orders the architectures PatchTST $>$ iTransformer $>$ CATS and the ratio orders them PatchTST $<$ iTransformer $<$ CATS, the same order as probing; both orderings hold in all 8 leave-one-dataset-out subsets (ratios $0.37$ to $0.65$, $1.19$ to $1.40$, and $15.2$ to $25.4$). CATS's cross-attention costs are large by construction, since cross-attention is its only path from the input. Per dataset, PatchTST relies more on attention for Electricity and Traffic, and iTransformer relies more on the FFN for ETTm1, ETTh1, ETTh2, and Exchange.

\paragraph{Single-scalar tokens.} In PatchTST at $P{=}1$, the FFN cost exceeds the attention cost on every dataset tested, as at $P{=}16$ (Table~\ref{tab:knockout_p}), although probing puts construction in attention at $P{=}1$ (\S\ref{sec:pointwise}). Knockout thus measures what the frozen model relies on, not which component makes the probed properties decodable.

\begin{table}[H]
\caption{PatchTST component knockout at $P{=}1$ and $P{=}16$ (MSE increase \%, mean $\pm$ sample std over 3 seeds; the $P{=}16$ row repeats Table~\ref{tab:knockout}). Weather at $P{=}1$ comes from a separate run with another test batch size, which changes how many windows at the end of the test split are kept; on ETTh1 and ETTm1, that run also gives $+5.1$/$+39.8$ and $+28.2$/$+48.3$ (Attn/FFN), close to the $P{=}1$ row.}
\label{tab:knockout_p}
\centering
\footnotesize
\setlength{\tabcolsep}{4pt}
\begin{tabular}{lcccccccc}
\toprule
 & \multicolumn{2}{c}{ETTh1} & \multicolumn{2}{c}{ETTm1} & \multicolumn{2}{c}{Weather} & \multicolumn{2}{c}{Exchange} \\
\cmidrule(lr){2-3}\cmidrule(lr){4-5}\cmidrule(lr){6-7}\cmidrule(lr){8-9}
$P$ & Attn & FFN & Attn & FFN & Attn & FFN & Attn & FFN \\
\midrule
1 & $+5.8$\stdv{1.0} & $+41.4$\stdv{0.8} & $+28.4$\stdv{8.2} & $+48.4$\stdv{9.1} & $+5.6$\stdv{8.4} & $+34.5$\stdv{4.8} & $+19.2$\stdv{11.2} & $+65.8$\stdv{1.9} \\
16 & $+1.6$\stdv{0.2} & $+19.5$\stdv{2.0} & $+17.6$\stdv{1.7} & $+53.1$\stdv{8.5} & $+34.8$\stdv{1.4} & $+36.9$\stdv{1.7} & $+11.4$\stdv{4.6} & $+106.1$\stdv{5.6} \\
\bottomrule
\end{tabular}
\end{table}

\paragraph{Scope.} Knockout shows what a frozen model relies on; a retrained model can compensate (\S\ref{sec:noattn}). Per-dataset costs vary widely, so the comparison across architectures rests not on single cells but on the ordering of the 8-dataset means, which holds when any one dataset is left out.

\section{Feature-Level Erasure}
\label{app:erasure}

We ask whether the forecaster uses an individual probed property by erasing it from the Layer-1 output (\texttt{1b\_ffn}) of frozen PatchTST ($P{=}16$) and measuring the forecast cost. For each constructed task, iterative nullspace projection (INLP)~\citep{ravfogel2020null} repeatedly trains a 4-class linear probe on the quartile labels and projects the span of the probe's mean-centered weight rows (at most 3 directions) out of the representation, until a probe retrained on the projected representation reaches at most 30\% (chance 25\%) or 40 iterations have run. The projection is applied to every token before Layer 2, and the cost is the increase in test MSE over the unmodified model on 200 test windows; the probes are fit on 200 other windows, whose own quartiles set the class boundaries. As a control, the same procedure runs on randomly permuted labels and stops at the same number of removed dimensions, so that, unlike the random-direction control of amnesic probing~\citep{elazar2021amnesic}, both removals come from the same probe mechanism. The two removals still take different shares of the representation energy (the summed squared norm of the token representations), so we compare the MSE increase per unit of removed energy.

On ETTm1, Weather, and Electricity ($d_{\text{model}}=128$; Table~\ref{tab:erasure}), erasure removes 33 to 94\% of the dimensions; it stays incomplete in the three cells with runs at the iteration cap (ETTm1 volatility and diff\_energy, Weather spectral\_entropy), where the probe after erasure still reaches 34 to 37\%. In these 9 cells, the energy-normalized cost exceeds the control in 6 and falls below it in 3 (Weather diff\_energy, Electricity volatility and diff\_energy), and the difference exceeds one standard deviation over model seeds only for ETTm1 diff\_energy (above the control) and Electricity volatility (below). With $d_{\text{model}}=16$, each INLP step removes up to 3 of the 16 dimensions, so the ETTh1 rows are coarse and shown for direction only. The test is therefore inconclusive about whether the forecaster uses an individual property: the properties may be spread over many directions, or INLP may remove more than the property itself: LEACE erases the linear information about a $k$-class label with an edit of rank at most $k{-}1$~\citep{belrose2023leace}, far fewer dimensions than INLP removes here. Minimal-rank erasure is left for future work.

\begin{table}[H]
\caption{Feature-level erasure in frozen PatchTST ($P{=}16$) at the Layer-1 output \texttt{1b\_ffn} (mean over 3 model seeds; $\pm$: sample std of the cost difference). Removed: share of dimensions removed by INLP. Probe: in-sample accuracy (chance 25\%) of a 4-class linear probe on the 200 fitting windows, trained for 1,500 steps before and 800 steps after erasure; not comparable to the main-text probing accuracies. Energy: share of representation energy removed by erasure and by the random-label control. Cost difference: MSE increase divided by the removed energy share, erasure minus control.}
\label{tab:erasure}
\centering
\small
\begin{tabular}{llcccc}
\toprule
Dataset ($d_{\text{model}}$) & Task & Removed (\%) & Probe (\%) & Energy (\%) & Cost difference \\
\midrule
ETTm1 (128) & volatility & $94$ & $80.6 \to 35.9$ & $52.8$ / $32.0$ & $+0.05$\stdv{0.07} \\
ETTm1 (128) & diff\_energy & $80$ & $62.7 \to 36.5$ & $54.8$ / $16.4$ & $+0.24$\stdv{0.16} \\
ETTm1 (128) & spectral\_entropy & $33$ & $76.9 \to 30.3$ & $2.2$ / $0.6$ & $+0.22$\stdv{0.26} \\
Weather (128) & volatility & $66$ & $68.3 \to 31.3$ & $12.4$ / $3.1$ & $+0.04$\stdv{0.09} \\
Weather (128) & diff\_energy & $55$ & $62.7 \to 31.7$ & $7.9$ / $2.2$ & $-0.11$\stdv{0.17} \\
Weather (128) & spectral\_entropy & $93$ & $66.4 \to 33.9$ & $75.6$ / $41.4$ & $+0.06$\stdv{0.09} \\
Electricity (128) & volatility & $80$ & $82.8 \to 29.8$ & $34.3$ / $67.9$ & $-0.15$\stdv{0.01} \\
Electricity (128) & diff\_energy & $84$ & $74.5 \to 28.7$ & $44.4$ / $72.8$ & $-0.03$\stdv{0.04} \\
Electricity (128) & spectral\_entropy & $84$ & $64.7 \to 29.5$ & $45.6$ / $72.5$ & $+0.02$\stdv{0.08} \\
\midrule
ETTh1 (16) & volatility & $88$ & $38.2 \to 25.1$ & $67.1$ / $72.6$ & $-0.05$\stdv{0.06} \\
ETTh1 (16) & diff\_energy & $88$ & $43.6 \to 29.6$ & $70.8$ / $67.4$ & $-0.04$\stdv{0.08} \\
ETTh1 (16) & spectral\_entropy & $63$ & $40.0 \to 29.7$ & $25.9$ / $22.7$ & $+0.14$\stdv{0.03} \\
\bottomrule
\end{tabular}
\end{table}

\clearpage

\section*{NeurIPS Paper Checklist}

\begin{enumerate}

\item {\bf Claims}
    \item[] Question: Do the main claims made in the abstract and introduction accurately reflect the paper's contributions and scope?
    \item[] Answer: \answerYes{}
    \item[] Justification: The abstract and \S\ref{sec:intro} state our three contributions (temporal probing framework, role allocation by tokenization and architecture, and a representational reading of prior results), which are supported in \S\ref{sec:results_start}, with the reading of prior results in \S\ref{sec:unified}. Forecasting-side interventions (activation patching, training without attention, and component knockout) are stated as establishing component-level reliance and sufficiency in the standard PatchTST, not feature-level causal construction; across architectures, knockout is stated only to match the ordering seen in probing, and whether individual probed features are used is stated as open (\S\ref{sec:limitations}).

\item {\bf Limitations}
    \item[] Question: Does the paper discuss the limitations of the work performed by the authors?
    \item[] Answer: \answerYes{}
    \item[] Justification: \S\ref{sec:limitations} states (i) the reach of the forecasting-side evidence: component-level reliance and sufficiency in PatchTST at $P{=}16$, no verdict on the shift seen in probing at $P{=}1$, only the ordering across architectures, and an inconclusive feature-level erasure test (Appendix~\ref{app:erasure}); (ii) the scope of the allocation: what varies, what does not, which regimes lie outside, and that lookback, depth, and the form of the mixer were not varied; (iii) the limits of the task suite: it is hand-picked to make construction visible, and the main conclusions rest on three second-order statistics of a token's segment (16 points in PatchTST); (iv) the conditions under which the framework applies, with pretrained foundation models left to future work; and (v) the absence of a formal theory: the phenomenon and its boundary conditions are established experimentally, not proven.

\item {\bf Theory assumptions and proofs}
    \item[] Question: For each theoretical result, does the paper provide the full set of assumptions and a complete (and correct) proof?
    \item[] Answer: \answerNA{}
    \item[] Justification: This paper presents an empirical study without formal theorems or proofs. The structural arguments for patch tokens (\S\ref{sec:ffn_dominance}) and single-scalar tokens (\S\ref{sec:pointwise}) are informal explanations grounded in the architecture, not theoretical results (\S\ref{sec:limitations}).

\item {\bf Experimental result reproducibility}
    \item[] Question: Does the paper fully disclose all the information needed to reproduce the main experimental results of the paper to the extent that it affects the main claims and/or conclusions of the paper (regardless of whether the code and data are provided or not)?
    \item[] Answer: \answerYes{}
    \item[] Justification: The forecasters are trained with each model's official code and scripts (Informer with the implementation in the PatchTST code base); Appendix~\ref{app:setup_details} gives their architecture-level and per-dataset architectural settings (Tables~\ref{tab:hparams_arch} to~\ref{tab:hparams_itrans} and the per-model paragraphs) and how probe samples are drawn, labeled, and split. \S\ref{sec:protocol} gives the probe training protocol (optimizer, learning rate, batch size, and step count), Appendices~\ref{app:patching} to~\ref{app:erasure} the forecasting-side interventions, and \S\ref{sec:datasets} the 8 standard public benchmarks with standard splits. The probing results of \S\ref{sec:results_start} are 9-run means (3 model $\times$ 3 probe seeds), and the forecasting-side interventions use 3 model seeds.

\item {\bf Open access to data and code}
    \item[] Question: Does the paper provide open access to the data and code, with sufficient instructions to faithfully reproduce the main experimental results, as described in supplemental material?
    \item[] Answer: \answerYes{}
    \item[] Justification: The code and the per-run result files are released at \url{https://github.com/mingyu-11/tokenization-component-roles}, with instructions for training the models, running the probes, and recomputing the tables. All datasets used (ETT, Weather, Electricity, Exchange, Traffic, Solar) are publicly available standard time series forecasting benchmarks, which we use following the PatchTST protocol, and are not redistributed. The paper also gives the probe architecture and training protocol (\S\ref{sec:protocol}), the per-dataset architectural hyperparameters (Appendix~\ref{app:setup_details}, Tables~\ref{tab:hparams_arch} to~\ref{tab:hparams_itrans}), the corruption procedure for activation patching (Appendix~\ref{app:patching}), and the seeds (3 model $\times$ 3 probe seeds for the probing results of \S\ref{sec:results_start}; 3 model seeds for the forecasting-side interventions).

\item {\bf Experimental setting/details}
    \item[] Question: Does the paper specify all the training and test details (e.g., data splits, hyperparameters, how they were chosen, type of optimizer) necessary to understand the results?
    \item[] Answer: \answerYes{}
    \item[] Justification: \S\ref{sec:setup} specifies models, datasets, and ablations. \S\ref{sec:protocol} reports the probe training protocol (Adam, lr=$10^{-3}$, batch=32{,}768, 30k steps, 80/20 split; evaluated every 500 steps, with the best test accuracy over the entire training trajectory reported), and Appendix~\ref{app:setup_details} describes how probe samples are drawn, labeled, and split. The forecasters are trained with each model's official code and scripts (Informer with the implementation in the PatchTST code base); Appendix~\ref{app:setup_details} gives their architecture-level and per-dataset architectural settings and the settings used for datasets the scripts do not cover.

\item {\bf Experiment statistical significance}
    \item[] Question: Does the paper report error bars suitably and correctly defined or other appropriate information about the statistical significance of the experiments?
    \item[] Answer: \answerYes{}
    \item[] Justification: Table~\ref{tab:patchtst_delta} and the per-dataset appendix probing tables behind Table~\ref{tab:noattn_r2} and Figure~\ref{fig:component_dominance} report 9-run means over 3 model seeds $\times$ 3 probe seeds (the probe seed sets the split, the initialization, and the batch draws; Appendix~\ref{app:setup_details}), with $\pm$ built from 9-run population standard deviations (for deltas, the standard deviations of the two probing points combined as in Table~\ref{tab:patchtst_delta}; see Appendix~\ref{app:full}), and the forecasting tables of Appendices~\ref{app:knockout} and~\ref{app:erasure} report means $\pm$ sample standard deviations over 3 seeds. Statistical tests include paired $t$-tests with Cohen's $d_z$ for the Layer-1 selectivity gain (\S\ref{sec:innate_constructed} and Appendix~\ref{app:selectivity}: $d_z = +1.67$, one-sided $p < 10^{-6}$, pooled over 24 dataset-task pairs) and for activation patching (Appendix~\ref{app:patching}: $d_z$ from $-0.85$ to $+3.30$ over 300 window-seed pairs and $p < 10^{-16}$ on the 100 seed-averaged windows per dataset, which the test treats as independent although they overlap).

\item {\bf Experiments compute resources}
    \item[] Question: For each experiment, does the paper provide sufficient information on the computer resources (type of compute workers, memory, time of execution) needed to reproduce the experiments?
    \item[] Answer: \answerYes{}
    \item[] Justification: Experiments ran on NVIDIA A100 (40GB), A40 (48GB), and RTX A6000 (48GB) GPUs (Appendix~\ref{app:setup_details}). Training a forecaster takes from minutes on ETTh1 to over 100 GPU-hours per seed for PatchTST at $P{=}1$ on Traffic, a probing run takes minutes, and activation patching takes under an hour per dataset. All reported results took over 700 GPU-hours, and the full project used more.

\item {\bf Code of ethics}
    \item[] Question: Does the research conducted in the paper conform, in every respect, with the NeurIPS Code of Ethics \url{https://neurips.cc/public/EthicsGuidelines}?
    \item[] Answer: \answerYes{}
    \item[] Justification: This work uses only publicly available time series forecasting benchmarks and does not involve human subjects, personal data, or sensitive applications. All authors have reviewed the NeurIPS Code of Ethics.

\item {\bf Broader impacts}
    \item[] Question: Does the paper discuss both potential positive societal impacts and negative societal impacts of the work performed?
    \item[] Answer: \answerNA{}
    \item[] Justification: This paper presents foundational research on the internal workings of time series transformers. It introduces no new datasets, deploys no models, and develops no applications, and we see no direct path from it to negative societal applications.

\item {\bf Safeguards}
    \item[] Question: Does the paper describe safeguards that have been put in place for responsible release of data or models that have a high risk for misuse (e.g., pre-trained language models, image generators, or scraped datasets)?
    \item[] Answer: \answerNA{}
    \item[] Justification: The paper does not release any models, datasets, or other high-risk artifacts. It analyzes forecasters that we train with the public code of PatchTST (which also provides our Informer implementation), iTransformer, CATS, and ModernTCN on standard forecasting benchmarks.

\item {\bf Licenses for existing assets}
    \item[] Question: Are the creators or original owners of assets (e.g., code, data, models), used in the paper, properly credited and are the license and terms of use explicitly mentioned and properly respected?
    \item[] Answer: \answerYes{}
    \item[] Justification: Code: PatchTST~\citep{nie2023time} (Apache-2.0), whose code base also provides the Informer~\citep{zhou2021informer} implementation we train, and iTransformer~\citep{liu2024itransformer}, CATS~\citep{kim2024cats}, and ModernTCN~\citep{luo2024moderntcn} (MIT); we train every forecaster ourselves and use no released weights. Data (\S\ref{sec:datasets}): ETT~\citep{zhou2021informer} (CC BY-ND 4.0); Weather~\citep{wu2021autoformer}, recorded by the Max Planck Institute for Biogeochemistry, Jena (CC BY 4.0); and, as prepared by \citet{lai2018modeling}, Electricity from the UCI ElectricityLoadDiagrams20112014 data (CC BY 4.0), Traffic from Caltrans PeMS (public domain under its conditions of use), Exchange (no license stated), and Solar-Energy from NREL's Solar Power Data for Integration Studies, for which, as NREL's data terms ask, we credit the U.S.\ Department of Energy (DOE)/NREL/ALLIANCE. No dataset is redistributed, and the released code keeps each upstream license.

\item {\bf New assets}
    \item[] Question: Are new assets introduced in the paper well documented and is the documentation provided alongside the assets?
    \item[] Answer: \answerYes{}
    \item[] Justification: The released code and per-run result files are documented in the repository README, and the repository states the license of our code (MIT) and those of the included upstream code. No new datasets or pre-trained models are released. The repository is \url{https://github.com/mingyu-11/tokenization-component-roles}.

\item {\bf Crowdsourcing and research with human subjects}
    \item[] Question: For crowdsourcing experiments and research with human subjects, does the paper include the full text of instructions given to participants and screenshots, if applicable, as well as details about compensation (if any)?
    \item[] Answer: \answerNA{}
    \item[] Justification: The paper does not involve crowdsourcing or research with human subjects.

\item {\bf Institutional review board (IRB) approvals or equivalent for research with human subjects}
    \item[] Question: Does the paper describe potential risks incurred by study participants, whether such risks were disclosed to the subjects, and whether Institutional Review Board (IRB) approvals (or an equivalent approval/review based on the requirements of your country or institution) were obtained?
    \item[] Answer: \answerNA{}
    \item[] Justification: The paper does not involve human subjects research; no IRB review applies.

\item {\bf Declaration of LLM usage}
    \item[] Question: Does the paper describe the usage of LLMs if it is an important, original, or non-standard component of the core methods in this research? Note that if the LLM is used only for writing, editing, or formatting purposes and does \emph{not} impact the core methodology, scientific rigor, or originality of the research, declaration is not required.
    \item[] Answer: \answerNA{}
    \item[] Justification: LLMs are not a component of the core methodology. The probing framework, ablation experiments, and activation patching analyses use standard linear/MLP probes and intervention techniques without LLMs.

\end{enumerate}


\begin{thebibliography}{99}

\bibitem[Alain and Bengio(2016)]{alain2016understanding}
Guillaume Alain and Yoshua Bengio.
Understanding intermediate layers using linear classifier probes.
\emph{arXiv preprint arXiv:1610.01644}, 2016.

\bibitem[Bao et~al.(2026)]{bao2026universal}
Anthony Bao, Venkata~Hasith Vattikuti, Jeffrey Lai, and William Gilpin.
Universal redundancies in time series foundation models.
In \emph{ICML}, 2026.

\bibitem[Belinkov(2022)]{belinkov2022probing}
Yonatan Belinkov.
Probing classifiers: Promises, shortcomings, and advances.
\emph{Computational Linguistics}, 48(1):207--219, 2022.

\bibitem[Belrose et~al.(2023)]{belrose2023leace}
Nora Belrose, David Schneider-Joseph, Shauli Ravfogel, Ryan Cotterell, Edward Raff, and Stella Biderman.
LEACE: Perfect linear concept erasure in closed form.
In \emph{NeurIPS}, 2023.

\bibitem[Chen et~al.(2025a)]{chen2025distributional}
Lei Chen, Joan Bruna, and Alberto Bietti.
Distributional associations vs in-context reasoning: A study of feed-forward and attention layers.
In \emph{ICLR}, 2025.

\bibitem[Chen et~al.(2025b)]{chen2025closer}
Yu Chen, Nathalia C\'espedes, and Payam Barnaghi.
A closer look at transformers for time series forecasting: Understanding why they work and where they struggle.
In \emph{ICML}, 2025.

\bibitem[Conneau et~al.(2018)]{conneau2018what}
Alexis Conneau, German Kruszewski, Guillaume Lample, Lo{\"\i}c Barrault, and Marco Baroni.
What you can cram into a single \$\&!\#* vector: Probing sentence embeddings for linguistic properties.
In \emph{ACL}, pages 2126--2136, 2018.

\bibitem[Dong et~al.(2025)]{dong2025attention}
Yihe Dong, Lorenzo Noci, Mikhail Khodak, and Mufan Li.
Is random attention sufficient for sequence modeling? Disentangling trainable components in the transformer.
\emph{arXiv preprint arXiv:2506.01115}, 2025.

\bibitem[Ekambaram et~al.(2023)]{ekambaram2023tsmixer}
Vijay Ekambaram, Arindam Jati, Nam Nguyen, Phanwadee Sinthong, and Jayant Kalagnanam.
TSMixer: Lightweight MLP-Mixer model for multivariate time series forecasting.
In \emph{KDD}, pages 459--469, 2023.

\bibitem[Elazar et~al.(2021)]{elazar2021amnesic}
Yanai Elazar, Shauli Ravfogel, Alon Jacovi, and Yoav Goldberg.
Amnesic probing: Behavioral explanation with amnesic counterfactuals.
\emph{Transactions of the Association for Computational Linguistics}, 9:160--175, 2021.

\bibitem[Elhage et~al.(2021)]{elhage2021mathematical}
Nelson Elhage, Neel Nanda, Catherine Olsson, Tom Henighan, Nicholas Joseph, Ben Mann, Amanda Askell, Yuntao Bai, Anna Chen, Tom Conerly, Nova DasSarma, Dawn Drain, Deep Ganguli, Zac Hatfield-Dodds, Danny Hernandez, Andy Jones, Jackson Kernion, Liane Lovitt, Kamal Ndousse, Dario Amodei, Tom Brown, Jack Clark, Jared Kaplan, Sam McCandlish, and Chris Olah.
A mathematical framework for transformer circuits.
\emph{Transformer Circuits Thread}, 2021.
URL \url{https://transformer-circuits.pub/2021/framework/index.html}.

\bibitem[Geva et~al.(2021)]{geva2021kvmemories}
Mor Geva, Roei Schuster, Jonathan Berant, and Omer Levy.
Transformer feed-forward layers are key-value memories.
In \emph{EMNLP}, pages 5484--5495, 2021.

\bibitem[Geva et~al.(2022)]{geva2022transformer}
Mor Geva, Avi Caciularu, Kevin Ro Wang, and Yoav Goldberg.
Transformer feed-forward layers build predictions by promoting concepts in the vocabulary space.
In \emph{EMNLP}, pages 30--45, 2022.

\bibitem[Guo et~al.(2024)]{guo2024ram}
Suhan Guo, Jiahong Deng, Yi Wei, Hui Dou, Furao Shen, and Jian Zhao.
RAM: Replace attention with MLP for efficient multivariate time series forecasting.
\emph{arXiv preprint arXiv:2410.24023}, 2024.

\bibitem[Han et~al.(2024)Han, Ye, and Zhan]{han2024capacity}
Lu Han, Han-Jia Ye, and De-Chuan Zhan.
The capacity and robustness trade-off: Revisiting the channel independent strategy for multivariate time series forecasting.
\emph{IEEE Transactions on Knowledge and Data Engineering}, 36(11):7129--7142, 2024.

\bibitem[Heimersheim and Nanda(2024)]{heimersheim2024use}
Stefan Heimersheim and Neel Nanda.
How to use and interpret activation patching.
\emph{arXiv preprint arXiv:2404.15255}, 2024.

\bibitem[Hewitt and Liang(2019)]{hewitt2019designing}
John Hewitt and Percy Liang.
Designing and interpreting probes with control tasks.
In \emph{EMNLP}, pages 2733--2743, 2019.

\bibitem[Hewitt and Manning(2019)]{hewitt2019structural}
John Hewitt and Christopher~D. Manning.
A structural probe for finding syntax in word representations.
In \emph{NAACL-HLT}, pages 4129--4138, 2019.

\bibitem[Hewitt et~al.(2021)]{hewitt2021conditional}
John Hewitt, Kawin Ethayarajh, Percy Liang, and Christopher~D. Manning.
Conditional probing: Measuring usable information beyond a baseline.
In \emph{EMNLP}, pages 1626--1639, 2021.

\bibitem[Ilbert et~al.(2024)]{ilbert2024samformer}
Romain Ilbert, Ambroise Odonnat, Vasilii Feofanov, Aladin Virmaux, Giuseppe Paolo, Themis Palpanas, and Ievgen Redko.
SAMformer: Unlocking the potential of transformers in time series forecasting with sharpness-aware minimization and channel-wise attention.
In \emph{ICML}, 2024.

\bibitem[Kaufman and Azencot(2024)]{kaufman2024analyzing}
Ilya Kaufman and Omri Azencot.
Analyzing deep transformer models for time series forecasting via manifold learning.
\emph{Transactions on Machine Learning Research}, 2024.

\bibitem[Kim et~al.(2024)]{kim2024cats}
Dongbin Kim, Jinseong Park, Jaewook Lee, and Hoki Kim.
Are self-attentions effective for time series forecasting?
In \emph{NeurIPS}, 2024.

\bibitem[Kim et~al.(2022)]{kim2022reversible}
Taesung Kim, Jinhee Kim, Yunwon Tae, Cheonbok Park, Jang-Ho Choi, and Jaegul Choo.
Reversible instance normalization for accurate time-series forecasting against distribution shift.
In \emph{ICLR}, 2022.

\bibitem[Lai et~al.(2018)]{lai2018modeling}
Guokun Lai, Wei-Cheng Chang, Yiming Yang, and Hanxiao Liu.
Modeling long- and short-term temporal patterns with deep neural networks.
In \emph{SIGIR}, pages 95--104, 2018.

\bibitem[Li and Janson(2024)]{li2024optimal}
Maximilian Li and Lucas Janson.
Optimal ablation for interpretability.
In \emph{NeurIPS}, 2024.

\bibitem[Liang et~al.(2025)]{liang2025why}
Zida Liang, Jiayi Zhu, and Weiqiang Sun.
Why attention fails: The degeneration of transformers into MLPs in time series forecasting.
\emph{arXiv preprint arXiv:2509.20942}, 2025.

\bibitem[Liu et~al.(2024)]{liu2024itransformer}
Yong Liu, Tengge Hu, Haoran Zhang, Haixu Wu, Shiyu Wang, Lintao Ma, and Mingsheng Long.
iTransformer: Inverted transformers are effective for time series forecasting.
In \emph{ICLR}, 2024.

\bibitem[Luo and Wang(2024)]{luo2024moderntcn}
Donghao Luo and Xue Wang.
ModernTCN: A modern pure convolution structure for general time series analysis.
In \emph{ICLR}, 2024.

\bibitem[Meng et~al.(2022)]{meng2022locating}
Kevin Meng, David Bau, Alex Andonian, and Yonatan Belinkov.
Locating and editing factual associations in GPT.
In \emph{NeurIPS}, 2022.

\bibitem[Nie et~al.(2023)]{nie2023time}
Yuqi Nie, Nam~H. Nguyen, Phanwadee Sinthong, and Jayant Kalagnanam.
A time series is worth 64 words: Long-term forecasting with transformers.
In \emph{ICLR}, 2023.

\bibitem[Olsson et~al.(2022)]{olsson2022induction}
Catherine Olsson, Nelson Elhage, Neel Nanda, Nicholas Joseph, Nova DasSarma, Tom Henighan, Ben Mann, Amanda Askell, Yuntao Bai, Anna Chen, Tom Conerly, Dawn Drain, Deep Ganguli, Zac Hatfield-Dodds, Danny Hernandez, Scott Johnston, Andy Jones, Jackson Kernion, Liane Lovitt, Kamal Ndousse, Dario Amodei, Tom Brown, Jack Clark, Jared Kaplan, Sam McCandlish, and Chris Olah.
In-context learning and induction heads.
\emph{Transformer Circuits Thread}, 2022.
URL \url{https://transformer-circuits.pub/2022/in-context-learning-and-induction-heads/index.html}.

\bibitem[Pimentel et~al.(2020)]{pimentel2020information}
Tiago Pimentel, Josef Valvoda, Rowan Hall~Maudslay, Ran Zmigrod, Adina Williams, and Ryan Cotterell.
Information-theoretic probing for linguistic structure.
In \emph{ACL}, pages 4609--4622, 2020.

\bibitem[Ravfogel et~al.(2020)]{ravfogel2020null}
Shauli Ravfogel, Yanai Elazar, Hila Gonen, Michael Twiton, and Yoav Goldberg.
Null it out: Guarding protected attributes by iterative nullspace projection.
In \emph{ACL}, pages 7237--7256, 2020.

\bibitem[Tang and Zhang(2025)]{tang2025patchmlp}
Peiwang Tang and Weitai Zhang.
Unlocking the power of patch: Patch-based MLP for long-term time series forecasting.
In \emph{AAAI}, 2025.

\bibitem[Tenney et~al.(2019)]{tenney2019bert}
Ian Tenney, Dipanjan Das, and Ellie Pavlick.
BERT rediscovers the classical NLP pipeline.
In \emph{ACL}, pages 4593--4601, 2019.


\bibitem[Tolstikhin et~al.(2021)]{tolstikhin2021mlp}
Ilya~O. Tolstikhin, Neil Houlsby, Alexander Kolesnikov, Lucas Beyer, Xiaohua Zhai, Thomas Unterthiner, Jessica Yung, Andreas Steiner, Daniel Keysers, Jakob Uszkoreit, Mario Lucic, and Alexey Dosovitskiy.
MLP-Mixer: An all-MLP architecture for vision.
In \emph{NeurIPS}, 2021.
\bibitem[van~Sprang et~al.(2024)]{vansprang2024concept}
Angela van Sprang, Erman Acar, and Willem Zuidema.
Interpretability for time series transformers using a concept bottleneck framework.
\emph{arXiv preprint arXiv:2410.06070}, 2024.

\bibitem[Vaswani et~al.(2017)]{vaswani2017attention}
Ashish Vaswani, Noam Shazeer, Niki Parmar, Jakob Uszkoreit, Llion Jones, Aidan~N. Gomez, {\L}ukasz Kaiser, and Illia Polosukhin.
Attention is all you need.
In \emph{NeurIPS}, pages 5998--6008, 2017.

\bibitem[Voita et~al.(2019)]{voita2019analyzing}
Elena Voita, David Talbot, Fedor Moiseev, Rico Sennrich, and Ivan Titov.
Analyzing multi-head self-attention: Specialized heads do the heavy lifting, the rest can be pruned.
In \emph{ACL}, pages 5797--5808, 2019.


\bibitem[Wang et~al.(2025)]{wang2025pattern}
Chengxin Wang, Yiran Zhao, Shaofeng Cai, and Gary Tan.
Investigating pattern neurons in urban time series forecasting.
In \emph{ICLR}, 2025.

\bibitem[Wang et~al.(2023)]{wang2023interpretability}
Kevin~Ro Wang, Alexandre Variengien, Arthur Conmy, Buck Shlegeris, and Jacob Steinhardt.
Interpretability in the wild: A circuit for indirect object identification in {GPT-2} small.
In \emph{ICLR}, 2023.

\bibitem[Wang et~al.(2026)]{wang2026combinationts}
Xiaorui Wang, Fanda Fan, Chenxi Wang, Yuxuan Yang, Rui Tang, Kuoyu Gao, Simiao Pang, Yuanfeng Shang, Zhipeng Liu, Wanling Gao, Lei Wang, and Jianfeng Zhan.
CombinationTS: A modular framework for understanding time-series forecasting models.
In \emph{ICML}, 2026.

\bibitem[Wili{\'n}ski et~al.(2025)]{wilinski2025exploring}
Micha{\l} Wili{\'n}ski, Mononito Goswami, Willa Potosnak, Nina {\.Z}ukowska, and Artur Dubrawski.
Exploring representations and interventions in time series foundation models.
In \emph{ICML}, pages 66861--66886, 2025.

\bibitem[Wu et~al.(2026)]{wu2025transformers}
Dennis Wu, Yihan He, Yuan Cao, Jianqing Fan, and Han Liu.
Understanding transformers for time series forecasting: A case study on {Moirai}.
In \emph{ICLR}, 2026.

\bibitem[Wu et~al.(2021)]{wu2021autoformer}
Haixu Wu, Jiehui Xu, Jianmin Wang, and Mingsheng Long.
Autoformer: Decomposition transformers with auto-correlation for long-term series forecasting.
In \emph{NeurIPS}, 2021.

\bibitem[Xu et~al.(2020)]{xu2020theory}
Yilun Xu, Shengjia Zhao, Jiaming Song, Russell Stewart, and Stefano Ermon.
A theory of usable information under computational constraints.
In \emph{ICLR}, 2020.

\bibitem[Y{\i}ld{\i}r{\i}m(2026)]{yildirim2026superposition}
Alper Y{\i}ld{\i}r{\i}m.
Superposition is not necessary: A mechanistic interpretability analysis of transformer representations for time series forecasting.
\emph{arXiv preprint arXiv:2605.05151}, 2026.

\bibitem[Yu et~al.(2026)]{yu2026understanding}
Annan Yu, Danielle~C. Maddix, Boran Han, Xiyuan Zhang, Abdul Fatir Ansari, Oleksandr Shchur, Christos Faloutsos, Andrew Gordon Wilson, Michael~W. Mahoney, and Yuyang Wang.
Understanding transformers for time series: Rank structure, flow-of-ranks, and compressibility.
In \emph{ICLR}, 2026.

\bibitem[Zeng et~al.(2023)]{zeng2023transformers}
Ailing Zeng, Muxi Chen, Lei Zhang, and Qiang Xu.
Are transformers effective for time series forecasting?
In \emph{AAAI}, 2023.

\bibitem[Zhang and Nanda(2024)]{zhang2024best}
Fred Zhang and Neel Nanda.
Towards best practices of activation patching in language models: Metrics and methods.
In \emph{ICLR}, 2024.

\bibitem[Zhou et~al.(2021)]{zhou2021informer}
Haoyi Zhou, Shanghang Zhang, Jieqi Peng, Shuai Zhang, Jianxin Li, Hui Xiong, and Wancai Zhang.
Informer: Beyond efficient transformer for long sequence time-series forecasting.
In \emph{AAAI}, 2021.

\bibitem[Zhou et~al.(2022)]{zhou2022fedformer}
Tian Zhou, Ziqing Ma, Qingsong Wen, Xue Wang, Liang Sun, and Rong Jin.
FEDformer: Frequency enhanced decomposed transformer for long-term series forecasting.
In \emph{ICML}, 2022.

\end{thebibliography}
\end{document}